\documentclass{article}

\PassOptionsToPackage{numbers, compress}{natbib}

\usepackage[eandd, final]{neurips_2026}

\usepackage[utf8]{inputenc} 
\usepackage[T1]{fontenc}    
\usepackage{hyperref}       
\usepackage{url}            
\usepackage{booktabs}       
\usepackage{amsfonts}       
\usepackage{nicefrac}       
\usepackage{microtype}      
\usepackage[table,dvipsnames]{xcolor}         

\usepackage{xspace}
\usepackage{graphicx}
\usepackage{multirow}
\newcommand{\tinycolorbox}[2]{%
  \begingroup
  \setlength{\fboxsep}{1pt}%
  \colorbox{#1}{#2}%
  \endgroup
}
\usepackage{pifont}
\newcommand{\xmark}{\ding{55}}
\newcommand{\cmark}{\ding{51}}
\usepackage[most]{tcolorbox}
\usepackage{caption}
\usepackage{subcaption}
\usepackage{enumitem}

\def\ourmethod{OpenView }
\def\ourmethodnospace{OpenView}
\def\ourbench{OpenView-Bench }
\def\ourbenchnospace{OpenView-Bench}
\def\ourdataset{OpenView-Dataset }
\def\ourdatasetnospace{OpenView-Dataset}

\makeatletter
\DeclareRobustCommand\onedot{\futurelet\@let@token\@onedot}
\def\@onedot{\ifx\@let@token.\else.\null\fi\xspace}

\def\eg{\emph{e.g}\onedot} 
\def\ie{\emph{i.e}\onedot}

\makeatother

\title{OpenView: Empowering MLLMs with\\ Out-of-view VQA}

\author{%
  Qixiang Chen$^{1}$ \quad
  Cheng Zhang$^{1}$ \quad
  Chi-Wing Fu$^{2}$ \quad
  Jingwen Ye$^{1}$ \quad
  Jianfei Cai$^{1}$ \\[2pt]
  $^{1}$Monash University \quad
  $^{2}$The Chinese University of Hong Kong
}

\begin{document}

\maketitle

\begin{abstract}
Recent multimodal large language models (MLLMs) show great potential in natural image understanding.
Yet, they perform well, mainly on reasoning in-view contents within the image frame. 
This paper presents the first study on out-of-view (OOV) understanding, i.e., the ability to reason about objects, activities, and scenes beyond the visible frame of a perspective view.
Our technical contributions are threefold.
First, we design \textbf{\ourmethodnospace}, a four-stage pipeline that leverages panoramic imagery for large-scale and diverse multi-choice VQA synthesis, where full-scene coverage enables flexible view framing while preserving global scene awareness.
Second, we curate \textbf{\ourdatasetnospace}, a high-quality synthetic dataset from diverse real-world panoramas to empower MLLMs upon supervised fine-tuning.
Third, we build \textbf{\ourbenchnospace}, a benchmark that jointly measures choice and rationale accuracy for interpretable and diagnosable evaluation.
Experimental results show that despite having a large gap from human performance in OOV VQA answer selection, multiple MLLMs, upon empowered by~\ourmethodnospace, can consistently boost their performance, uplifted from 48.3\% to 62.3\% on average. 
Code, benchmark, and data are available at \textcolor{magenta}{\url{https://github.com/q1xiangchen/OpenView}}.
\end{abstract}
    
\section{Introduction}
\label{sec:intro}
When people browse photo albums, unseen surroundings can be intuitively imagined and memorable moments can also be recalled.
This process reveals how humans integrate visual cues (\eg, scene structure, camera viewpoint, object appearance, and lighting)
with world knowledge to form probabilistic expectations about unseen regions out of the visible field of view (FoV).
This out-of-view (OOV) capability represents a fundamental component of human visual intelligence.
Yet, a key question arises: \textit{Can Multimodal Large Language Models (MLLMs) demonstrate a similar ability?}


Existing benchmarks for evaluating MLLMs on visual question answering (VQA) cover a wide range of domains,~\eg, mathematics~\cite{lu2023mathvista,wang2024measuring,xiao2024logicvista}, document understanding~\cite{masry2022chartqa,liu2024ocrbench,wang2024charxiv,yang2025effective}, video comprehension~\cite{wu2024longvideobench,fu2025video,lin2025towards,zhu2025vau}, spatial reasoning~\cite{ray2024sat,tong2024cambrian,yang2025thinking,gholami2025spatial,dongfang2025multimodal} general knowledge~\cite{Antol_2015_ICCV,Goyal_2017_CVPR,singh2019towards,Marino_2019_CVPR,hudson2019gqa,schwenk2022okvqa,zhang2024mme,zellers2019recognition}, and other specific aspects~\cite{liao2025thinking,li2023evaluating,Guan_2024_CVPR,chen2024we,ke2025explain,ke2026view2space}, as well as multi-disciplinary combinations to access all-round capability~\cite{yu2023mm,li2024seed,liu2024mmbench,yue2024mmmu,mialon2024gaia,zhang2024provision,ying2024mmt}. 
However, all of them focus primarily on the visual contents within the given image (in-view), without explicitly challenging MLLMs to infer plausible unseen content beyond the image frame (out-of-view).
OOV understanding bridges this gap by explicitly requiring models to integrate visible cues with contextual priors to reason about content beyond the captured frame.
It is essential for tasks that require anticipating information beyond the captured view, ~\eg, image outpainting, where generators extrapolate plausible scene continuations consistent with visible cues~\cite{Tang2023mvdiffusion,zheng2025panorama};
world modeling, which reasons about partially-observed environments to predict future states~\cite{bar2025navigation,dong2025unified, yang2025mindjourney,team2025hunyuanworld,hyworld22026};
Yet, existing MLLMs show limited capabilities in OOV reasoning, as current benchmarks and provide insufficient guidance to foster such contextually and spatially conditioned visual reasoning.


\begin{figure}[t]
    \centering
    \includegraphics[trim=0 8 13 0, clip, width=0.9\linewidth]{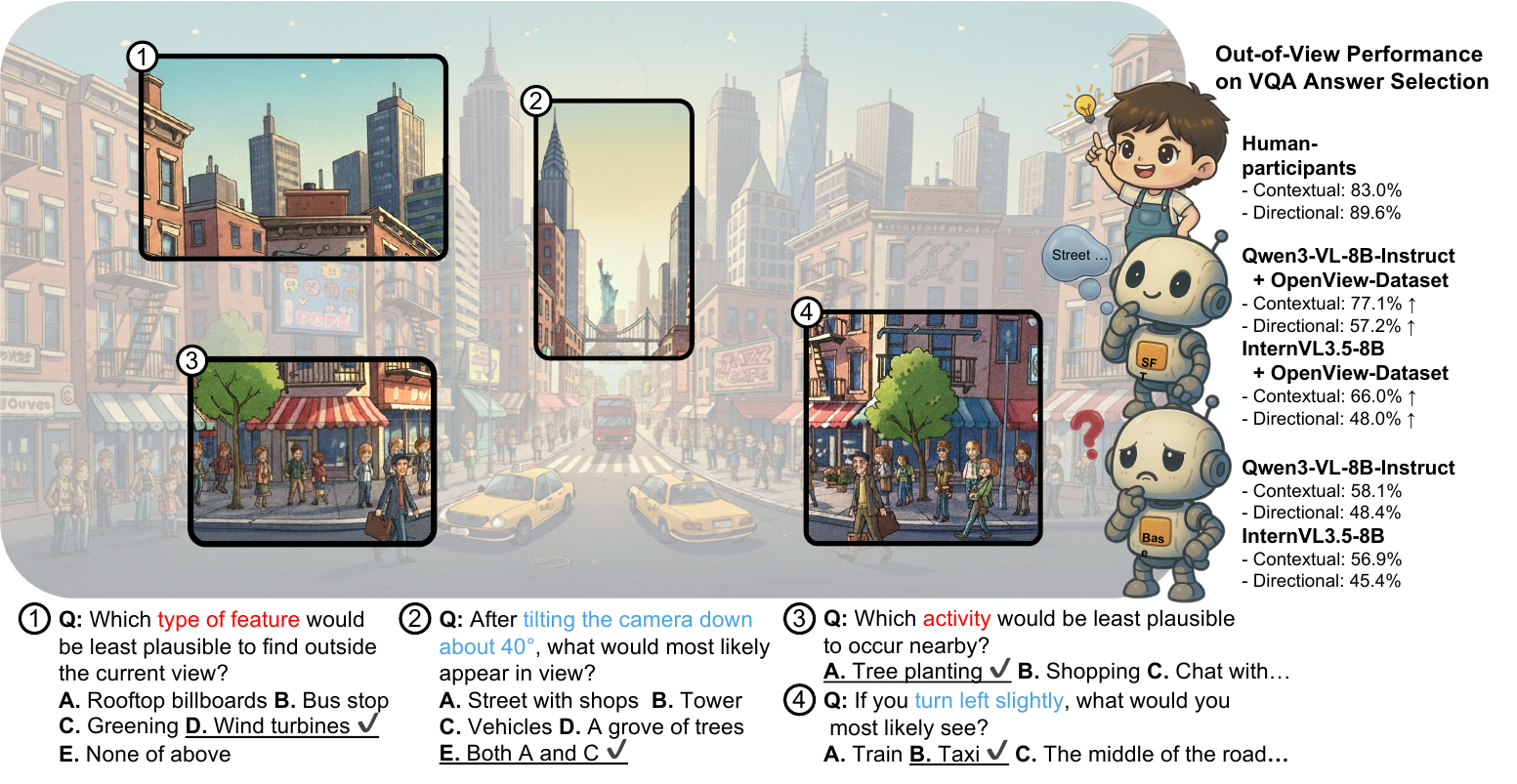}
    \caption{\textbf{Examples of out-of-view visual question answering (VQA).} \ourmethod synthesizes multi-choice out-of-view VQA data, featuring both \textcolor{red}{contextual} and \textcolor{Cerulean}{directional} question types. Empowered with \ourmethodnospace, 
    several MLLMs can largely improve their performance (mid right vs. bottom right) on out-of-view VQA, making their capability closer to human performance (top right).}
    \vspace{-1.5em}
\label{fig:teaser}
\end{figure}

 
However, it is inherently challenging to construct training data for OOV reasoning.
Manual creation requires extensive human effort to design VQAs, considering both visible cues and unseen context while constructing plausible, discriminative and difficulty-balanced options.
Recent trends in dataset curation leverage fully- or semi-agentic pipelines to reduce manual labor~\cite{wang2024measuring,zhang2024provision,yang2025effective,dongfang2025multimodal}.
Yet, directly cropping subviews from ordinary images introduces new complications. 
Without access to full scene context, the question design process may lead to factual inconsistencies or hallucinated content, while the limited FoV and view cropping further restrict the diversity and quality of the generated questions.
To work around these limitations, we propose using panoramic imagery as the foundation for dataset construction, as panoramas naturally provide full $180^\circ\!\times 360^\circ$ views of the target environments, enabling flexible view selection while preserving global scene awareness.


This work presents the first study on OOV understanding in a single-view setting and we propose \textbf{\ourmethodnospace}, an automatic four-stage pipeline for generating high-quality multi-choice VQA data from panoramic imagery.
\ourmethod leverages the full-scene coverage of panoramas and the strong vision-language reasoning ability to synthesize scene-grounded and context-rich VQAs.
Specifically, we build a data preprocessor for panorama filtering, a visual analyzer for semantic and spatial annotation, a proposal generator for constructing VQAs, and a proposal refiner for quality assurance and augmentation. 
Together, these four stages form the \ourmethod pipeline, which is streamlined, efficient, and scalable.


Excitingly, the \ourmethod pipeline enables an automatic creation of \ourdatasetnospace, a large-scale synthetic data built upon a collection of diverse real-world panoramas. 
To support principled evaluation, we further develop \ourbenchnospace, a manually-verified benchmark curated from high-quality subsets of the panoramas, explicitly excluding samples used in training.
In addition to verifying semantic correctness, we perform difficulty control via a description-based test to maintain a balanced level of reasoning complexity.
We further introduce comprehensive evaluation metrics on OOV understanding, jointly assessing choice accuracy and rationale correctness. This design serves as a verification mechanism to distinguish genuine reasoning from correct guesses.
As Figure~\ref{fig:teaser} shows, we organize our OOV questions in two categories:
(i) contextual questions assess whether objects, attributes, or conditions are plausible beyond the visible region, and
(ii) directional questions emphasize spatial continuity under explicit changes in camera view direction.
These components establish a structured framework for evaluating the OOV reasoning ability of MLLMs.


Experimental results reveal that current MLLMs exhibit limited performance on OOV VQAs, compared with human participants on the same set of test data; see Figure~\ref{fig:teaser} (top right vs.~bottom right).
Leveraging \ourdataset for supervised fine-tuning, multiple MLLMs show consistent and notable improvements, underscoring the effectiveness of our synthetic data in enhancing OOV performance.
Overall, this work lays the foundation for advancing visual understanding beyond the visible FoV.
Main contributions are as follows:
\begin{itemize}
    \item We introduce a new MLLM understanding task, out-of-view (OOV) understanding, and conduct the first systematic study by formulating the OOV understanding task as a single-view multi-choice VQA problem.
    \item We design \ourmethodnospace, a four-stage pipeline to automatically generate a large volume of multi-choice VQA problems, leveraging panoramic imagery to enable scene-grounded and context-rich VQA synthesis.
    \item We create \ourbenchnospace, a benchmark with both contextual and directional question types, to jointly evaluate choice and rationale OOV accuracy.
    \item We construct \ourdatasetnospace, a synthetic dataset of over 158k OOV VQAs, spanning diverse real-world scenarios and consistently boosting the OOV performance of various existing MLLMs. 
\end{itemize}

\section{Related Work}
\label{sec:related-work}

\noindent\textbf{VQA benchmarks.}
Early work establishes benchmarks for visual question answering on natural images,~\eg, VQA~\cite{Antol_2015_ICCV} and VQA v2~\cite{Goyal_2017_CVPR}, followed by datasets that emphasize compositional reasoning,~\eg, GQA~\cite{hudson2019gqa}, and knowledge-intensive QA,~\eg, OK-VQA~\cite{Marino_2019_CVPR}, VCR~\cite{zellers2019recognition}, and A-OKVQA~\cite{schwenk2022okvqa}. Subsequent efforts expand the scope to text-heavy scenarios and charts~\cite{singh2019towards,masry2022chartqa,liu2024ocrbench,wang2024charxiv}, mathematical reasoning in visual contexts~\cite{lu2023mathvista,wang2024measuring,xiao2024logicvista}, and long or interleaved video understanding~\cite{wu2024longvideobench,fu2025video,zhu2025vau}.
In parallel, comprehensive evaluation suites are introduced to probe broad multimodal capabilities and generalization,~\eg, MM-Vet~\cite{yu2023mm}, SEED-Bench~\cite{li2024seed}, MMBench~\cite{liu2024mmbench}, MMMU~\cite{yue2024mmmu}, GAIA~\cite{mialon2024gaia}, MMT-Bench~\cite{ying2024mmt}, and MME-RealWorld~\cite{zhang2024mme}. 
Recent studies also begin to examine spatial memory, camera-centric understanding, and multi-view reasoning~\cite{yang2025thinking,liao2025thinking,lin2025towards,gholami2025spatial,dongfang2025multimodal,ke2025explain}.

Despite this rapid progress across domains and scales, OOV-related evaluation remains limited. 
Existing benchmarks assess reasoning with visual evidence confined predominantly to the visible frame.
They rarely provide protocols or supervision signals to encourage the models to infer scene elements beyond the captured view.
A recent benchmark~\cite{yin2025spatial} studies spatial mental modeling under multi-view observations, where models establish cross-view correspondence and recover spatial layouts from alternative views.
In contrast, our work focuses on single-view OOV understanding, where unseen regions are inherently unobservable, requiring models to reason under uncertainty over multiple hypotheses in a contextually and spatially conditioned manner.

\noindent\textbf{Out-of-view vs.~out-of-knowledge.}
Knowledge-centric VQA benchmarks~\cite{wang2017fvqa,shahMYP19,zellers2019recognition,Marino_2019_CVPR,schwenk2022okvqa} emphasize answering questions that require external knowledge, often through reasoning or retrieval over commonsense, encyclopedic facts, or structured knowledge sources. 
Such settings primarily evaluate the model’s ability beyond shallow visual recognition by inferring or accessing non-visual knowledge.
While OOV reasoning also involves commonsense knowledge, it introduces a new dimension of difficulty: different questions require different depths of visual understanding to effectively support commonsense reasoning.
Rather than relying solely on standalone factual knowledge, the model must integrate sufficient visible cues to infer the content that lies beyond the observed view. 
To this end, our benchmark complements the existing efforts by enabling quantitative evaluation of OOV understanding without reducing the question to pure knowledge retrieval or textual reasoning.

\noindent\textbf{Agentic data synthesis.}
Recent efforts in synthetic dataset creation increasingly adopt fully or semi-agentic workflows to reduce manual annotation cost and to enhance scalability. This approach combines powerful models (\eg, GPT‑4o~\cite{hurst2024gpt}), auxiliary tools (\eg, scene graph generators or spatial reasoning modules), and multi-stage coordination among generation, verification, and refinement agents~\cite{liu2023visual,zhang2024provision,ying2024mmt}.
For instance, large language models are orchestrated not just to generate data but also to critique, filter, or augment data, aiming to yield higher-fidelity synthetic supervision than naive prompt-based generation~\cite{madaan2023self,yang2025effective,fu2025video}.

Despite these advances, fully automatic pipelines often face notable challenges (\eg, vulnerable to factual drift and hallucination), especially in the absence of ground-truth context and the difficulty of maintaining spatial validity in the visual data. 
These limitations motivate us to leverage panoramic imagery to build a four-stage generative pipeline, grounding the question design in full 360$^\circ$ views and integrating quality control at necessary stages.

\begin{figure*}[t]
    \centering
    \includegraphics[trim=0 5 0 4, clip, width=\linewidth]{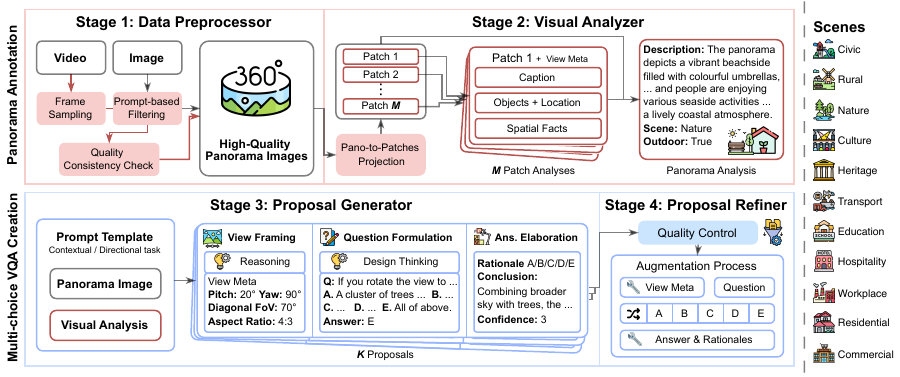}
    \caption{
    Overview of the OpenView pipeline.
    \textbf{Top: Panorama Annotation} (Section~\ref{sec:pano_annotation}) includes Stage~1, which collects, samples, and filters panoramic images, and Stage~2, which produces a spatial-grounded caption per local patch and a comprehensive summary per panorama.
    \textbf{Bottom: Multi-choice VQA Creation} (Section~\ref{sec:mcvqa_generation}) includes Stage~3, which generates multi-choice VQA proposals via view framing, question formulation, and answer elaboration for contextual and directional OOV tasks, and Stage~4, which improves the synthesis quality through format refinement, confidence-based filtering, and an augmentation to shuffle options and jitter views.
    \textbf{Right:} 
    Our generated VQAs cover diverse scene categories, both functional and environmental.
    }
    \label{fig:generation_pipeline}
\end{figure*}

\section{\ourmethod for Dataset Synthesis}
\label{sec:methods}

\noindent\textbf{Motivation of VQA:}
Out-of-view (OOV) reasoning is fundamentally underdetermined, as multiple hypotheses about unseen regions may be consistent with the observed image. Rather than enforcing a single deterministic ground truth or attempting to recover a full distribution over scene completions, we cast OOV understanding as comparative plausibility reasoning among a small set of candidate hypotheses, instantiated as a multi-choice VQA task that provides a tractable and automatically-evaluable proxy in line with the long-standing practice of reducing inherently uncertain QA to plausibility comparison among candidates~\cite{roemmele2011choice,bisk2020piqa,zellers2019hellaswag}.

\noindent\textbf{Pipeline Overview:} Figure~\ref{fig:generation_pipeline} shows our four-stage pipeline for automatic multi-choice VQA generation.
Throughout the pipeline, we employ Qwen2.5-VL-72B~\cite{Qwen2.5-VL} as the assistant model, leveraging its strong visual-language reasoning capability to support all four stages.

\subsection{Panorama Annotation}
\label{sec:pano_annotation}
\noindent\textbf{Source Data.}
We collected 5,289 panorama images and 3,034 panorama videos from five public datasets~\cite{mapillary_metropolis_dataset,Matterport3D,huang2024360loc,chen2024360+,wallingford2024image}, carefully considering scene coverage and image resolution.
We categorize the scenes into 11 major types; see Figure~\ref{fig:generation_pipeline} (right), which highlights the functional and environmental diversity of the scenes in the panorama collection. 
Further information on the panorama sources, sampling strategies, and scene classification taxonomy is provided in the Appendix.

\noindent\textbf{Stage 1: Data Preprocessor.}
\label{sec:stage1}
Although the collected panoramas have broad scene coverage, a substantial portion of them exhibit various quality issues and lack informativeness,~\eg, invalid equirectangular projections, visual artifacts, foreground obstructions, motion blur, and synthetic contents. 
Filtering out such cases improves the overall scene fidelity and realism.
To effectively identify defective samples, we employ prompt-based filtering by leveraging the assistant model to assess visual quality on various aspects.
For video filtering, we further perform a quality consistency check to eliminate video sources that contain multiple invalid frames, as such cases often indicate unreliable remaining frames.

\noindent\textbf{Stage 2: Visual Analyzer.}
\label{sec:stage2}
Accurate and structured visual annotation is crucial for VQA design~\cite{li2023evaluating,ying2024mmt,lin2025towards,gholami2025spatial,dongfang2025multimodal}, as annotations provide rich semantic grounding.
Yet, panoramic imagery poses unique challenges, due to the geometric distortion introduced by the equirectangular projection.
To this end, we divide each panorama into 12 perspective-projected views to obtain a set of image patches per panorama.
We then perform a spatial-focused analysis on each image patch using the assistant model and subsequently condense the patches into a concise text that highlights the key characteristics of the scene.
This visual analysis stage builds a global semantic representation, capturing the spatial relationships among objects and the overall scene description.
Section~\ref{sec:ablation_visual_analyzer} demonstrates the benefits of this stage for VQA creation.

\subsection{Multi-choice VQA Creation}
\label{sec:mcvqa_generation}

\noindent\textbf{Stage 3: Proposal Generator.}
\label{sec:stage3}

Following human design logic, we construct VQA proposals, by leveraging the reasoning capability of the assistant model, conditioned on full panoramic images and the visual analysis results from Stage 2.
Overall, Stage 3 has three coherent steps:
(i) {\em view framing\/} selects insightful perspective views with zero roll, oriented approximately perpendicular to the ground plane, to simulate natural camera poses of real-world photos;
(ii) {\em question formulation\/} creates an OOV question based on each selected view and the surrounding visual contents in the associated panorama, with five multi-choice options: 
four options correspond to concrete candidates, whereas the fifth serves as a compositional choice to accommodate cases where multiple ambiguous candidates remain plausible or none is fully correct, ~\eg, \emph{``Both A and C''} or \emph{``None of the above''}.
Options are prompted to be concise, mutually exclusive, and plausible, with negative choices produced either by contradicting facts or by introducing less plausible alternatives given visual evidence;
and (iii) {\em answer elaboration\/} provides concise rationale for each option, as well as a confidence score of the generated proposal in range 1 to 3.
Notably, these rationales justify the correct answer and offer the reasoning behind the incorrect options, guiding the MLLMs to learn the underlying visual-language alignment and relation, capturing the relative plausibility of competing OOV hypotheses, rather than focusing solely on the option selection.
At each step, the assistant model begins by articulating a thinking trace to enhance the interpretability of the creation process and to improve the robustness of the outputs.

\noindent\textbf{Stage 4: Proposal Refiner.}
\label{sec:stage4}
While the proposal generator significantly reduces human effort in VQA design, we observe that the outputs may contain poorly formulated questions, trivial questions, biased option ordering, and unbalanced answer distributions, thereby affecting the robustness of the synthetic data.

To retain the proposals with full confidence and correct minor formatting errors, we design the proposal refiner, as a quality control.
In addition, we augment each retained proposal through option shuffling and slight view jittering to further improve the data diversity and reliability.
Note that option shuffling means randomizing the options in the generated VQAs, reducing positional bias and preventing the models from memorizing option patterns.
On the other hand, we slightly jitter the view,~\ie, the perspective-projected image frame from the associated panorama, enhancing the visual diversity without altering the VQA semantics, while maintaining spatial fidelity.
Please refer to an ablation study in  Section~\ref{sec:ablation_proposal_refiner} that studies the effectiveness of the proposal refiner.

\begin{figure}[t]
    \centering
    \begin{subfigure}[h]{0.428\linewidth}
        \centering
        \includegraphics[trim=65 65 65 58, clip, width=\linewidth]{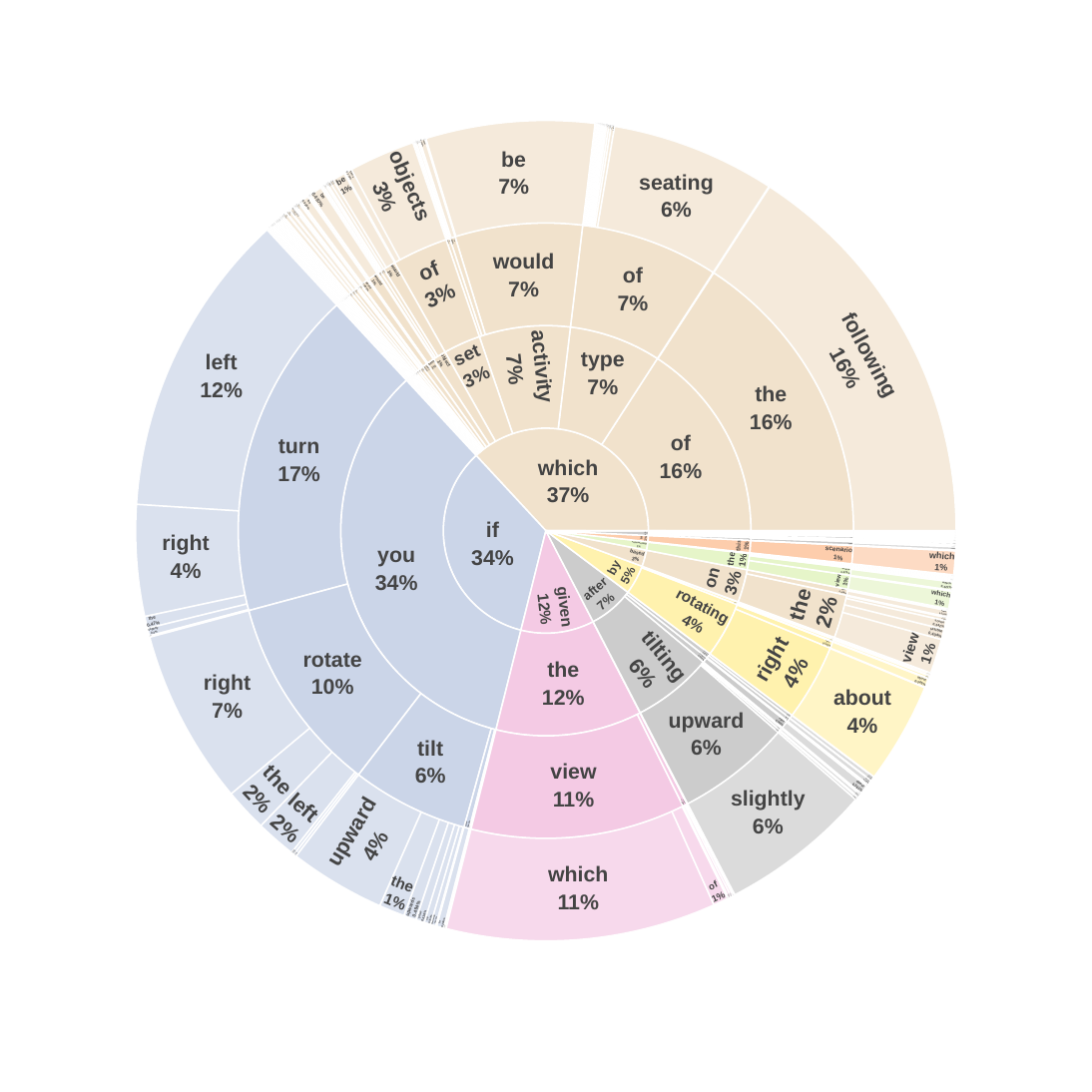}
        \caption{Distribution of the first four words of the questions}
        \label{fig:question_sunburst}
    \end{subfigure}
    \hfill
    \begin{subfigure}[h]{0.53\linewidth}
        \centering
        \begin{subfigure}[t]{0.49\linewidth}
            \centering
            \includegraphics[width=\linewidth]{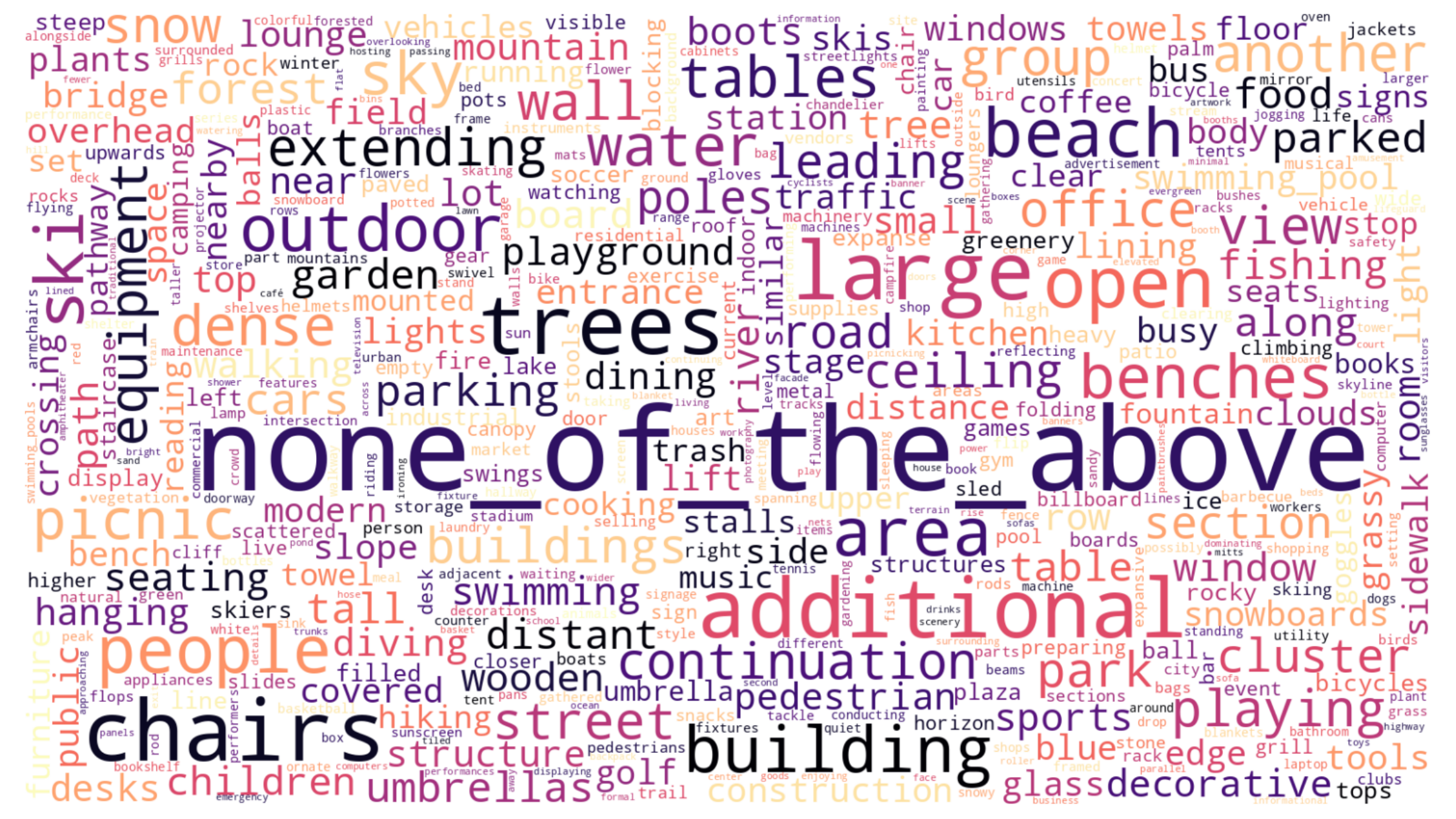}
            \caption{Word cloud of options}
            \label{fig:option_wordcloud}
        \end{subfigure}
        \hfill
        \begin{subfigure}[t]{0.49\linewidth}
            \centering
            \includegraphics[width=\linewidth]{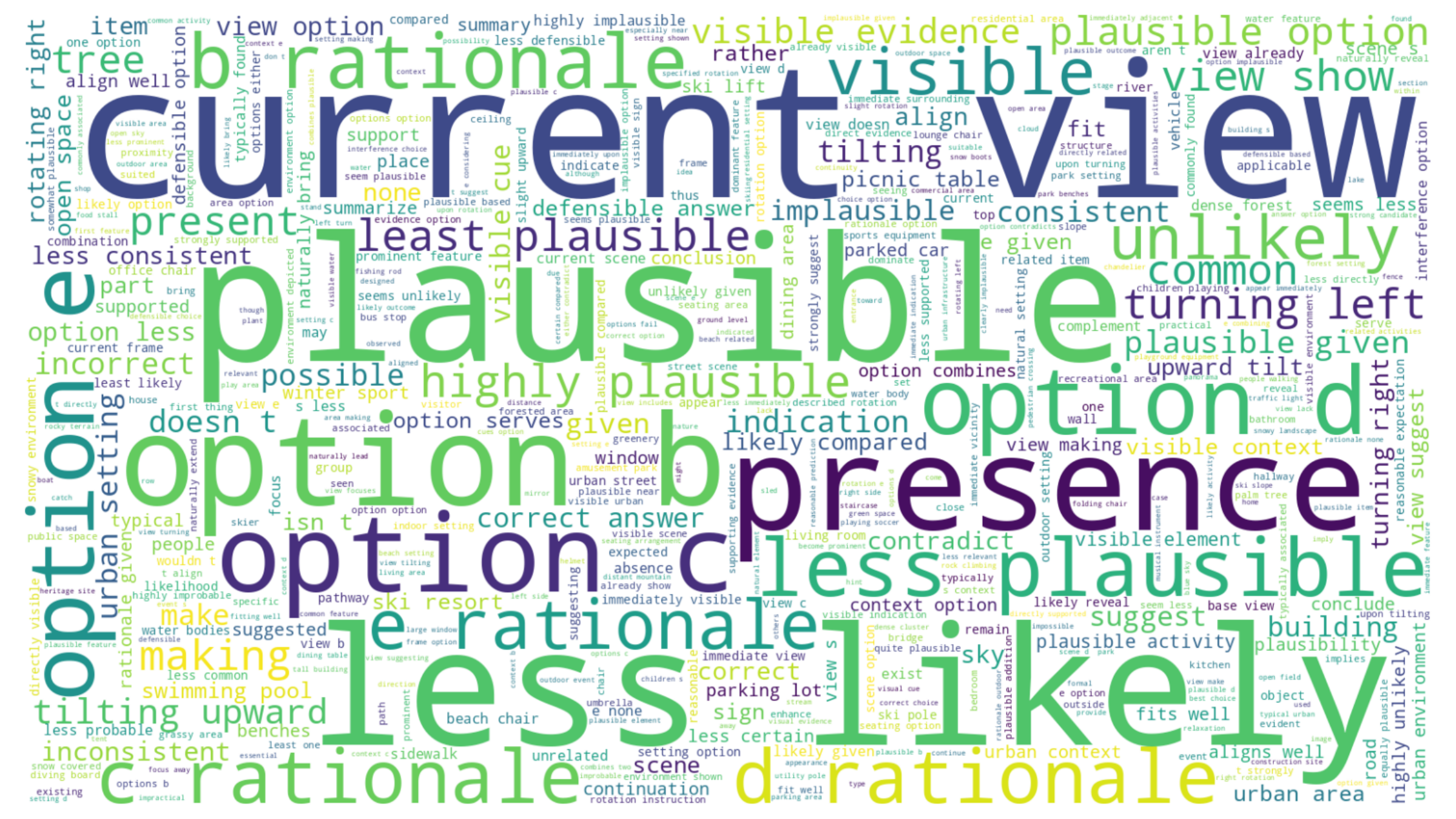}
            \caption{Word cloud of rationales}
            \label{fig:rationale_wordcloud}
        \end{subfigure}

        \includegraphics[width=0.916\linewidth]{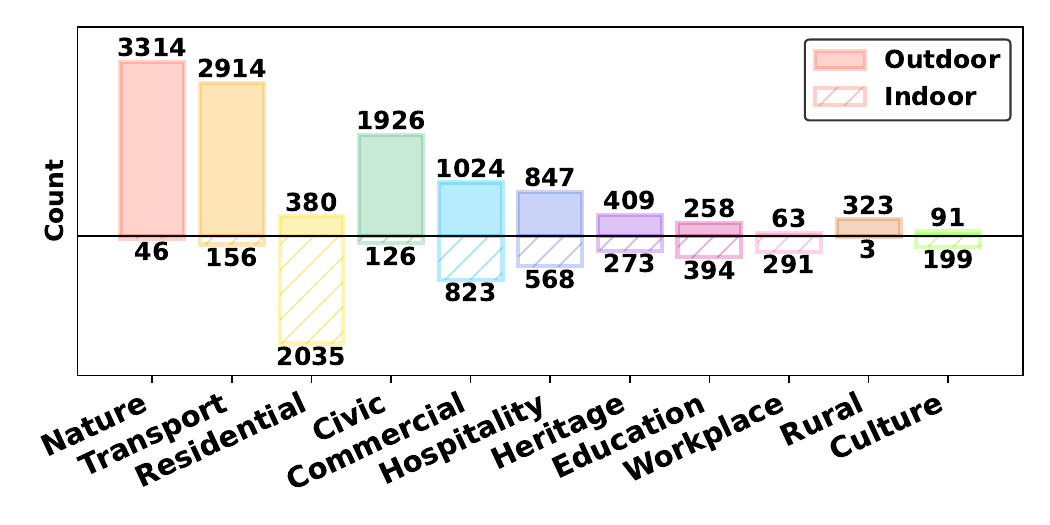}
        \caption{Distribution of scene categories across indoor and outdoor environments}
        \label{fig:label_dist}
    \end{subfigure}
    \caption{\textbf{An analysis on \ourdatasetnospace.} 
    Question patterns (a) exhibit high linguistic consistency, which guides the model to focus on OOV understanding.
    The option word cloud (b) reveals diverse object and scene terms, whereas the rationale word cloud (c) highlights recurring reasoning phrases commonly-used across questions.
    Scene category distribution (d) across indoor and outdoor 
    environments, illustrating the broad coverage of real-world scenarios represented in the collection.}
    \label{fig:dataset_analysis}
    \vspace{-1.5em}
\end{figure}

\subsection{Prompt Design and \ourdataset analysis}
\label{sec:dataset_analysis}

For each stage in \ourmethodnospace, we carefully design and iteratively refine the prompts through extensive qualitative inspection to enhance the quality of the results.
Details on the prompt templates and output examples are given in the Appendix.

We employ the \ourmethod pipeline to generate \ourdatasetnospace, which contains over 158k realistic multi-choice OOV VQAs from 16k filtered panoramas, spanning a broad range of real-world scenarios (see Figure~\ref{fig:label_dist}).
The question distribution (see Figure~\ref{fig:question_sunburst}) based on the first four words in the questions illustrates how the contextual and directional questions are formulated. 
Contextual questions typically begin with phrases such as \emph{``which of the following ...},'' \emph{``which type of ...},'' and \emph{``given the view, which ...},'' representing queries that target global scene semantics, whereas directional questions often start with expressions such as \emph{``if you turn ...},'' \emph{``after tilting upward ...},'' and \emph{``by rotating right ...},'' emphasizing visual reasoning under different camera view changes.

Further, we study the lexical composition through word cloud visualizations of both options and rationales (see Figures~\ref{fig:option_wordcloud} and \ref{fig:rationale_wordcloud}), overviewing the vocabulary diversity and reasoning expressions in the dataset, suggesting a broad range of lexical variety in scene-related and reasoning terms, and indicating the semantic diversity of the dataset.

Overall, the resulting synthetic dataset, \ourdatasetnospace, attains high coverage, visual richness, and structural coherence, providing a strong foundation for OOV understanding and model guidance.
Excitingly, as demonstrated in the experiments presented in Section~\ref{sec:main_results}, fine-tuning various MLLMs on \ourdataset consistently leads to improvements in their OOV understanding performance, manifesting the effectiveness of our data design.

\section{\ourbench}
\label{sec:bench}
Building on the \ourmethod pipeline, we construct \ourbenchnospace, a manually-verified benchmark with a holistic evaluation protocol that isolates single-view OOV understanding as a distinct capability beyond in-view tasks.
In total, the benchmark comprises 1,155 high-quality OOV VQAs with balanced task difficulty, scene coverage and answer distribution. Detailed benchmark statistics are provided in the Appendix.

\noindent\textbf{Benchmark Construction.}
We select seven representative panoramic images for each of the 11 scene categories, captured from distinct locations, and generate initial question proposals using the visual analyzer and proposal generator.
We then conduct multi-round human verification with GPT-4o~\cite{hurst2024gpt} as the assistant model, following recent LLM-assisted evaluation practices~\cite{madaan2023self,wang2024measuring}.
Annotators revise view framing, questions, options, and rationales, while removing trivial or hallucinated cases.
We further apply a description-based control test by replacing the image with assistant-generated object descriptions, and selectively revise questions that remain answerable under this setting, as they are considered to rely primarily on object co-occurrence statistics, thereby maintaining a balanced difficulty.
Finally, we augment the verified proposals, balance the answer distribution, and conduct cross-review for factual and logical coherence, resulting in 1,155 proposals.
More details on the construction and the user study are provided in the Appendix.

\noindent\textbf{Evaluation Metrics.}
\label{sec:evaluation_metrics}
For OOV reasoning, an MLLM is expected to not only choose the right answer but also provide a coherent and logical explanation for both chosen and rejected options.
This motivates us to design an evaluation protocol that has two parts:
(i) choice accuracy, which measures the proportion of correctly-selected options among all the questions, and  
(ii) rationale accuracy, which measures the reliability of the model in justifying the correctly-chosen answers through reasonable option-wise explanations.  

Following recent evaluation practices~\cite{yu2023mm,Guan_2024_CVPR,wang2024charxiv}, we adopt the LLM-as-the-judge paradigm to assess the correctness of the predicted response with the human revised rationales.
We employ GPT5-mini~\cite{openai2025gpt5systemcard} as the LLM judge and report its inter-annotator agreement analysis in the Appendix.
The choice accuracy is evaluated by exact matching with regex, whereas LLM judges the rationale by returning a binary correctness label.
Finally, the joint accuracy metric measures the proportion of questions for which both the selected answers and its rationales are correct.

\section{Experiments}
\label{sec:experiments}

\textbf{Setting.}
Five proprietary (Gemini-3-flash~\cite{deepmind2025gemini3}, GPT-4o~\cite{hurst2024gpt}, GPT-4o mini~\cite{openai2024gpt4omini}, GPT-5~\cite{openai2025gpt5systemcard} and GPT-5.4~\cite{openai2026gpt54}) and twelve recent open-source MLLMs 
(GLM-4.1V-9B-Base, GLM-4.1V-9B-Thinking~\cite{vteam2025glm45vglm41vthinkingversatilemultimodal}, 
InternVL3.5-8B~\cite{wang2025internvl3_5}, 
LLaVA-NeXT-Mistral-7B~\cite{liu2024llavanext},
MiniCPM-V 4.5~\cite{yu2025minicpm},
Ovis2.5-9B~\cite{lu2025ovis25technicalreport}, 
Qwen3-VL-8B-Instruct, Qwen3-VL-8B-Thinking~\cite{Qwen3-VL}, and 
Qwen2.5-VL series~\cite{Qwen2.5-VL}) are used.
%
We additionally evaluate a description-based setting, where visual inputs are replaced with object descriptions generated by GPT-4o.
Based on the OOV task formulation, we design a unified prompt template to instruct the model to first generate option-wise rationales, then provide the final answer in a structured tag.
For a fair comparison, we adopt the recommended inference configuration provided by each model.

\noindent\textbf{Implementation Details.}
To study the effectiveness of \ourdataset on 
empowering existing MLLMs to handle OOV tasks.
We use LoRA~\cite{hu2022lora} to fine-tune the MLLM models. 
Specifically, LLaVA-NeXT-Mistral-7B and InternVL3.5-8B are trained with a batch size of 8, a LoRA rank of 8, and an alpha value of 16,
whereas Qwen2.5-VL-7B and Qwen3-VL-8B are fine-tuned with a batch size of 128, a LoRA rank of 64, and an alpha value of 64.
All models are trained for one epoch with the vision tower frozen and a learning rate of $1\times10^{-4}$, using two NVIDIA A100 80GB GPUs.

\subsection{Main Results}
\label{sec:main_results}

\textbf{Results on \ourbenchnospace.}
Table~\ref{tab:oov_bench_results} compares human performance with various proprietary and open-source MLLMs on \ourbenchnospace.
A substantial gap is evident between humans and MLLMs. While humans surpass 83\% choice accuracy across both task types, model accuracies still fall short of this level.
Across the MLLM models, answer accuracy is moderately strong, 
yet considering rationale evaluation reveals a lower joint accuracy.
This discrepancy is particularly evident in directional tasks.
Among the proprietary models, GPT-5.4 achieve the highest joint performance, largely attributed to its exceptional rationale reliability.
Although MiniCPM-V 4.5 and QwenVL series achieve the highest performance in open-source models, the gap persists compared with leading proprietary models and human-level understanding.
More detailed examples are provided in the Appendix.

\begin{table}[t]
\centering
\caption{
\textbf{Evaluation on \ourbenchnospace.}
Results are reported as percentages for choice, rationale, and joint accuracy across contextual and directional tasks.
Joint accuracy columns are \tinycolorbox{gray!10}{shaded} as the primary comparison metric, with models ranked by the overall score.
The best results among all models and within open-source models are highlighted in \tinycolorbox{red!35}{red} and \tinycolorbox{red!10}{light red}, respectively.
$^\dagger$indicates description-based test setting;
$^\ddagger$indicates that thinking mode is enabled.}
\label{tab:oov_bench_results}
\resizebox{0.9\textwidth}{!}{
\begin{tabular}{l|c|cc>{\columncolor{gray!10}}c cc>{\columncolor{gray!10}}c|cc>{\columncolor{gray!10}}c}
\toprule
\multirow{2}{*}{\textbf{Model}} & 
\multirow{2}{*}{\textbf{Rank}} &
\multicolumn{3}{c}{\textbf{Contextual}} &
\multicolumn{3}{c}{\textbf{Directional}} &
\multicolumn{3}{|c}{\textbf{Overall}} \\
\cmidrule(lr){3-5} \cmidrule(lr){6-8} \cmidrule(lr){9-11}
 & & Choice & Rationale & Joint 
 & Choice & Rationale & Joint
 & Choice & Rationale & Joint \\
\midrule
\quad Human-level & -- & \textbf{83.02} & - & - & \textbf{89.58} & - & - & \textbf{86.00} & - & - \\
\midrule
\quad $^\dagger$GPT-4o~\cite{hurst2024gpt} & -- & 64.14 & - & - & 41.74 & - & - & 52.99 & - & - \\
\quad $^\dagger$GPT-5\cite{openai2025gpt5systemcard} & -- & 63.79 & - & - & 44.52 & - & - & 54.20 & - & - \\
\midrule
\multicolumn{11}{l}{\textit{Proprietary Models}} \\
\quad GPT-4o-mini~\cite{openai2024gpt4omini} & 9 & 63.10 & 93.99 & 59.31 & 48.35 & 87.05 & 42.09 & 55.76 & 90.99 & 50.74 \\
\quad GPT-4o~\cite{hurst2024gpt} & 6 & 71.72 & 83.89 & 60.17 & 55.83 & 89.41 & 49.92 & 63.81 & 86.30 & 55.07 \\
\quad Gemini-3-flash~\cite{deepmind2025gemini3} & 8 & \cellcolor{red!35}77.59 & 78.89 & 61.21 & \cellcolor{red!35}62.43 & 70.19 & 43.82 & \cellcolor{red!35}70.04 & 75.03 & 52.55\\
\quad GPT-5~\cite{openai2025gpt5systemcard} & 2 & 70.86 & 91.73 & 65.00 & 57.04 & 95.43 & 54.43 & 63.98 & 93.37 & 59.74 \\
\quad GPT-5.4~\cite{openai2026gpt54} & 1 & 70.52 & \cellcolor{red!35}99.02 & \cellcolor{red!35}69.83 & 59.30 & \cellcolor{red!35}96.48 & \cellcolor{red!35}57.21 & 64.94 & \cellcolor{red!35}97.87 & \cellcolor{red!35}63.56 \\
\midrule
\multicolumn{11}{l}{\textit{Open-source Models}} \\
\quad LLaVA-NeXT-Mistral-7B~\cite{liu2024llavanext} & 17 & 42.07 & 39.34 & 16.55 & 29.04 & 29.94 & 8.69 & 35.58 & 35.52 & 12.64 \\
\quad GLM-4.1V-9B-Base~\cite{vteam2025glm45vglm41vthinkingversatilemultimodal} & 14 & 56.38 & 87.46 & 49.31 & 42.78 & 79.10 & 33.84 & 49.61 & 83.89 & 41.62 \\
\quad GLM-4.1V-9B-Thinking~\cite{vteam2025glm45vglm41vthinkingversatilemultimodal}  & 11 & 52.41 & 92.11 & 48.27 & 45.04 & 90.73 & 40.86 & 48.74 & 91.47 & 44.58 \\
\quad InternVL3.5-8B~\cite{wang2025internvl3_5} & 12 & 56.90 & 90.00 & 51.21 & 45.39 & 79.69 & 36.17 & 51.17 & 85.45 & 43.72 \\
\quad $^\ddagger$InternVL3.5-8B~\cite{wang2025internvl3_5} & 15 & 49.48 & 82.58 & 40.86 & 42.26 & 85.60 & 36.17 & 45.89 & 83.96 & 38.53 \\
\quad Qwen3-VL-8B-Instruct~\cite{Qwen3-VL} & 10 & 58.10 & 94.96 & 55.17 & 48.35 & 90.65 & 43.83 & 53.25 & 93.01 & 49.53 \\
\quad Qwen3-VL-8B-Thinking~\cite{Qwen3-VL} & 7 & 67.41 & 91.05 & 61.38 & 49.57 & 91.23 & 45.22 & 58.53 & 91.12 & 53.33 \\
\quad Ovis2.5-9B~\cite{lu2025ovis25technicalreport} & 16 & 66.21 & 65.89 & 43.63 & 43.65 & 66.14 & 28.87 & 54.98 & 65.98 & 36.28 \\
\quad MiniCPM-V 4.5~\cite{yu2025minicpm} & 3 & \cellcolor{red!10}68.79 & 93.23 & 64.13 & \cellcolor{red!10}57.04 & 89.33 & \cellcolor{red!10}50.95 & \cellcolor{red!10}62.94 & 91.47 & \cellcolor{red!10}57.57 \\
\midrule
\multicolumn{11}{l}{\textit{Qwen2.5-VL series}~\cite{Qwen2.5-VL}} \\
\quad Qwen2.5-VL-3B-Instruct   & 18 & 57.93 & 13.99 & 8.10 & 44.70 & 12.06 & 5.39 & 51.34 & 13.15 & 6.75 \\
\quad Qwen2.5-VL-7B-Instruct   & 13 & 61.03 & 81.36 & 49.65 & 45.39 & 76.63 & 34.78 & 53.25 & 79.35 & 42.25 \\
\quad Qwen2.5-VL-32B-Instruct  & 4 & 66.38 & 95.58 & 63.45 & 53.22 & \cellcolor{red!10}94.12 & 50.09 & 59.83 & \cellcolor{red!10}94.93 & 56.80 \\
\quad Qwen2.5-VL-72B-Instruct  & 5 & 68.10 & \cellcolor{red!10}95.70 & \cellcolor{red!10}65.17 & 51.13 & 92.52 & 47.31 & 59.65 & 94.34 & 56.27 \\
\bottomrule
\end{tabular}
}
\end{table}

\begin{figure}[t]
    \centering
    \begin{minipage}[t]{0.48\linewidth}
        \centering
        \vspace{0pt}
        \includegraphics[width=\linewidth, trim=25 8 15 12, clip]{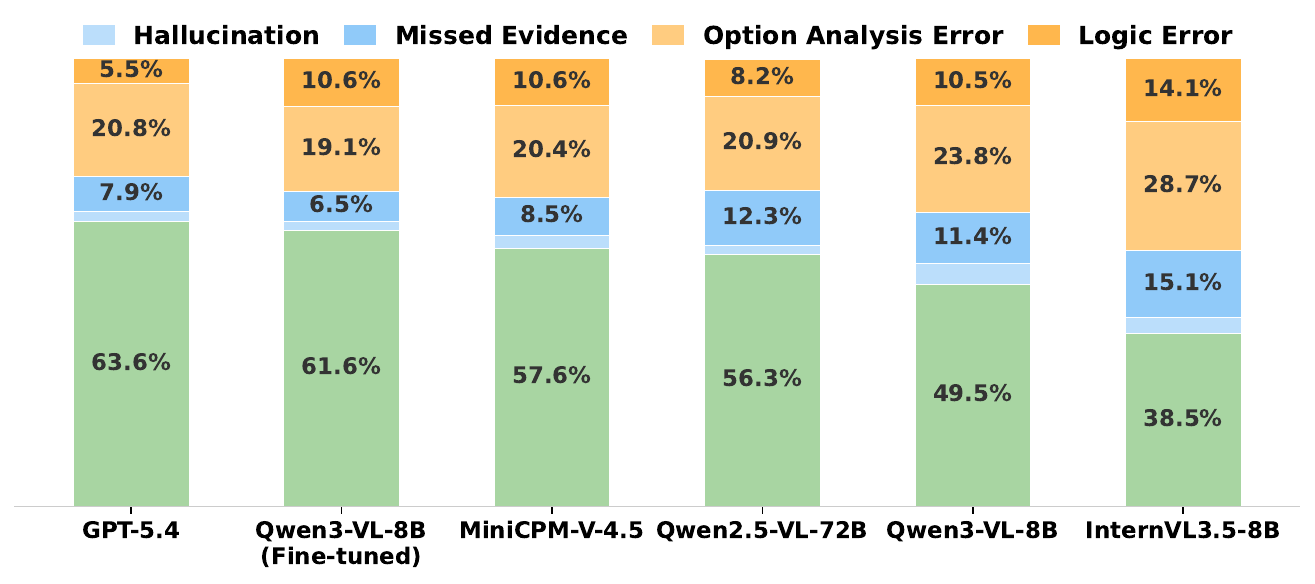}
    \end{minipage}
    \hfill
    \begin{minipage}[t]{0.51\linewidth}
        \centering
        \vspace{4pt}
        \includegraphics[width=\linewidth]{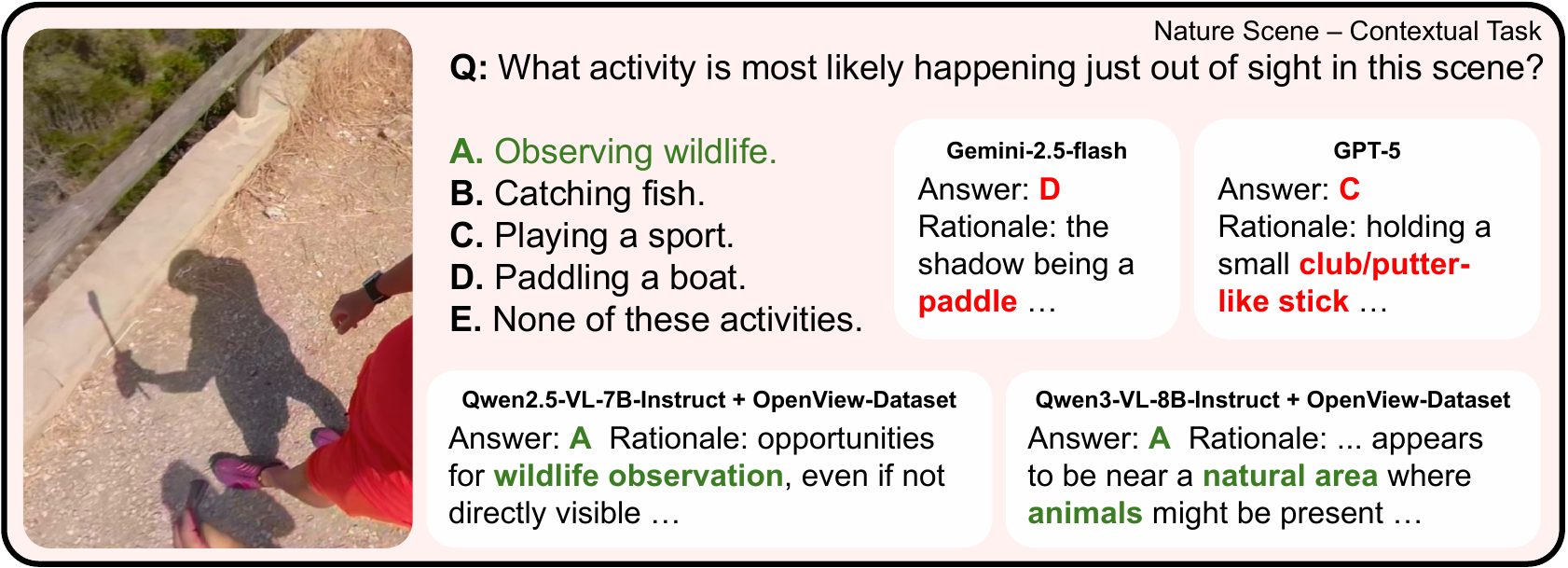}
    \end{minipage}

    \caption{
    \textbf{Left:} Breakdown of error cases across models.
    \textbf{Right:} An example test case.
    Baseline MLLMs misinterpret the scene, while fine-tuned models correctly infer the unseen activity.
    }
    \vspace{-1.5em}
    \label{fig:error_and_test_case}
\end{figure}

\noindent\textbf{What Makes OOV Difficult?}
To diagnose the performance gap, we analyze the responses of six representative models selected from different performance tiers (\ie, ranks 1, 3, 5, 10, 15, and the best fine-tuned variant). 
Following our definition of OOV understanding, we categorize the errors into two primary types upon examination, as shown in Figure~\ref{fig:error_and_test_case}~(left).

\textcolor{cyan!60}{Perception limitation:} 
These failures manifest as \emph{missed evidence}, where key objects or cues present in the frame are overlooked, and \emph{hallucination}, where models generate descriptions inconsistent with the actual image content.
The former leaves models without the grounding necessary to infer what lies beyond the visible region, while the latter introduces actively misleading evidence into the reasoning trace.
This is evidenced by the description-based setting, which lacks rich visual cues and consistently underperforms the original model (\eg, 53\% vs.\ 64\% choice acc. for GPT-4o in Table~\ref{tab:oov_bench_results}), confirming that adequate visual perception is a contributing factor for effective OOV reasoning.

\textcolor{orange!80}{Reasoning limitation:}
Reasoning errors constitute the dominant failure mode, accounting for approximately 70\% of all errors across models.
\emph{Option analysis error} reflects models' difficulty in correctly assessing each candidate against the visual context, including conflating in-frame absence with out-of-view implausibility, misjudging combined options and misweighting evidence, while \emph{logic error} manifest as self-contradictory reasoning in which final selections directly contradict the model's own preceding analysis.
This is further reflected in the consistently lower rationale accuracy on directional questions compared to contextual ones across most models, as these questions additionally require camera rotation reasoning. We provide breakdown by question type in Appendix.

Together, they underscore that OOV understanding uniquely demands the integration of accurate visual perception with structured reasoning in a contextually and spatially conditioned manner, as failures in either dimension compromise models' ability to reason about content beyond the captured frame.

\begin{table}[t]
\centering
\caption{
\textbf{OOV performance boosted by fine-tuning using \ourdatasetnospace.} 
Results are reported as percentages for choice, rationale, and joint accuracy across contextual and directional tasks.
The final column reports the relative improvement in overall joint accuracy.
After supervised fine-tuning, four open-source models show consistent joint improvements, with gains highlighted in \tinycolorbox{cyan!15}{blue} and minor rationale drops indicated in \tinycolorbox{gray!20}{gray}.
}
\label{tab:oov_finetune_results}
\resizebox{0.9\textwidth}{!}{
\begin{tabular}{lcccccc|cccc}
\toprule
\multirow{2}{*}{\textbf{Model}} &
\multicolumn{3}{c}{\textbf{Contextual}} &
\multicolumn{3}{c}{\textbf{Directional}} &
\multicolumn{4}{|c}{\textbf{Overall}} \\
\cmidrule(lr){2-4} \cmidrule(lr){5-7} \cmidrule(lr){8-11}
 & Choice & Rationale & Joint 
 & Choice & Rationale & Joint
 & Choice & Rationale & Joint & \% gain \\
\midrule
LLaVA-NeXT-Mistral-7B & 42.07 & 39.34 & 16.55 & 29.04 & 29.94 & 8.69 & 35.58 & 35.52 & 12.64 & \multirow{2}{*}{$\uparrow$ 332\%} \\
+ \ourdataset  & 71.55 & 90.84 & 65.00 & 49.57 & 89.12 & 44.18 & \cellcolor{cyan!15}60.61 & \cellcolor{cyan!15}90.14 & \cellcolor{cyan!15}\textbf{54.63} &  \\
\midrule
InternVL3.5-8B  & 56.90 & 90.00 & 51.21 & 45.39 & 79.69 & 36.17 & 51.17 & 85.45 & 43.72 & \multirow{2}{*}{$\uparrow$ 18\%} \\
+ \ourdataset  & 66.03 & 90.86 & 59.99 & 48.00 & 89.13 & 42.78 & \cellcolor{cyan!15}57.06 & \cellcolor{cyan!15}90.14 & \cellcolor{cyan!15}\textbf{51.43} &  \\
\midrule
Qwen2.5-VL-7B-Instruct  & 61.03 & 81.36 & 49.65 & 45.39 & 76.63 & 34.78 & 53.25 & 79.35 & 42.25 & \multirow{2}{*}{$\uparrow$ 39\%} \\
+ \ourdataset  & 72.76 & 93.36 & 67.93 & 56.00 & 87.89 & 49.22 & \cellcolor{cyan!15}64.42 & \cellcolor{cyan!15}90.99 & \cellcolor{cyan!15}\textbf{58.62} &  \\
\midrule
Qwen3-VL-8B-Instruct  & 58.10 & 94.96 & 55.17 & 48.35 & 90.65 & 43.83 & 53.25 & 93.01 & 49.53 & \multirow{2}{*}{$\uparrow$ 24\%} \\
+ \ourdataset  & 77.07 & 93.29 & 71.90 & 57.22 & 89.36 & 51.13 & \cellcolor{cyan!15}67.19 & \cellcolor{gray!20}91.62 & \cellcolor{cyan!15}\textbf{61.56} &  \\
\bottomrule
\end{tabular}
}
\vspace{-1.3em}
\end{table}

\noindent\textbf{\ourdataset Guides MLLMs with Consistent Improvements.}
We fine-tune four open-source MLLMs on \ourdataset and evaluate them on \ourbenchnospace.
As Table~\ref{tab:oov_finetune_results} reports, supervised fine-tuning yields substantial gains across all models.
Notably, the overall joint accuracy of LLaVA-Next-Mistral-7B increases from 12.64\% to 54.63\% accompanied by the rationale accuracy increase from 35.52\% to 90.14\%. 
Although the rationale accuracy decreases for Qwen3-VL-8B-Instruct, indicating reduced justification reliability for some correctly answered questions, its joint accuracy still improves, reflecting better alignment between choice predictions and their justifications.
After fine-tuning, Qwen3-VL-8B-Instruct achieves the strongest overall OOV performance among open-source models.
An example test case is visualized in Figure~\ref{fig:error_and_test_case}~(right).
These results demonstrate that \ourdataset effectively guides MLLMs towards better OOV understanding and strengthens their ability to reason in a structured and interpretable manner.

\noindent\textbf{Cross-paradigm Generalization to MindCube.}
\label{sec:cross-paradigm}
While our \ourdataset trains models on \emph{single-view, plausibility-based} OOV reasoning, MindCube~\cite{yin2025spatial} evaluates a distinct setting: \emph{multi-view spatial mental modeling}, where models recover 3D layouts under explicit camera trajectories (Rotation, Around, Among) and infer occluded structure from cross-view correspondence. 
As shown in Table~\ref{tab:mindcube}, fine-tuning with \ourdataset consistently improves performance across both models. 
For Qwen2.5-VL-7B-Instruct, the overall accuracy increases from 31.02\% to 37.73\%, with notable gains on the Among category. 
The improvement is more pronounced for Qwen3-VL-8B-Instruct, where the overall score rises from 29.23\% to 41.50\%, and the Among category improves from 29.48\% to 43.05\%. 
That training on \emph{single-view plausibility comparison} yields gains on \emph{multi-view 3D spatial reasoning} indicates that the visual reasoning skills induced by \ourdataset are transferable to settings beyond its training paradigm, rather than being narrowly tied to the OOV question format.

\begin{table}[t]
\centering
\begin{minipage}[t]{0.48\linewidth}
\centering
\captionof{table}{\textbf{Performance on the MindCube benchmark}. Fine-tuning on \ourdataset improves performance across all question types.}
\label{tab:mindcube}
\resizebox{\linewidth}{!}{
\begin{tabular}{c|c|ccc}
\toprule
\textbf{Model} & \textbf{Overall} & \textbf{Rotation} & \textbf{Among} & \textbf{Around} \\
\midrule
Qwen2.5-VL-7B-Instruct & 31.02 & 37.28 & 31.06 & 27.02 \\
+ \ourdataset & 37.73 & 39.50 & 38.65 & 27.72 \\
$\Delta$ & +6.71 & +2.22 & +7.59 & +0.70 \\
\midrule
Qwen3-VL-8B-Instruct & 29.23 & 35.25 & 29.48 & 23.27 \\
+ \ourdataset & 41.50 & 35.62 & 43.05 & 29.86 \\
$\Delta$ & +12.27 & +0.37 & +13.57 & +6.59 \\
\bottomrule
\end{tabular}}
\end{minipage}
\hfill
\begin{minipage}[t]{0.475\linewidth}
\centering
\captionof{table}{\textbf{Visual Analyzer (VA) and Proposal Refiner (PR).}
We report overall accuracies on contextual and directional tasks.
}
\label{tab:ablation_visual_analyzer_and_proposal_refiner}
\resizebox{\linewidth}{!}{
\begin{tabular}{lcc|ccc}
\toprule
\textbf{Model} & \textbf{VA} & \textbf{PR} &
\textbf{Choice} & \textbf{Rationale} & \textbf{Joint} \\
\midrule
\multirow{3}{*}{InternVL3.5-8B}
 & \xmark & \cmark & 55.58 & 88.01 & 48.92 \\
 & \cmark & \xmark & 54.81 & 87.99 & 48.23 \\
 & \cmark & \cmark & 54.98 & 89.45 & 49.18 \\
\midrule
\multirow{3}{*}{Qwen2.5-VL-7B-Instruct}
 & \xmark & \cmark & 62.08 & 88.15 & 54.72 \\
 & \cmark & \xmark & 61.13 & 85.69 & 52.38 \\
 & \cmark & \cmark & 64.87 & 89.30 & 57.93 \\
\bottomrule
\end{tabular}}
\end{minipage}
\vspace{-0.5em}
\end{table}

\subsection{Ablation Studies}
We conduct ablation studies to examine the effects of two major components in \ourmethod pipeline.
Specifically, we fine-tune InternVL3.5-8B and Qwen2.5-VL-7B-Instruct under different configurations using a subset of 5k panoramas and evaluate their performance on \ourbenchnospace. 

\phantomsection
\noindent\textbf{Role of Visual Analyzer.} 
\label{sec:ablation_visual_analyzer}
As shown in Table~\ref{tab:ablation_visual_analyzer_and_proposal_refiner}, fine-tuning on the synthetic dataset without the visual analysis stage consistently reduces performance for both models.
The visual analyzer handles equirectangular distortion by producing structured annotations from 12 perspective-projected views.
These annotations provide spatial evidence for reliable VQA generation and cover the full panorama view by view, allowing the proposal generator to use the entire $360^\circ$ scene.

\phantomsection
\noindent\textbf{Role of Proposal Refiner.}
\label{sec:ablation_proposal_refiner}
Table~\ref{tab:ablation_visual_analyzer_and_proposal_refiner} also shows that applying the proposal refiner, which performs quality control and augmentation process, leads to additional performance gains.
These refinements strengthen the robustness of the synthesized supervision and help models learn more effectively from the generated VQAs.
Additional ablation and application studies are provided in the Appendix.

\section{Conclusion}
\label{sec:conclusion}
We presented the first systematic study on single-view OOV understanding, challenging the MLLMs to reason beyond the visible image frame, and introduced \ourmethodnospace, a streamlined generation pipeline, to efficiently synthesize scene-grounded and context-rich OOV VQAs from panoramic imagery.
Leveraging this pipeline, we constructed \ourdatasetnospace, which provides high-quality synthetic supervision to fine-tune MLLMs, and \ourbenchnospace, a benchmark that enables us to assess both choice and rationale accuracies for contextual and directional tasks.
Experiment results show that training on \ourdataset yields consistent and substantial improvements for multiple MLLMs, enhancing not only their ability to select correct answers but also their capacity to produce logical and coherent rationales.
Overall, our results demonstrate that OOV understanding is learnable through \ourmethodnospace, marking an important step toward more advanced multimodal intelligence.

\noindent{\textbf{Limitations \& Future work.}}
Although our synthetic dataset provides effective supervision for improving OOV reasoning in several MLLMs, the construction of candidate options relies on prompt-based generation and confidence filtering, which may not always produce ideally balanced distractors.
Our pipeline currently uses a single-pass generator and does not perform iterative adversarial distractor filtering, as in HellaSwag~\cite{zellers2019hellaswag}.
Closing this gap, for example by re-scoring distractors against an evolving discriminator, is a promising direction for future work.

\begin{ack}
This research was supported by Monash eResearch capabilities, including M3 HPC.
We thank Zhuokun Chen, Yizhou Tian, Lantian Wei, Tangyuan Zhang, and Liyun Zhu for helpful discussions and assistance with data preparation.
The authors declare no competing interests.
\end{ack}

\bibliographystyle{plainnat}
\bibliography{main}


\appendix

\clearpage

\noindent Below is the outline of this supplement:
\begin{itemize}
    \item \textbf{Section}~\ref{supp:more_pipeline} provides additional details on the source data and the \ourmethod pipeline.
    \item \textbf{Section}~\ref{supp:more_dataset} presents further statistics, a human quality audit, computational costs, and visualizations of \ourdatasetnospace.
    \item \textbf{Section}~\ref{supp:more_benchmark} details the statistics and construction process of \ourbenchnospace, along with the user study and rationale evaluation agreement analysis.
    \item \textbf{Section}~\ref{supp:more_results} reports a more comprehensive description-based evaluation, question-rephrasing robustness, few-shot/CoT baselines, and transfer from spatial reasoning training, followed by failure-case and per-scene analyses, data-scaling ablations, and qualitative visualizations.
    \item \textbf{Section}~\ref{supp:application} demonstrates the image outpainting application using \ourdatasetnospace-tuned MLLMs.
\end{itemize}

\noindent\textbf{Broader Impacts.}
This work aims to enhance MLLMs with out-of-view (OOV) understanding, enabling them to reason about unseen regions beyond the visible field of view under partial observations.
It can potentially benefit applications that require scene understanding from limited visual input, such as image out-painting, world modeling and embodied navigation.
Our dataset and benchmark target generic environmental reasoning and do not involve identity recognition, personal profiling, or sensitive attribute inference. 
Incorrect predictions about unseen scene content could lead to misleading interpretations of the surrounding environment. Model outputs should therefore be treated as plausible inferences rather than verified observations.

\noindent\textbf{Data Release and Privacy.}
We release VQA annotations and projection metadata without redistributing raw panoramas or projected image crops.
Users should obtain images directly from the original dataset providers under their respective access conditions and terms of use, then preprocess them locally into question views using our code.

\section{\ourmethodnospace: More Details}
\label{supp:more_pipeline}

\subsection{Source Data Details}
\noindent\textbf{Panorama Sources.} 
Table~\ref{tab:dataset_sources} summarizes the panoramic images and videos collected from five public datasets:
(i) \textbf{Matterport3D}~\cite{Matterport3D} is a large-scale RGB-D dataset, containing 10,800 panoramas across 90 building-scale indoor environments, where each panorama is reconstructed by stitching its skybox images, during the collection process; 
(ii) \textbf{Mapillary Metropolis}~\cite{mapillary_metropolis_dataset} is a multimodal city-scale benchmark with 27,745 high-resolution panoramic street views; 
(iii) \textbf{360Loc}~\cite{huang2024360loc} provides panoramic videos (panoramic image sequences) recorded in four campus locations, including atrium, concourse, piazza, and hall areas; 
(iv) \textbf{360+x}~\cite{chen2024360+} covers 28 panoptic multi-modal scenes across 17 cities and includes 464 high-resolution panoramic videos; and
(v) \textbf{360-1M}~\cite{wallingford2024image} is a large-scale real-world dataset of over one million panoramic videos, from which we retain a high-quality subset of videos with resolutions above $4096\times2048$.

\begin{table}[h]
\centering
\caption{Details of source datasets for panoramic images and videos.}
\label{tab:dataset_sources}
\resizebox{0.6\linewidth}{!}{
\begin{tabular}{lcccc}
\toprule
\textbf{Dataset} & \textbf{\#~Images} & \textbf{\#~Videos} & \textbf{Sampling Strategy} \\
\midrule
Matterport3D~\cite{Matterport3D} & 2{,}295 & -- & All frames \\
Mapillary Metropolis~\cite{mapillary_metropolis_dataset} & 2{,}994 & -- & Random sample 10\% \\
360Loc~\cite{huang2024360loc} & -- & 14 & Frame step: 10 \\
360+x~\cite{chen2024360+} & -- & 232 & 5 frames per video \\
360-1M~\cite{wallingford2024image} & -- & 2{,}792 & 5 frames per video \\
\bottomrule
\end{tabular}
}
\end{table}

\noindent\textbf{Sampling Strategies.}
We employ dataset-specific sampling strategies tailored to the characteristics and conditions of each dataset.
For panoramic image sources, we include all panoramas available from Matterport3D, which covers various indoor building scenes, and include ten percent of the panoramas from Mapillary Metropolis via randomly sampling to balance the street-view contents in our collection.
For panoramic video sources, we sample 360Loc with a frame step of ten, as its videos are recorded while traversing the campus.
For 360+x and 360-1M, which contain diverse real-world scenarios, we uniformly select five frames from the middle segments of each video, effectively maintaining the temporal diversity and avoiding subtitle overlays or empty frames at the start and end of the videos.

\noindent\textbf{Scene Classification Taxonomy.} 
To facilitate a clear understanding of the scene distribution in our panorama collection, we cluster the panoramas into groups of similar functional and environmental characteristics.
Below, we summarize the 11 scene categories in our taxonomy, each accompanied by representative contents:
\begin{tcolorbox}[width=\linewidth, colframe=black, colback=red!6, boxsep=0mm, arc=2mm, left=1mm, right=2mm, top=2mm, bottom=2mm]
\begin{itemize}[leftmargin=12pt]
    \item \textbf{Civic}: Squares, parks, playgrounds, botanical gardens, zoos, and other community spaces.
    \item \textbf{Rural}: Agricultural and countryside areas such as farms, villages, and fields.
    \item \textbf{Nature}: Mountains, forests, beaches, rivers, and other wilderness settings.
    \item \textbf{Culture}: Museums, theaters, concert halls, and sports arenas.
    \item \textbf{Heritage}: Historic and religious sites including temples, monuments, and ruins.
    \item \textbf{Transport}: Roads, stations, airports, bus stops, ports, and parking areas.
    \item \textbf{Education}: Campuses, schools, libraries, and other learning environments.
    \item \textbf{Hospitality}: Hotels, resorts, recreation centers, and conference halls.
    \item \textbf{Workplace}: Offices, labs, hospitals, and institutional buildings.
    \item \textbf{Residential}: Homes, apartments, courtyards, and living spaces.
    \item \textbf{Commercial}: Shops, malls, markets, restaurants, and cafés.
\end{itemize}
\end{tcolorbox}

\subsection{Further Method Details}
In the following parts, we provide further details about our \ourmethod pipeline.

\noindent\textbf{Stage 2: Visual Analyzer.}
To prepare for a structured visual analysis, each panorama is divided into 12 perspective-projected views.
These views form a grid of three rows and four columns with slight overlap (for the top and bottom rows), due to the spherical geometry of the projection.
Each patch is defined by its view metadata, including intrinsic parameters (diagonal FoV and aspect ratio) and extrinsic parameters (yaw and pitch).

After that, we employ the assistant model to analyze each patch to generate (i) a short caption that describes the local content, (ii) a list of objects with their spatial locations represented by nine coarse directional regions (\eg, top-left, center, bottom-right), and (iii) the spatial relationships among the visible objects. 
We then assign a numerical identifier to each patch and also include the indices of its adjacent views in the metadata to provide additional contextual information.
These patches and their analyses are further jointly processed by the assistant model to generate a global summary to categorize the overall scene category and characterizes the outdoor and indoor states.

Overall, this process provides a visual analysis with fine-grained spatial annotations from the perspective-projected views, together with an overall scene description of each panorama.

\noindent\textbf{Stage 3: Proposal generator.}
Similar to Stage 2, where each perspective-projected view is defined by its view metadata, the assistant model frames the appropriate question view by specifying the intrinsic and extrinsic parameters. 
In this work, we set the roll angle to zero to keep each projected view perpendicular to the ground plane to simulate the natural image style and the associated camera pose.
We intentionally avoid introducing additional challenges involving complex camera pose reasoning. For example, asking about the appearance of a view after rotating to the right when the current view is captured with a 30$^\circ$ roll. 
This would mix the camera geometry with out-of-view (OOV) reasoning, which is not the focus of this study.

\noindent\textbf{Stage 4: Proposal Refiner.}
The quality control module corrects minor formatting issues and removes proposals with major structural errors or low confidence to improve the reliability of the overall synthesis. During the final stage of proposal generation, each candidate is assigned a confidence score from 1 (low) to 3 (high), and only those with a score of 3 are retained.

The augmentation module further introduces two strategies to increase the number of samples and to enhance the robustness of the dataset:
(i) An option shuffle strategy, in which we randomize the order of the options together with their corresponding rationale. 
This strategy aims to minimize positional bias in the synthetic data and prevent the model from learning the superficial option patterns rather than genuine OOV understanding during fine-tuning.
(ii) A view jittering strategy, in which we randomly rotate the viewing direction slightly within $\pm~3.6^\circ$ in yaw and pitch. 
This introduces additional view diversity while maintaining the semantic of the VQA instances.

\subsection{Prompt Templates}
Tables~\ref{tab:prompt_stage1_filter} to~\ref{tab:prompt_stage4} provide all the prompt templates created for the four stages of \ourmethodnospace.
Table~\ref{tab:prompt_stage1_filter} presents the prompt for filtering out defective panorama samples.
Tables~\ref{tab:prompt_stage2_caption} and~\ref{tab:prompt_stage2_summary} show the templates employed for visual analysis, including patch-wise analysis and overall summarization. 
Note that we represent perspective-projected view locations (\ie, the extrinsic parameters) using UV coordinates (u, v $\in [0,1]$), which can be directly converted into yaw and pitch. 
This representation maintains a direct correspondence to the image domain, avoiding the need to introduce camera-pose conventions during the model inference and making the spatial grounding more intuitive for the model.

For multi-choice VQA proposal generation, we form a structured prompting setup that includes a base instruction prompt (Tables~\ref{tab:prompt_stage3_part1}-\ref{tab:prompt_stage3_part3} in three parts), two task-specific few-shot prompts (Tables~\ref{tab:prompt_stage3_contextual} and~\ref{tab:prompt_stage3_directional}), and a user prompt (Table~\ref{tab:prompt_stage3_user}).
The instruction prompts define the core idea and general rules of OOV VQA design, whereas the task-specific prompts outline the specific objectives and provide example questions as part of the system prompts.
The user prompt provides the panorama image and its visual analysis, enabling the model to generate multi-choice VQA proposals for each panoramic image.
Moreover, Table~\ref{tab:prompt_stage4} shows the prompt for minor-format fixing in the Proposal Refiner.

Finally, Table~\ref{tab:prompt_inference} presents the unified inference prompt in \ourbenchnospace, and Table~\ref{tab:prompt_eval} provides the detailed instructions for the judge model (\ie, GPT5-mini) to evaluate the correctness of a model’s rationale against the ground truth.

\section{More Dataset Statistic and Visualizations}
\label{supp:more_dataset}

\subsection{More \ourdataset Statistics}
\begin{figure}[h]
    \centering
    \begin{subfigure}[t]{0.32\linewidth}
        \includegraphics[width=\linewidth]{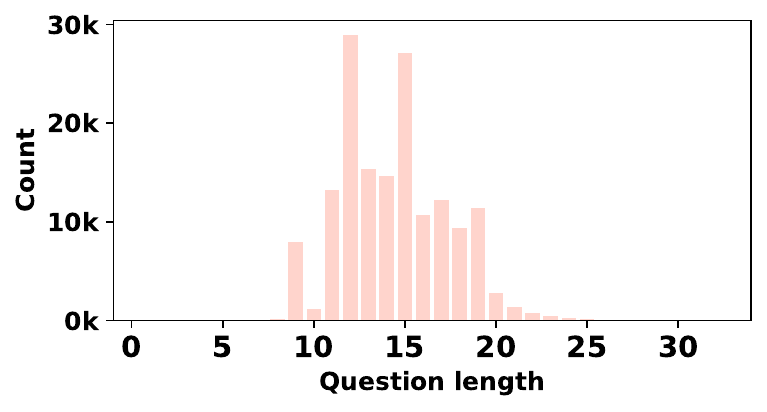}
        \caption{Distribution of the question length}
        \label{fig:question_len_dist}
    \end{subfigure}
    \begin{subfigure}[t]{0.32\linewidth}
        \includegraphics[trim=5 0 7 7, clip, width=\linewidth]{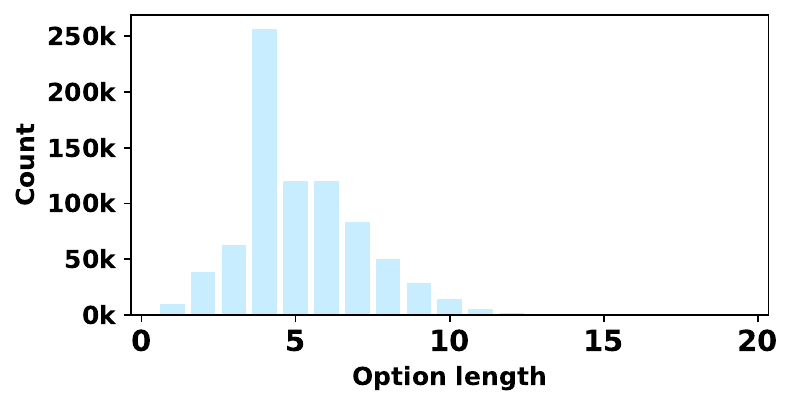}
        \caption{Distribution of the option length}
        \label{fig:option_len_dist}
    \end{subfigure}
    \begin{subfigure}[t]{0.315\linewidth}
        \includegraphics[trim=5 0 0 0, clip, width=\linewidth]{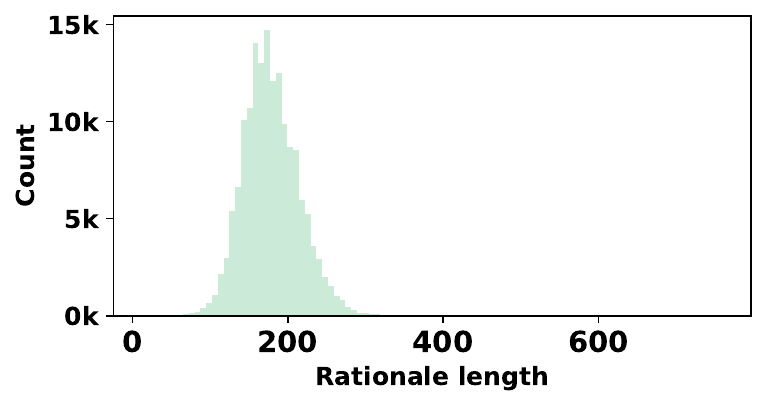}
        \caption{Distribution of the rationale length}
        \label{fig:rationale_len_dist}
    \end{subfigure}
    \caption{\textbf{Additional statistical analysis of \ourdatasetnospace.} Word length distributions are shown for questions, options, and rationales.}
    \label{fig:more_dataset_analysis}
\end{figure}

When constructing \ourdatasetnospace, the proposal generator produces three proposals per panorama to balance synthesis diversity with computational efficiency, ensuring that the overall inference cost of large-scale generation remains tractable.
Figure~\ref{fig:more_dataset_analysis} presents additional statistics on the length distributions of questions, options, and rationales in \ourdatasetnospace.

As shown in Figures~\ref{fig:question_len_dist} and~\ref{fig:option_len_dist}, both questions and options are intentionally concise, with question lengths centered around a mean of about 15 words and each option lengths most frequently observed at approximately four words.
In contrast, the rationale distribution in Figure~\ref{fig:rationale_len_dist} forms a clear and narrow peak around a mean of 190 words, indicating that each question is paired with a detailed and informative explanation.
These characteristics help to guide the models to connect the visual evidence with the structured reasoning and their relation with contextual knowledge, thereby improving their ability to perform reliable OOV understanding.

\subsection{Human Quality Audit}
\label{supp:dataset_audit}
We conduct a human audit of 110 VQA proposals sampled from the 158k training set, covering all 11 scene categories with ten proposals per category (five contextual and five directional).
Three annotators independently classify the principal defect of each proposal along four dimensions: weak distractors, ambiguous questions, incorrect answers, and flawed rationales.
A proposal is flagged only when at least two of the three annotators agree.

Under this criterion, 71.8\% of proposals pass all checks.
The remaining proposals are flagged for weak distractors (10.0\%), ambiguous questions (8.2\%), incorrect answers (5.5\%), or flawed rationales (4.5\%).
Inter-annotator agreement is high across the four dimensions, with Fleiss' $\kappa$ values of 0.929, 0.958, 0.869, and 0.925, respectively.

\subsection{Inference Computational Cost}
\label{sec:inference_cost}
We analyze the per-stage inference cost of the \ourmethod pipeline.
Table~\ref{tab:pipeline_cost} reports the computational costs.
Notably, we keep the proportion of image tokens large, ensuring the model receives rich visual context for producing accurate outputs.

\begin{table}[ht]
\centering
\caption{\textbf{Computational cost of one forward pass for each stage in \ourmethodnospace.}
We round the average token counts to integers and report the average runtime on two A100 80GB GPUs in seconds.}
\label{tab:pipeline_cost}
\resizebox{0.7\linewidth}{!}{
\begin{tabular}{lccccc}
\toprule
\textbf{Stage/Process} & \textbf{Avg. Input Tokens} & & \textbf{Avg. Output Tokens} & & \textbf{Time~(s)} \\
\midrule
S1/Prompt-based Filtering & 2{,}442 && 75 && 4.74 \\
S2/Patch Analysis & 1{,}205 && 121 && 1.21 \\
S2/Panorama Analysis & 5{,}278 && 101 && 4.65 \\
S3/Proposal Generation & 5{,}538 && 1{,}669 && 44.89 \\
S4/Format Refine & 459 && 99 && 2.12 \\
\bottomrule
\end{tabular}
}
\end{table}

\subsection{Visualizations of Proposals}
\begin{figure*}[ht]
    \centering
    \includegraphics[width=\linewidth]{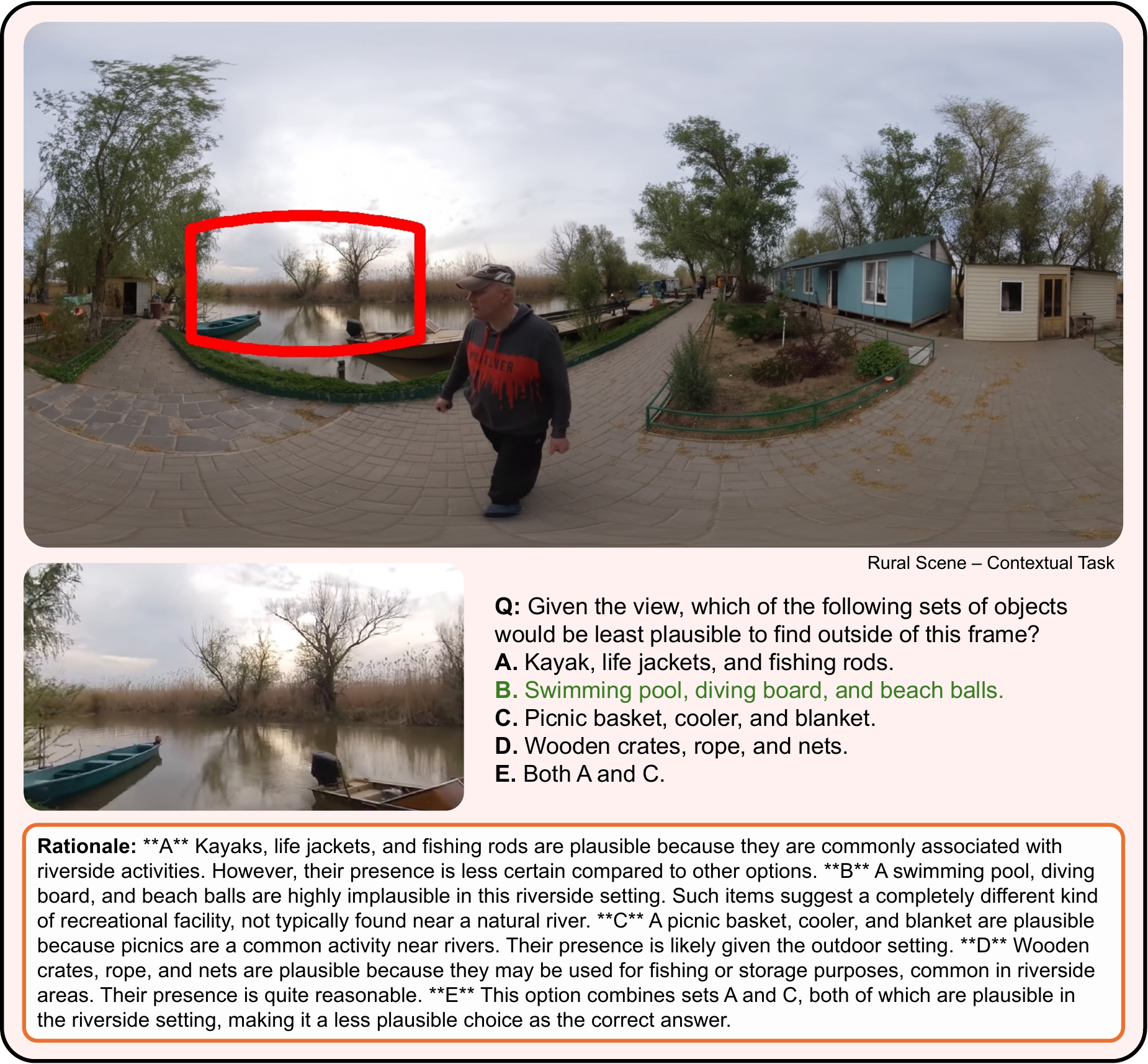}
    \caption{Example of a contextual question on a rural scene. The corresponding perspective projection is outlined in red and the correct answer is highlighted in green.}
    \label{fig:example_rural}
\end{figure*}

\begin{figure*}[ht]
    \centering
    \includegraphics[width=\linewidth]{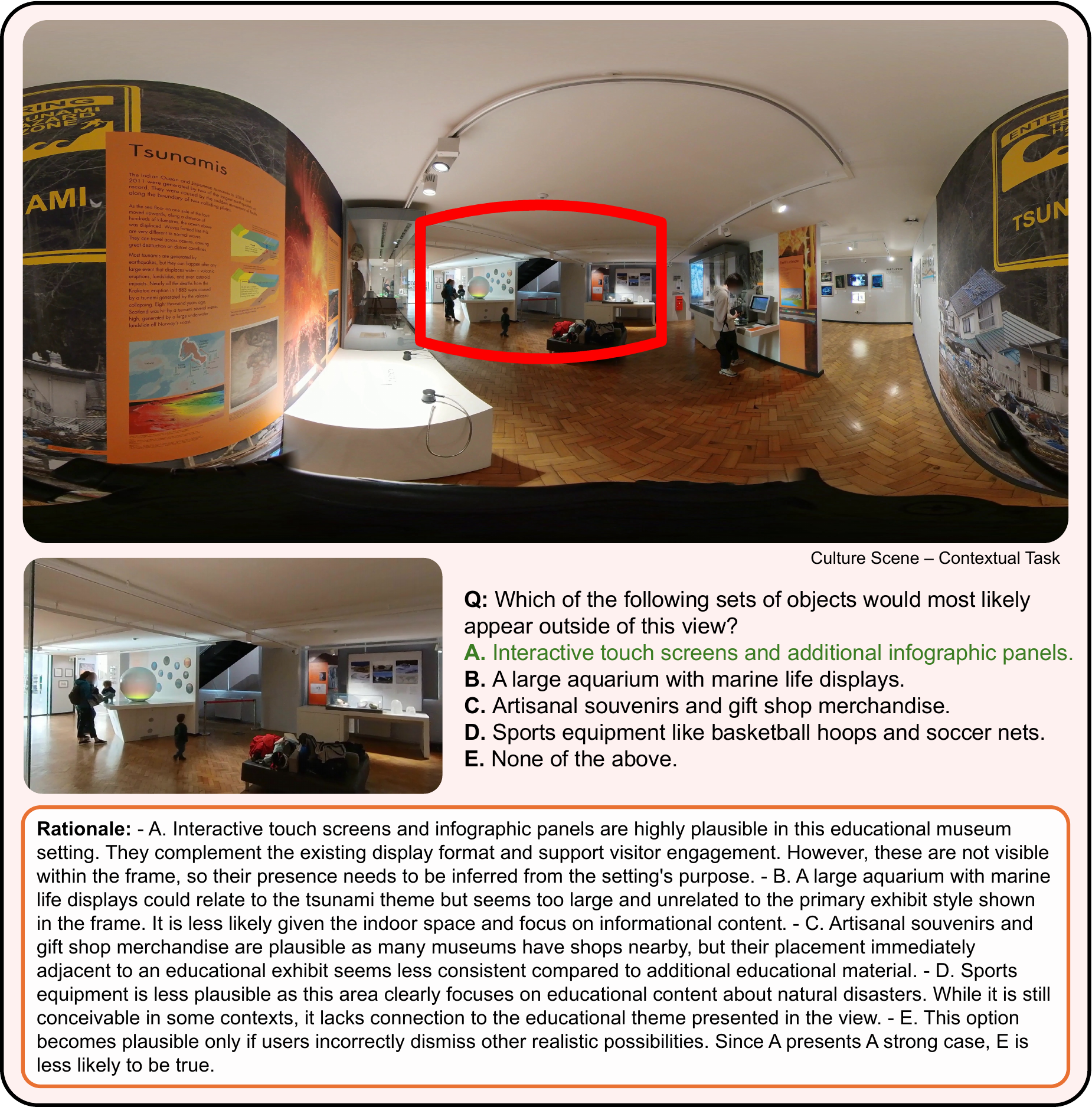}
    \caption{Example of a contextual question on a culture scene. The corresponding perspective projection is outlined in red and the correct answer is highlighted in green.}
    \label{fig:example_culture}
\end{figure*}

\begin{figure*}[ht]
    \centering
    \includegraphics[width=\linewidth]{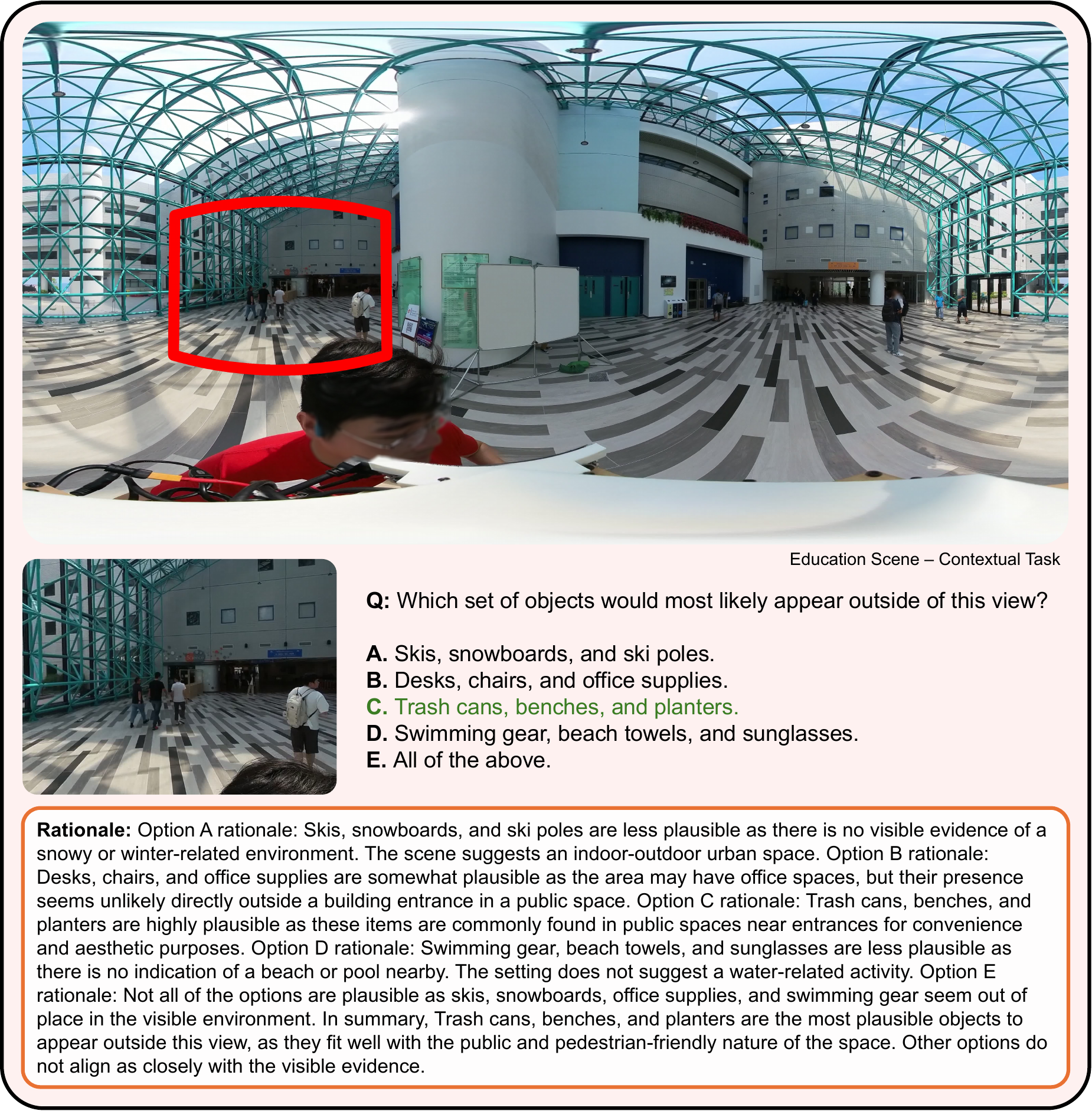}
    \caption{Example of a contextual question on an education scene. The corresponding perspective projection is outlined in red and the correct answer is highlighted in green.}
    \label{fig:example_education}
\end{figure*}

\begin{figure*}[ht]
    \centering
    \includegraphics[width=\linewidth]{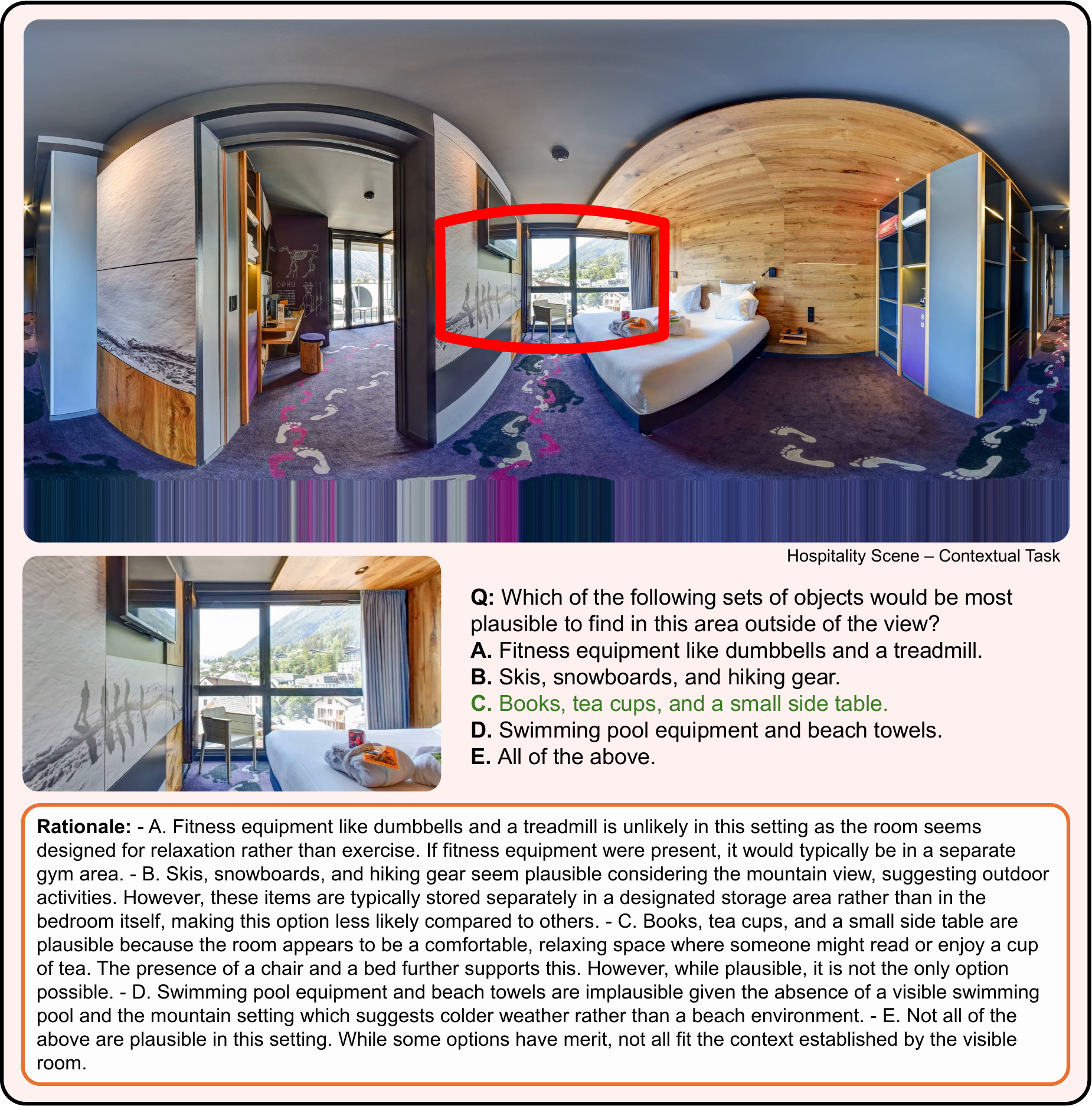}
    \caption{Example of a contextual question on a hospitality scene. The corresponding perspective projection is outlined in red and the correct answer is highlighted in green.}
    \label{fig:example_hospitality_2}
\end{figure*}

\begin{figure*}[ht]
    \centering
    \includegraphics[width=\linewidth]{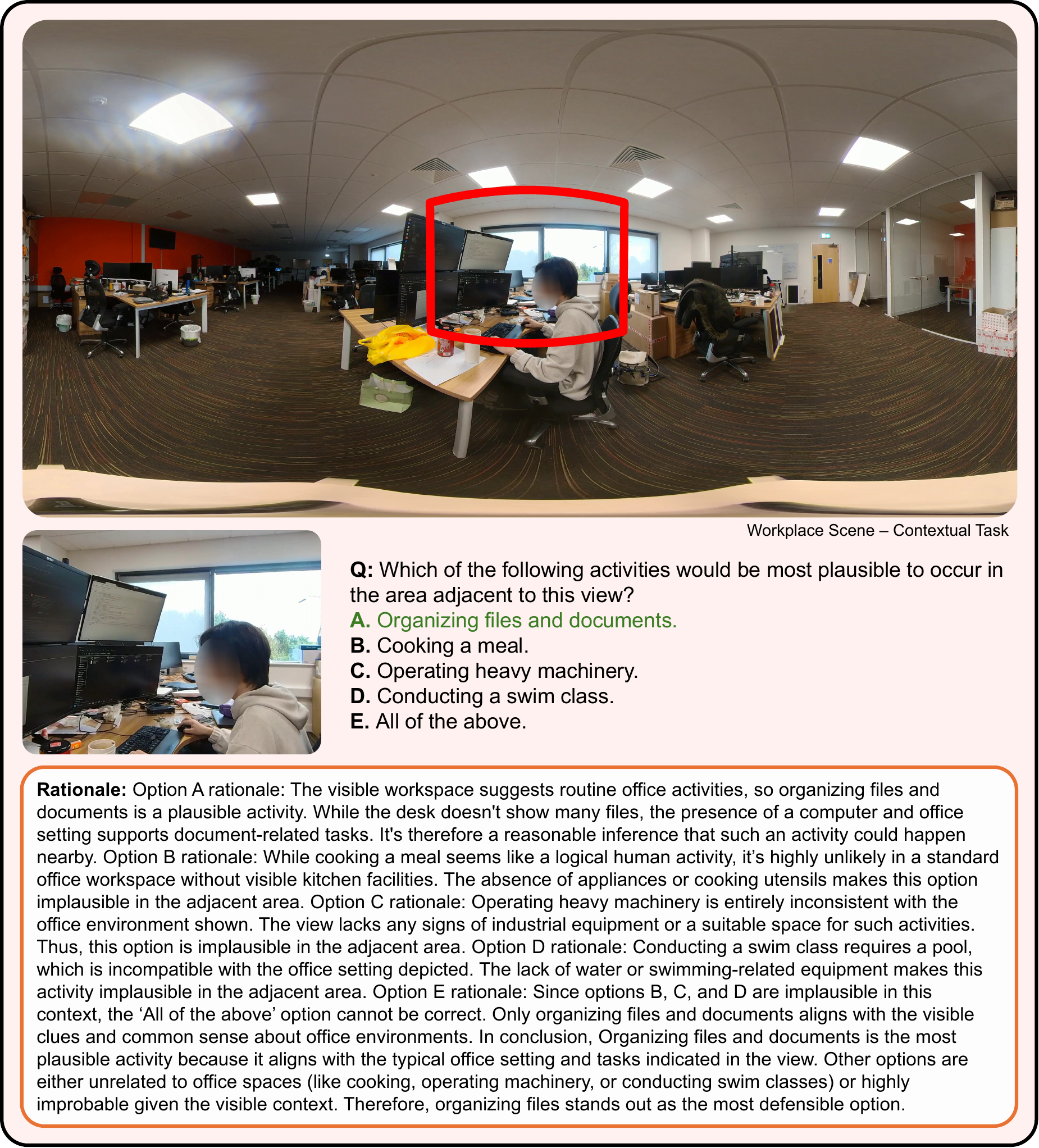}
    \caption{Example of a contextual question on a workplace scene. The corresponding perspective projection is outlined in red and the correct answer is highlighted in green.}
    \label{fig:example_workplace}
\end{figure*}

\begin{figure*}[ht]
    \centering
    \includegraphics[width=\linewidth]{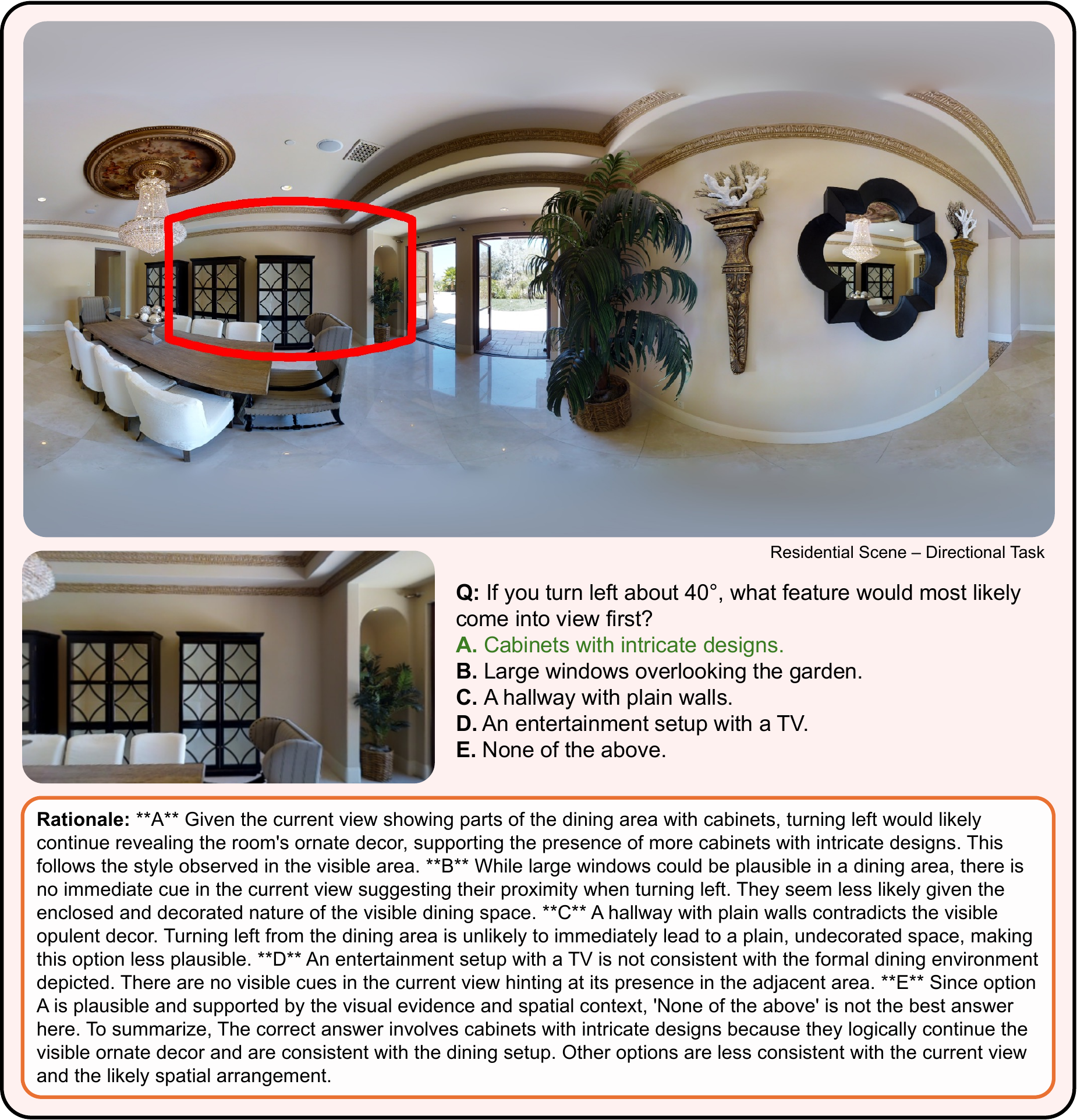}
    \caption{Example of a directional question on a residential scene. The corresponding perspective projection is outlined in red and the correct answer is highlighted in green.}
    \label{fig:example_residential}
\end{figure*}

\begin{figure*}[ht]
    \centering
    \includegraphics[width=\linewidth]{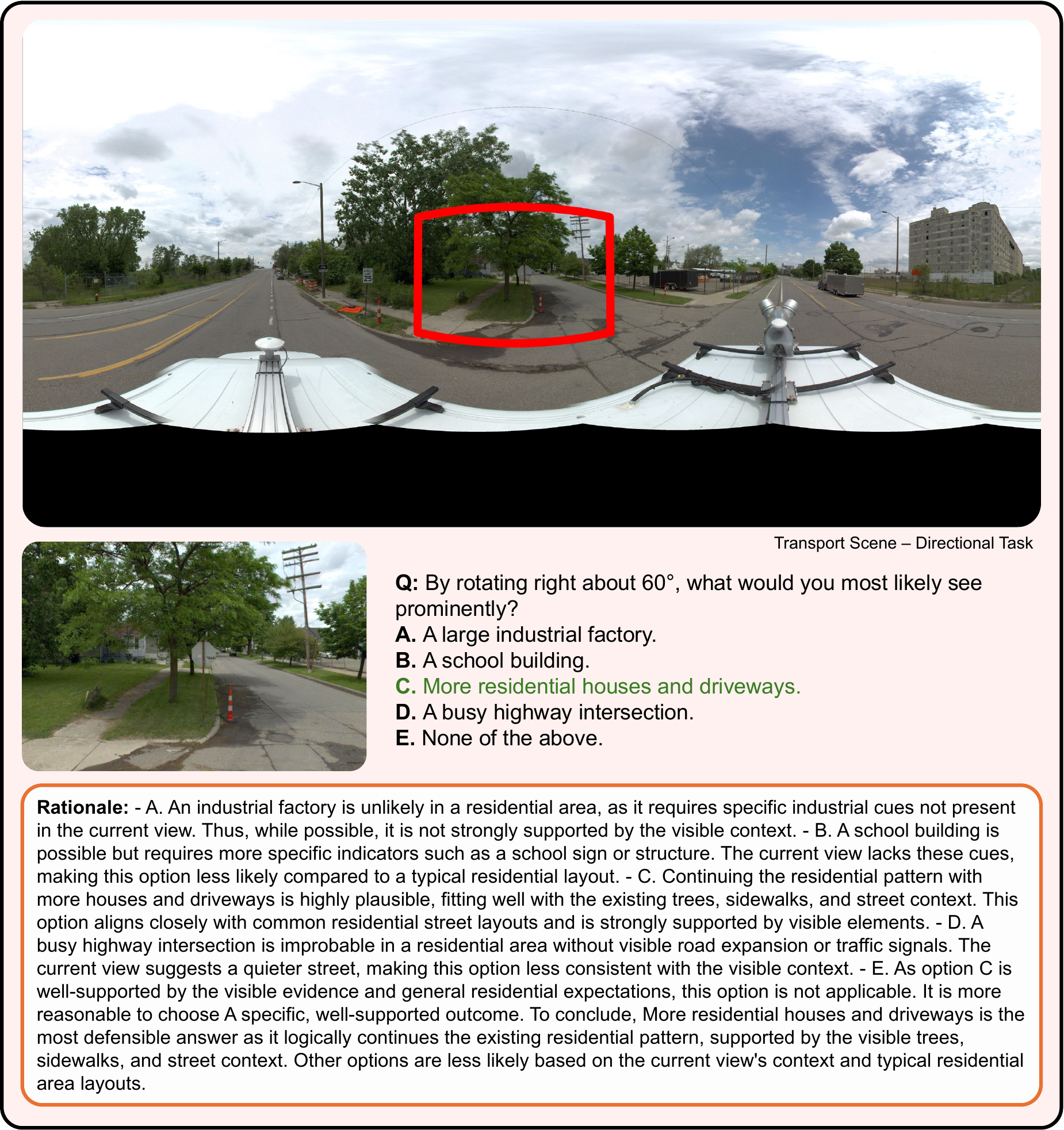}
    \caption{Example of a directional question on a transport scene. The corresponding perspective projection is outlined in red and the correct answer is highlighted in green.}
    \label{fig:example_transport}
\end{figure*}

\begin{figure*}[ht]
    \centering
    \includegraphics[width=\linewidth]{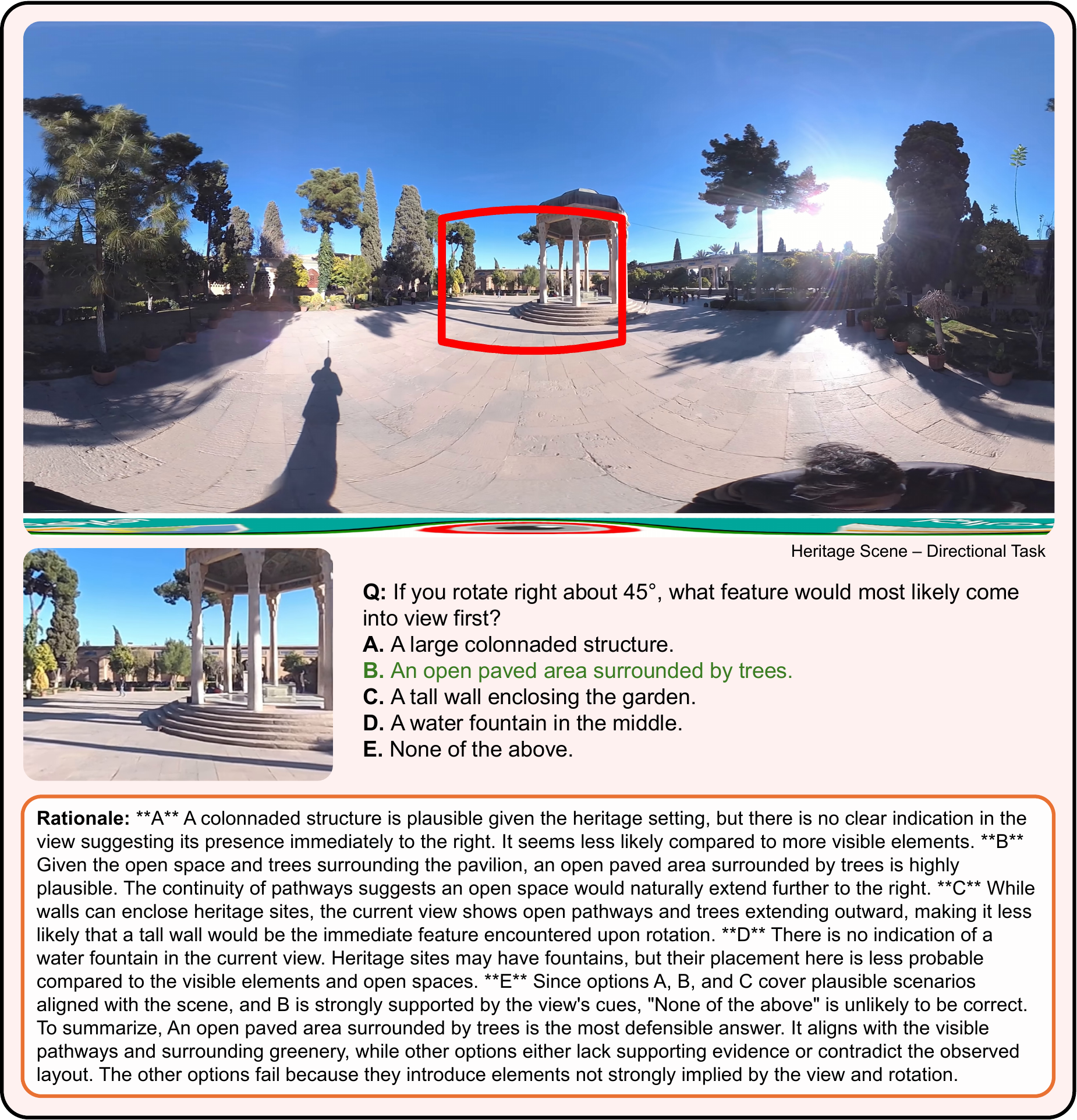}
    \caption{Example of a directional question on a heritage scene. The corresponding perspective projection is outlined in red and the correct answer is highlighted in green.}
    \label{fig:example_heritage}
\end{figure*}

\begin{figure*}[ht]
    \centering
    \includegraphics[width=\linewidth]{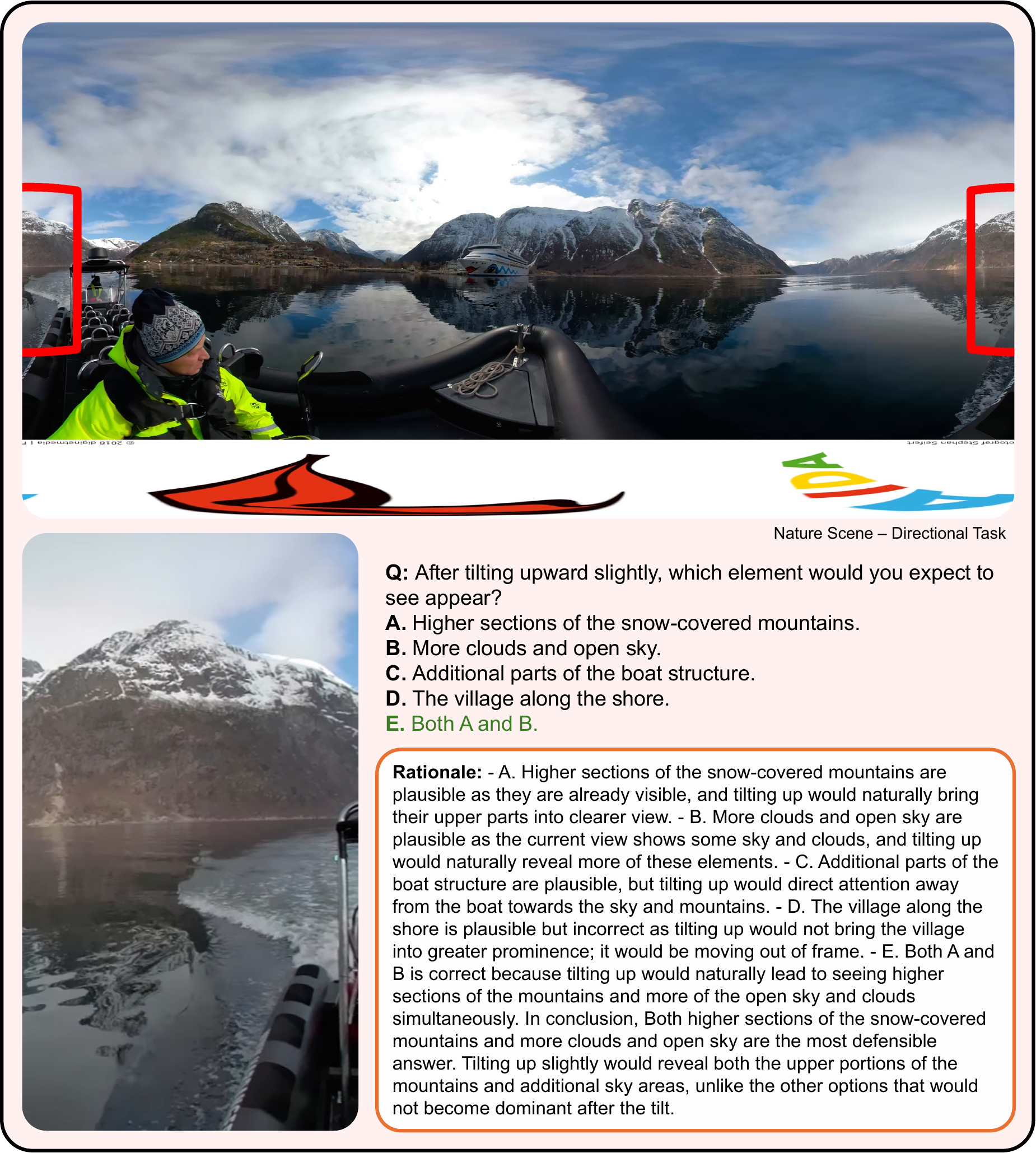}
    \caption{Example of a directional question on a nature scene. The corresponding perspective projection is outlined in red and the correct answer is highlighted in green.}
    \label{fig:example_nature_3}
\end{figure*}

\begin{figure*}[ht]
    \centering
    \includegraphics[width=\linewidth]{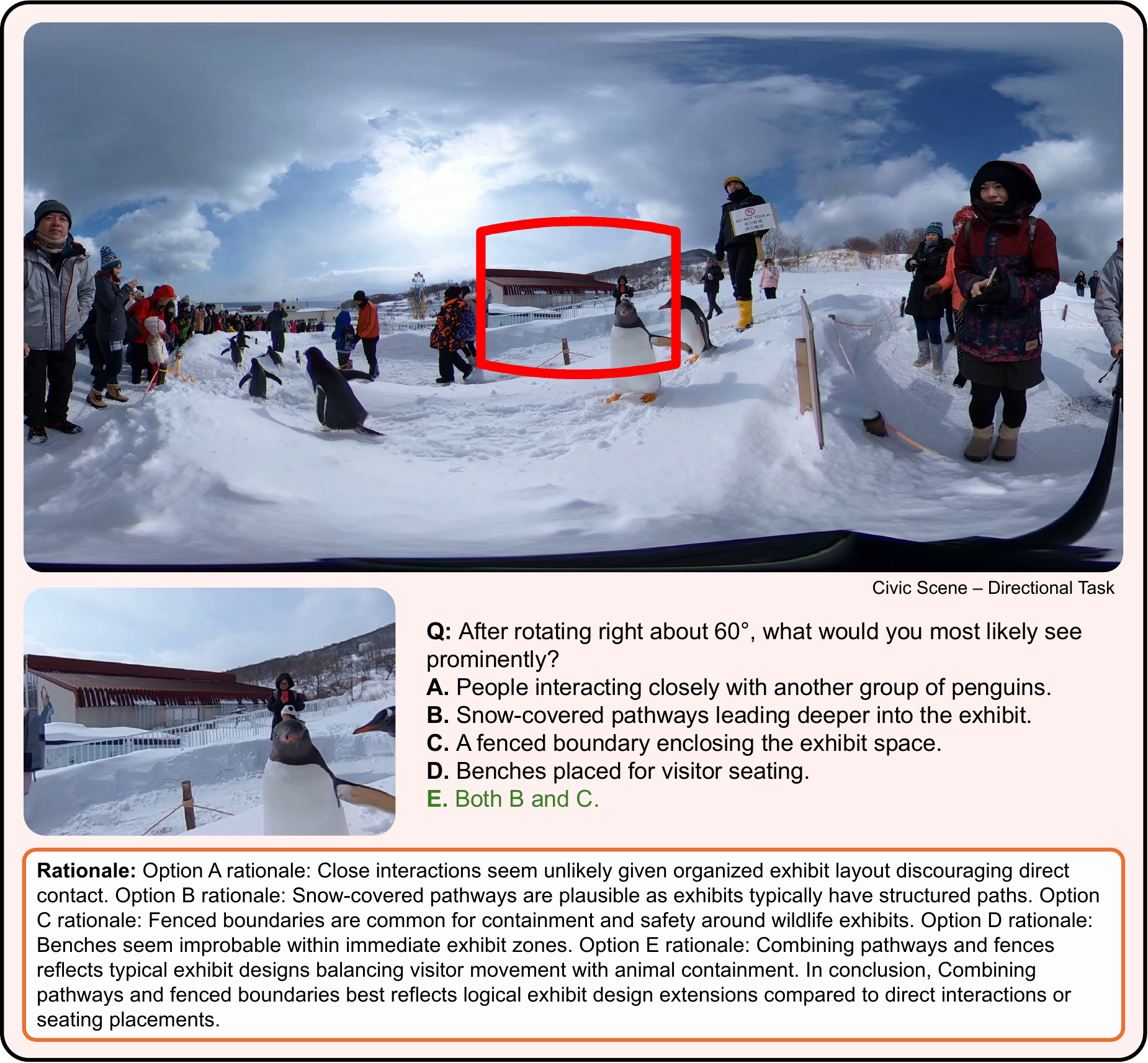}
    \caption{Example of a directional question on a civic scene. The corresponding perspective projection is outlined in red and the correct answer is highlighted in green.}
    \label{fig:example_civic_2}
\end{figure*}

\begin{figure*}[ht]
    \centering
    \includegraphics[width=\linewidth]{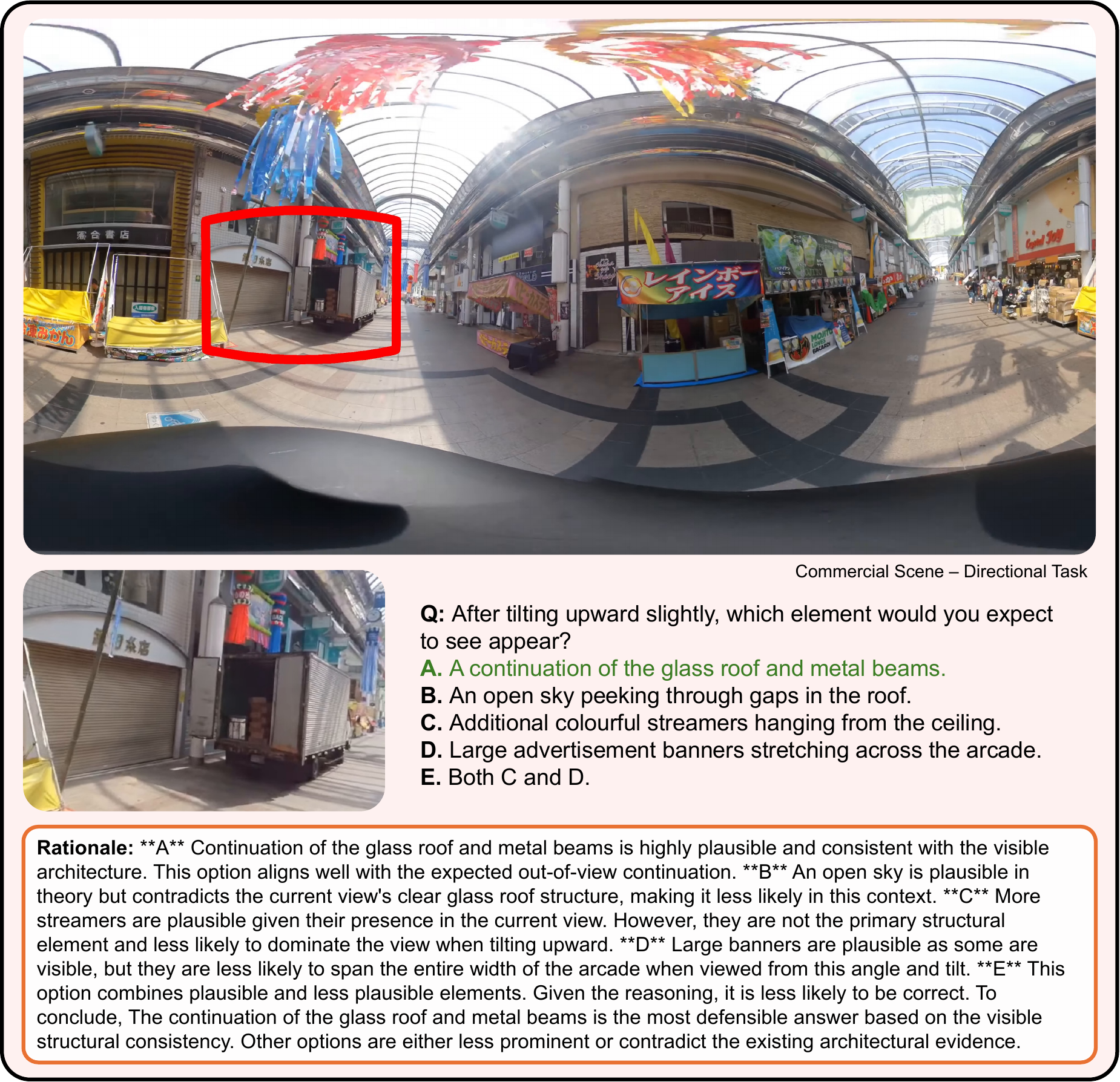}
    \caption{Example of a directional question on a commercial scene. The corresponding perspective projection is outlined in red and the correct answer is highlighted in green.}
    \label{fig:example_commercial}
\end{figure*}

Figures~\ref{fig:example_rural} to~\ref{fig:example_commercial} show representative example visualizations of the synthetic VQAs across 11 scene categories, including 5 contextual and 6 directional questions.

\section{Further Benchmark Details}
\label{supp:more_benchmark}

\subsection{Benchmark Statistics}

\begin{figure}[ht]
    \centering
    \includegraphics[width=0.65\linewidth]{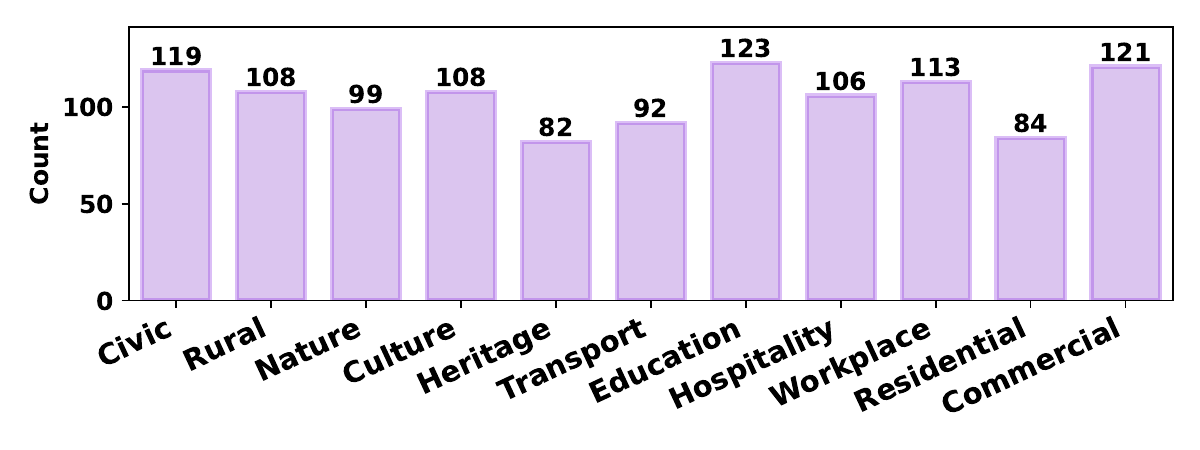}
    \caption{Distribution of scene categories across VQAs in the \ourbenchnospace.}
    \label{fig:test_category_dist}
\end{figure}

In constructing \ourbenchnospace, we consider not only the diversity of the perspective-projected views produced from different view metadata but also the overall composition of the benchmark across scenes, task types, and answer options.
Figure~\ref{fig:test_category_dist} shows the distribution of scene categories in the benchmark.
The benchmark draws on 77 source panoramas: 47 outdoor and 30 indoor panoramas, contributing 61.4\% and 38.6\% of the questions, respectively.
At the task level, the benchmark contains 580 contextual questions and 575 directional questions, resulting in a nearly balanced composition of task types.
At the answer level, the choices are evenly distributed across the five options (A: 234, B: 228, C: 238, D: 228, E: 227).

The benchmark spans diverse scenes with balanced task and answer distributions. Future work could further expand its scene and geographic coverage.

\vspace{-1em}
\vspace{-1em}

\begin{figure*}[ht]
    \centering
    \includegraphics[width=\linewidth]{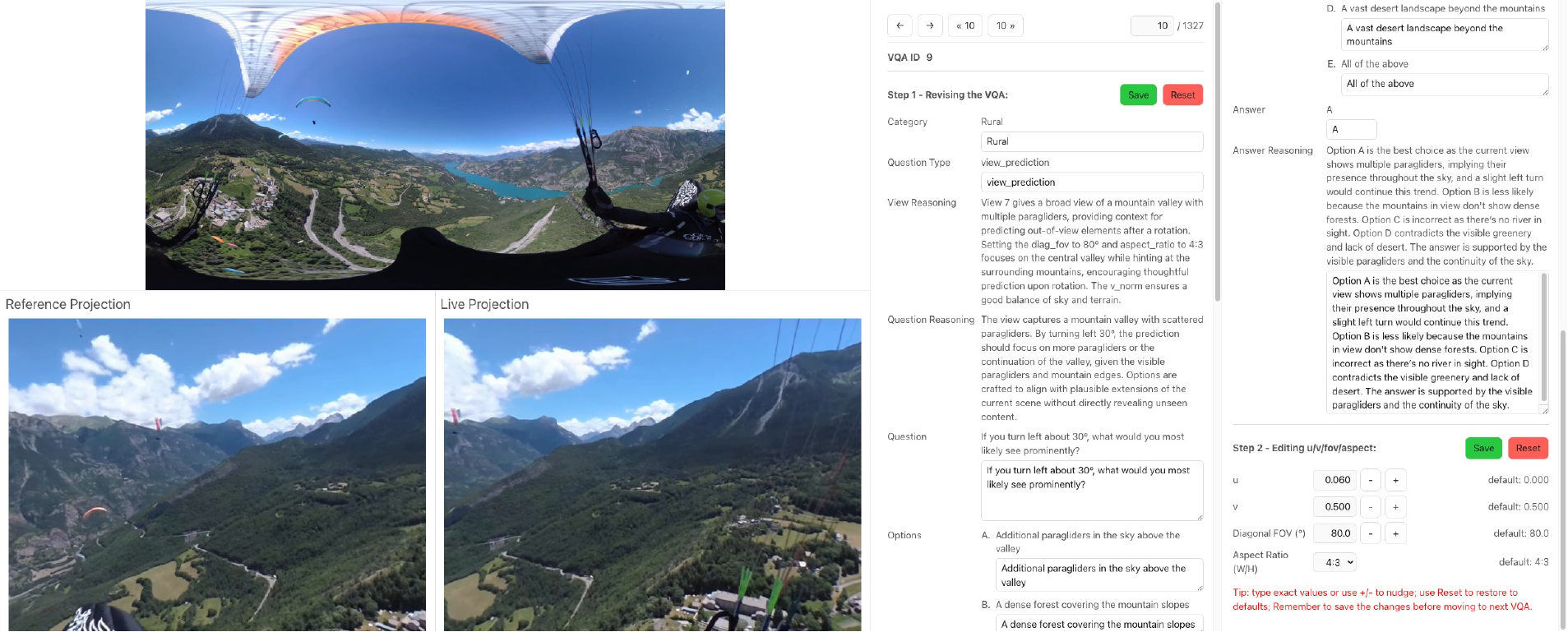}
    \caption{\textbf{VQA validation interface.} Each annotator first examines the reference projection and the VQA proposal displayed in the right panel, while consulting the full panorama in the top-left for additional context. They then review the rationale in the \textit{Answer Reasoning} section and revise the question accordingly. Any modification of the view metadata is immediately reflected in the live projection grid. The interface provides default fields along with editable input boxes to support efficient correction and refinement.}
    \label{fig:validation_interface}
\end{figure*}

\subsection{Construction Details}
\noindent\textbf{Separation of training and benchmark sources.}
We collected 21,090 panoramic images, of which 17,209 remained after Stage~1 preprocessing.
Of these, 16,463 were used to construct \ourdatasetnospace.
The remaining 746 panoramas were excluded from training-set construction because they belong to the same scenes (\eg, in Matterport3D and 360Loc) or source videos (\eg, in 360+x and 360-1M) as the 77 panoramas selected for \ourbenchnospace, with seven representative panoramas per scene category.
Thus, the exclusion operates at the scene or source-video level rather than only at the individual-image level.

\noindent\textbf{Verification and refinement.}
Five volunteers (aged 20–30) perform multi-round proposal verification and refinement, where each proposal requires roughly four to five minutes on average.
To facilitate efficient review and modification, we develop a dedicated web-based annotation interface (see Figure~\ref{fig:validation_interface}) that provides a user friendly access to the proposal, as well as the editable real-time projection function with the associated panorama as a reference.

As an additional quality control step, we conduct a description-based test in which only textual descriptions of visible objects, generated by an assistant model, are provided without the image. This test is designed to identify questions that can be answered without in-depth spatial reasoning.
Based on the results, we highlight approximately 25\% of the questions that can still be correctly answered under this setting and send them back to annotators for targeted revision. During this process, annotators are encouraged to refine the question wording or options so that answering the question requires grounding in spatial cues or scene layout rather than relying solely on object-level descriptions for contextual knowledge reasoning.

\subsection{User Study}
We conducted a user study with ten volunteers (aged 20–30) who were not involved in the annotation process, to evaluate the reliability and difficulty of the benchmark.
For each task, five questions were randomly sampled from each scene category, and each participant was assigned 11 randomly selected questions.
Across all responses, human participants achieved an average choice accuracy of 86.0\%.
Among the questions, 33 were answered by multiple participants, yielding a Fleiss’s $\kappa$ of 0.82, indicating strong agreement.
These results suggest that the benchmark is reasonable and moderately challenging, offering a meaningful basis for evaluating model reasoning ability.

We acknowledge that human participants may reach correct answers through faulty reasoning.
Consequently, the current human--model performance gap can only be assessed through choice accuracy.

\subsection{Agreement on Rationale Evaluation}
We report inter-annotator agreement on a balanced subset of 150 response sampled from GPT-5, Qwen2.5-VL-7B-Instruct, and its fine-tuned variant (\ie, Qwen2.5-VL-7B-Instruct + \ourdatasetnospace) in Table~\ref{tab:taskwise_alignment}.
Three human annotators independently label rationale correctness. 
Consensus labels are obtained via majority voting.
We report accuracy and Cohen’s kappa against the human consensus, Cohen’s kappa with each annotator and Fleiss’ kappa among all annotators.
GPT5-mini exhibits moderate agreement with human annotations across both tasks. Notably, the agreement between GPT5-mini and the human majority ($\kappa = 0.417$) is comparable to the agreement observed between individual human annotators ($\kappa \approx 0.44$--$0.53$), suggesting that the judge provides a reasonably consistent signal for automated evaluation.

\begin{table}[h]
  \centering
  \caption{Agreement between GPT5-mini judge and human.}
  \label{tab:taskwise_alignment}
  \resizebox{0.8\linewidth}{!}{
      \begin{tabular}{lcc|cccc}
        \toprule
        Task Type 
        & Acc. (majority)
        & $\kappa$(majority)
        & $\kappa$(Human1) 
        & $\kappa$(Human2) 
        & $\kappa$(Human3) 
        & Fleiss' $\kappa$ (all) \\
        \midrule
        Contextual   & 90.12 & 0.648 & 0.344 & 0.648 & 0.613 & 0.611 \\
        Directional  & 82.61 & 0.425 & 0.497 & 0.425 & 0.425 & 0.618 \\
        \midrule
        \textbf{Overall} & \textbf{86.67} & \textbf{0.417} & \textbf{0.439} & \textbf{0.532} & \textbf{0.517} & \textbf{0.616} \\
        \bottomrule
      \end{tabular}
    }
\end{table}

\noindent\textbf{Robustness to an Alternative Judge.}
We further replace GPT-5-mini with GPT-5.6-Sol~\cite{openai2026gpt56sol} to assess the judge robustness.
On the same subset of 150 model responses, GPT-5.6-Sol achieves higher agreement with the human consensus, improving overall Cohen's $\kappa$ from 0.417 to 0.488.
We then re-evaluate all 18 model entries in Table~\ref{tab:oov_bench_results} using GPT-5.6-Sol.
The resulting model rankings exhibit a Spearman rank correlation coefficient of 0.930 with the original rankings, indicating strong rank agreement across the two judges.
These results suggest that GPT-5.6-Sol acts as a stricter judge while achieving very strong rank agreement with the GPT-5-mini-based evaluation, supporting the robustness of the rationale evaluation protocol. 

\section{More Results and Visualizations}
\label{supp:more_results}

\subsection{Description-based Evaluation}
\label{supp:description_only}
We evaluate five models under a description-based setting, replacing visual inputs with object descriptions generated by GPT-4o.
Table~\ref{tab:description_only} includes GPT-5, GPT-4o and three additional open-source models, spanning ranks 2, 3, 6, 10, and 16 in Table~\ref{tab:oov_bench_results}.

Replacing images with object descriptions consistently reduces choice accuracy on both contextual and directional tasks for all five models.
The drop is particularly pronounced for MiniCPM-V 4.5, at 30.00\% and 24.87\% on contextual and directional tasks, respectively.
These results provide additional evidence that adequate visual perception contributes to effective OOV reasoning, consistent with the perception limitation discussed in Section~\ref{sec:main_results}.

\begin{table}[htbp]
\centering
\caption{\textbf{Performance under the description-based setting.} We report choice accuracy. $^\dagger$denotes replacing images with object descriptions, with percentage-point drops shown in parentheses. Ranks follow Table~\ref{tab:oov_bench_results}.}
\label{tab:description_only}
\resizebox{0.8\textwidth}{!}{%
\begin{tabular}{l|c|ccc}
\toprule
\textbf{Model} & \textbf{Rank} & \textbf{Contextual} & \textbf{Directional} & \textbf{Overall} \\
\midrule
GPT-5 & 2 & 70.86 & 57.04 & 63.98 \\
$^\dagger$GPT-5 & -- & 63.79 ($-7.07$) & 44.52 ($-12.52$) & 54.20 ($-9.78$) \\
\midrule
MiniCPM-V 4.5 & 3 & 68.79 & 57.04 & 62.94 \\
$^\dagger$MiniCPM-V 4.5 & -- & 38.79 ($-30.00$) & 32.17 ($-24.87$) & 35.50 ($-27.44$) \\
\midrule
GPT-4o & 6 & 71.72 & 55.83 & 63.81 \\
$^\dagger$GPT-4o & -- & 64.14 ($-7.58$) & 41.74 ($-14.09$) & 52.99 ($-10.82$) \\
\midrule
Qwen3-VL-8B-Instruct & 10 & 58.10 & 48.35 & 53.25 \\
$^\dagger$Qwen3-VL-8B-Instruct & -- & 54.31 ($-3.79$) & 35.30 ($-13.05$) & 44.85 ($-8.40$) \\
\midrule
Ovis2.5-9B & 16 & 66.21 & 43.65 & 54.98 \\
$^\dagger$Ovis2.5-9B & -- & 60.00 ($-6.21$) & 41.39 ($-2.26$) & 50.74 ($-4.24$) \\
\bottomrule
\end{tabular}%
}
\end{table}

\subsection{Robustness to Question Rephrasing}
\label{supp:rephrased_benchmark}
We rephrase \ourbench with DeepSeek-V4-Flash~\cite{deepseekai2026deepseekv4} and evaluate four models from Table~\ref{tab:oov_bench_results}.
In Table~\ref{tab:rephrased_benchmark}, the original-question rows report results on \ourbenchnospace, while the rephrased-question rows report results on its rephrased version.
Overall performance changes little, and our fine-tuned Qwen3-VL-8B remains ahead of GPT-5.
The fine-tuning gains therefore persist after independently rephrasing every question.

\begin{table}[ht]
\centering
\caption{\textbf{Performance on rephrased questions.} We compare model performance on the original questions and their DeepSeek-V4-Flash rephrasings. Choice, rationale, and joint accuracy are reported as percentages; joint accuracy columns are \tinycolorbox{gray!10}{shaded}.}
\label{tab:rephrased_benchmark}
\resizebox{\textwidth}{!}{%
\begin{tabular}{l|cc>{\columncolor{gray!10}}c cc>{\columncolor{gray!10}}c|cc>{\columncolor{gray!10}}c}
\toprule
\multirow{2.5}{*}{\textbf{Model}} & \multicolumn{3}{c}{\textbf{Contextual}} & \multicolumn{3}{c}{\textbf{Directional}} & \multicolumn{3}{|c}{\textbf{Overall}} \\
\cmidrule(lr){2-4}\cmidrule(lr){5-7}\cmidrule(lr){8-10}
 & Choice & Rationale & Joint & Choice & Rationale & Joint & Choice & Rationale & Joint \\
\midrule
\multicolumn{10}{l}{\textit{Original questions}} \\
GPT-5 & 70.86 & 91.73 & 65.00 & 57.04 & 95.43 & 54.43 & 63.98 & 93.37 & 59.74 \\
GPT-4o & 71.72 & 83.89 & 60.17 & 55.83 & 89.41 & 49.91 & 63.81 & 86.30 & 55.06 \\
Qwen3-VL-8B-Instruct & 58.10 & 94.96 & 55.17 & 48.35 & 90.65 & 43.83 & 53.25 & 93.01 & 49.53 \\
Qwen3-VL-8B + \ourdatasetnospace & 77.07 & 93.29 & 71.90 & 57.22 & 89.36 & 51.13 & 67.19 & 91.62 & 61.56 \\
\midrule
\multicolumn{10}{l}{\textit{Rephrased questions}} \\
GPT-5 & 68.28 & 93.43 & 63.79 & 57.57 & 96.07 & 55.30 & 62.94 & 94.64 & 59.57 \\
GPT-4o & 71.38 & 86.71 & 61.90 & 50.78 & 88.70 & 45.04 & 61.13 & 87.54 & 53.51 \\
Qwen3-VL-8B-Instruct & 61.38 & 96.07 & 58.97 & 50.09 & 93.75 & 46.96 & 55.76 & 95.03 & 52.99 \\
Qwen3-VL-8B + \ourdatasetnospace & 77.41 & 94.88 & 73.45 & 56.87 & 91.13 & 51.83 & 67.19 & 93.30 & 62.68 \\
\bottomrule
\end{tabular}%
}
\end{table}

\subsection{Few-shot and CoT Baselines}
\label{supp:prompting_baselines}
We evaluate 2-shot, 4-shot, and CoT prompting on three open-source models in Table~\ref{tab:prompting_baselines}.
The 2-shot setting uses one contextual and one directional VQA exemplar; the 4-shot setting uses two of each, drawn from the training set without scene overlap with the benchmark.
For CoT, we instruct the model to first describe the visible scene and infer likely contents outside the frame in each direction, then reason through the options and select the most defensible answer.

Only Qwen3-VL-8B benefits from few-shot prompting, with joint accuracy increasing from 49.53\% (zero-shot) to 55.58\% (2-shot) and 56.80\% (4-shot), while the other two models do not benefit.
CoT improves both choice and rationale accuracy for Qwen3-VL-8B and InternVL3.5-8B, increasing their joint accuracy from 49.53\% to 57.66\% and from 43.72\% to 51.26\%, respectively.
This prompting strategy allows the models to verbalize the visible scene before reasoning about the options, while LLaVA-NeXT-Mistral-7B continues to perform poorly.
These results suggest that prompt engineering can improve OOV performance, although the benefits vary across models.

\begin{table}[htbp]
\centering
\caption{\textbf{Effects of prompting strategies.} We compare zero-shot performance with 2-shot, 4-shot, and CoT prompting for three open-source MLLMs. Choice, rationale, and joint accuracy are reported as percentages; joint accuracy columns are \tinycolorbox{gray!10}{shaded}.}
\label{tab:prompting_baselines}
\resizebox{\textwidth}{!}{%
\begin{tabular}{l|cc>{\columncolor{gray!10}}c cc>{\columncolor{gray!10}}c|cc>{\columncolor{gray!10}}c}
\toprule
\multirow{2.5}{*}{\textbf{Model}} & \multicolumn{3}{c}{\textbf{Contextual}} & \multicolumn{3}{c}{\textbf{Directional}} & \multicolumn{3}{|c}{\textbf{Overall}} \\
\cmidrule(lr){2-4}\cmidrule(lr){5-7}\cmidrule(lr){8-10}
 & Choice & Rationale & Joint & Choice & Rationale & Joint & Choice & Rationale & Joint \\
\midrule
Qwen3-VL-8B (zero-shot) & 58.10 & 94.96 & 55.17 & 48.35 & 90.65 & 43.83 & 53.25 & 93.01 & 49.53 \\
\quad + 2-shot & 65.34 & 98.15 & 64.14 & 50.96 & 92.15 & 46.96 & 58.18 & 95.54 & 55.58 \\
\quad + 4-shot & 66.03 & 98.43 & 65.00 & 51.48 & 94.26 & 48.52 & 58.79 & 96.61 & 56.80 \\
\quad + CoT & 65.17 & 99.74 & 65.00 & 52.00 & 96.66 & 50.26 & 58.61 & 98.38 & 57.66 \\
\midrule
InternVL3.5-8B (zero-shot) & 56.90 & 90.00 & 51.21 & 45.39 & 79.69 & 36.17 & 51.17 & 85.45 & 43.72 \\
\quad + 2-shot & 49.83 & 82.70 & 41.21 & 38.43 & 81.90 & 31.48 & 44.16 & 82.35 & 36.36 \\
\quad + 4-shot & 45.69 & 75.47 & 34.48 & 32.17 & 71.35 & 22.96 & 38.96 & 73.78 & 28.74 \\
\quad + CoT & 61.55 & 96.08 & 59.14 & 49.22 & 87.99 & 43.30 & 55.41 & 92.50 & 51.26 \\
\midrule
LLaVA-NeXT-Mistral-7B (zero-shot) & 42.07 & 39.34 & 16.55 & 29.04 & 29.94 & 8.69 & 35.58 & 35.52 & 12.64 \\
\quad + 2-shot & 33.28 & 28.50 & 9.48 & 33.39 & 22.40 & 7.48 & 33.33 & 25.45 & 8.48 \\
\quad + 4-shot & 28.62 & 33.73 & 9.66 & 26.61 & 31.37 & 8.35 & 27.62 & 32.60 & 9.00 \\
\quad + CoT & 27.59 & 41.88 & 11.55 & 32.52 & 38.50 & 12.52 & 30.04 & 40.06 & 12.03 \\
\bottomrule
\end{tabular}%
}
\end{table}

\subsection{Cross-paradigm Generalization to \ourbenchnospace}
\label{supp:mindcube_transfer}
To examine whether training for multi-view spatial reasoning transfers to OOV understanding, we evaluate MindCube-RL-PlainCogMap-FromSFT~\cite{yin2025spatial} on \ourbenchnospace.
This model is based on Qwen2.5-VL-3B and fine-tuned with reinforcement learning for multi-view spatial cognitive-map reasoning.
As shown in Table~\ref{tab:mindcube_transfer}, the model achieves 43.9\% choice accuracy and 9.5\% joint accuracy, compared with 51.3\% and 6.8\%, respectively, for the base model.
Thus, reinforcement learning for spatial cognitive-map reasoning provides a modest improvement in joint accuracy but reduces choice accuracy.
In contrast, training on \ourdataset improves performance on MindCube (Section~\ref{sec:cross-paradigm}).
This asymmetric transfer suggests that the two benchmarks assess related but distinct reasoning capabilities.

\begin{table}[htbp]
\centering
\caption{\textbf{MindCube-RL on \ourbenchnospace.} We compare MindCube-RL with its Qwen2.5-VL-3B-Instruct base model. Choice, rationale, and joint accuracy are reported as percentages; joint accuracy columns are \tinycolorbox{gray!10}{shaded}.}
\label{tab:mindcube_transfer}
\resizebox{\textwidth}{!}{%
\begin{tabular}{l|cc>{\columncolor{gray!10}}c cc>{\columncolor{gray!10}}c|cc>{\columncolor{gray!10}}c}
\toprule
\multirow{2}{*}{\textbf{Model}} & \multicolumn{3}{c}{\textbf{Contextual}} & \multicolumn{3}{c}{\textbf{Directional}} & \multicolumn{3}{|c}{\textbf{Overall}} \\
\cmidrule(lr){2-4}\cmidrule(lr){5-7}\cmidrule(lr){8-10}
 & Choice & Rationale & Joint & Choice & Rationale & Joint & Choice & Rationale & Joint \\
\midrule
Qwen2.5-VL-3B-Instruct & 57.93 & 13.99 & 8.10 & 44.70 & 12.06 & 5.39 & 51.34 & 13.15 & 6.75 \\
MindCube-RL (3B) & 48.97 & 34.51 & 16.90 & 38.78 & 5.38 & 2.09 & 43.90 & 21.70 & 9.52 \\
\bottomrule
\end{tabular}
}
\end{table}

\begin{figure}[ht]
    \centering
    \includegraphics[width=\linewidth]{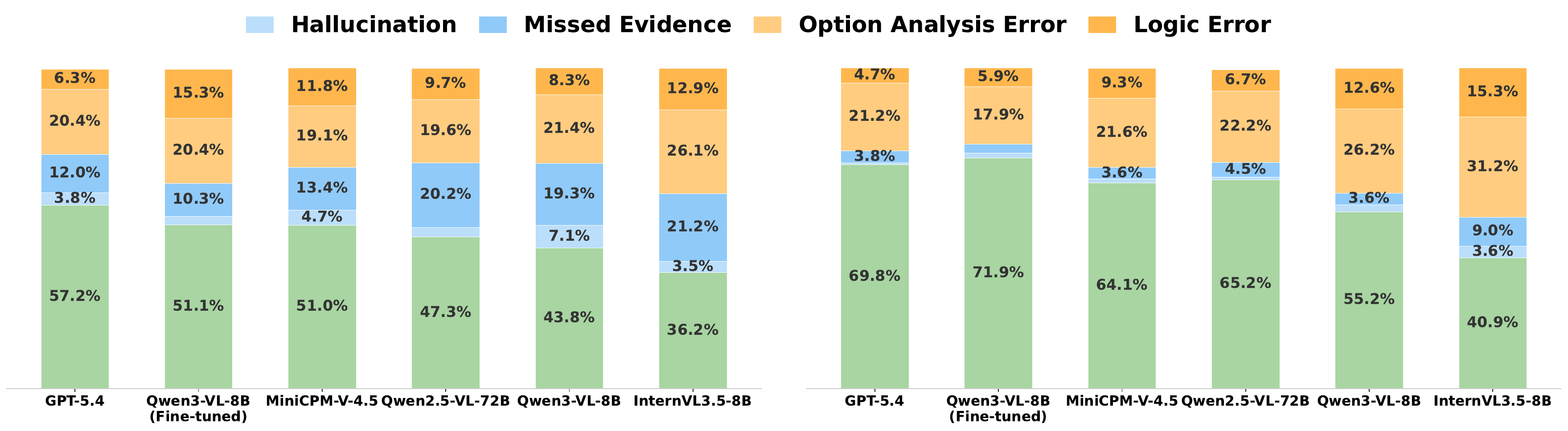}
    \caption{
    \textbf{Response distribution across models on directional (\textit{left}) and contextual (\textit{right}) tasks.}
    Each stacked bar shows the percentage of \textcolor{Green!60}{correct predictions} and error cases, with error subtypes grouped into \textcolor{cyan!60}{perception limitations} (hallucination, missed evidence) and \textcolor{orange!80}{reasoning limitations} (option analysis error, logic error).
    }
    \label{fig:error_breakdown_by_qtype}
\end{figure}

\subsection{Detailed Failure Case Analysis}
To better understand the sources of model failures, we perform an empirical failure analysis by grouping incorrect predictions into perception and reasoning limitations, each further divided into two sub-error types.
Figure~\ref{fig:error_breakdown_by_qtype} presents the error distribution for directional and contextual questions separately. 
Across all models, directional questions yield substantially lower accuracy than contextual ones, with gaps ranging from 5\% to 21\% (see Table~\ref{tab:oov_bench_results}), confirming that camera rotation instruction introduces a distinct layer of difficulty. 

Decomposing the errors reveals that this gap stems predominantly from perception failures rather than from reasoning failures. 
Specifically, missed evidence increases roughly three to four times on directional questions, averaging $\sim$16\% of total questions compared to $\sim$5\% on contextual ones, suggesting that camera rotation systematically disrupts models' ability to ground relevant spatial cues from the visible scene. 
Hallucination rates likewise rise modestly under directional settings. 
These results suggest that camera rotation instruction in the question primarily impairs the perceptual grounding stage, with the resulting errors propagating downstream to affect the final answer.
In contrast, reasoning errors remain uniformly high across both question types, suggesting that models fundamentally lack OOV-specific reasoning guidance, as such scenarios are largely underrepresented in existing training data and benchmarks.

These findings suggest that advancing OOV reasoning requires progress on two fronts. 
On the perceptual side, models need to move beyond object-level recognition toward richer geometric and spatial understanding, enabling more accurate and holistic scene grounding from the visible frame. 
On the reasoning side, models must learn to dynamically extract question-relevant visual evidence and leverage it to assess option plausibility, a capability that, when grounded in strong perception, promisingly extends well beyond single-view OOV settings, as reflected in the cross-paradigm generalization results on MindCube (Section~\ref{sec:cross-paradigm}).

\subsection{Per-scene Performance Analysis}
We present the detailed analysis of the joint accuracy for five representative MLLMs and one fine-tuned variant across 11 scene categories in Table~\ref{tab:scene_joint_acc}. 
Overall, GPT-5.4 achieves the strongest scene-level performance among base MLLMs, leading in four scene categories and maintaining competitive accuracy across most others. 
Despite this, performance across scenes remains within a broadly comparable range, with most base MLLM joint accuracies falling between 45\% and 66\%.
Commonly challenging scenes, \eg, \textit{Culture}, \textit{Residential} and \textit{Rural}, tend to yield lower accuracies across multiple models.
On the other hand, the fine-tuned model achieves the best performance in several categories, including \textit{Commercial}, \textit{Heritage}, \textit{Nature}, \textit{Residential}, and \textit{Rural}. 
These consistent gains in joint accuracy across diverse scene categories suggest that training with \ourdataset equips the model with stronger visual grounding and contextual reasoning, enabling more robust out-of-view understanding across real-world environments.  

Nevertheless, a notable performance gap persists across scene categories, suggesting that certain environments impose greater demands on visual understanding and contextual reasoning. 
A more systematic investigation of scene-robust training strategies and model designs remains an open and promising direction.

\begin{table}[h]
    \centering
    \caption{\textbf{Joint accuracy across scene categories.} Performance of five representative MLLMs and our fine-tuned model on \ourbenchnospace. The best result in each category is highlighted in bold.}
    \label{tab:scene_joint_acc}
    \resizebox{\linewidth}{!}{
    \begin{tabular}{l@{\hspace{14pt}}c@{\hspace{10pt}}c@{\hspace{10pt}}c@{\hspace{10pt}}c@{\hspace{10pt}}c@{\hspace{10pt}}c@{\hspace{10pt}}c@{\hspace{10pt}}c@{\hspace{10pt}}c@{\hspace{10pt}}c@{\hspace{10pt}}c}
    \toprule
    \textbf{Model} & \textbf{Civic} & \textbf{Commercial} & \textbf{Culture} & \textbf{Education} & \textbf{Heritage} & \textbf{Hospitality} & \textbf{Nature} & \textbf{Residential} & \textbf{Rural} & \textbf{Transport} & \textbf{Workplace} \\
    \midrule
    GPT-5.4 & \textbf{72.27} & 62.81 & 52.78 & 62.60 & 57.32 & \textbf{68.87} & 64.65 & 51.19 & 53.70 & \textbf{76.08} & \textbf{73.45} \\
    GPT-5 & 57.98 & 61.15 & 54.63 & \textbf{66.67} & 50.00 & 66.04 & 62.63 & \textbf{54.76} & 49.07 & 66.30 & 64.61 \\
    MiniCPM-V 4.5 & 61.35 & 56.20 & \textbf{57.40} & 51.22 & \textbf{63.41} & 55.66 & 67.68 & 51.19 & 41.67 & 69.56 & 61.06 \\
    Qwen2.5-VL-32B-Instruct & 53.78 & 54.55 & 41.67 & 61.79 & 57.32 & 61.32 & 61.62 & 47.62 & 56.48 & 67.39 & 61.06 \\
    Qwen3-VL-8B-Instruct & 47.90 & 51.24 & 41.67 & 54.47 & 45.12 & 52.83 & 62.63 & 32.14 & 36.12 & 60.87 & 56.63 \\
    Qwen3-VL-8B-Instruct (SFT) & 62.19 & \textbf{66.94} & 54.63 & 55.28 & \textbf{63.41} & 60.38 & \textbf{72.73} & \textbf{54.76} & \textbf{62.96} & 70.65 & 54.87 \\
    \bottomrule
    \end{tabular}}
    \end{table}

\subsection{Role of Data Scaling}
Table~\ref{tab:ablation_data_scale} reports results on scaling the training set, which consistently improves the overall joint accuracy. 
While larger training set introduces richer visual and contextual diversity in terms of enhancing the generalization, the improvement in rationale precision does not always align with the gains in choice accuracy. 
This inconsistency suggests that reasoning about OOV tasks remains inherently challenging, particularly under the multi-choice VQA setting.

\begin{table}[h]
\centering
\caption{
\textbf{Ablation study on data scaling.} 
We report the number of training samples at different scales, together with the overall accuracies on the contextual and directional tasks.}
\label{tab:ablation_data_scale}
\resizebox{0.8\linewidth}{!}{
\begin{tabular}{cc|ccc|ccc}
\toprule
\multirow{2}{*}{\textbf{\#~\!Panorama}} & 
\multirow{2}{*}{\textbf{\#~\!Training samples}} & 
\multicolumn{3}{c|}{\textbf{InternVL3.5-8B}} & 
\multicolumn{3}{c}{\textbf{Qwen2.5-VL-7B-Instruct}} \\
\cmidrule(lr){3-5} \cmidrule(lr){6-8}
 &  & \textbf{Choice} & \textbf{Rationale} & \textbf{Joint} & \textbf{Choice} & \textbf{Rationale} & \textbf{Joint} \\
\midrule
1k  & 9{,}596   & 52.99 & 85.13 & 45.11 & 61.90 & 85.03 & 52.63 \\
3k  & 29{,}028  & 55.50 & 87.68 & 48.66 & 61.04 & 87.66 & 53.51 \\
5k  & 48{,}296  & 54.98 & 89.45 & 49.18 & \textbf{64.87} & 89.30 & 57.93 \\
10k & 96{,}470  & \textbf{57.92} & 86.85 & 50.30 & 63.90 & 89.84 & 57.41 \\
16k & 158{,}352 & 57.06 & \textbf{90.14} & \textbf{51.43} & 64.42 & \textbf{90.99} & \textbf{58.62} \\
\bottomrule
\end{tabular}
}
\end{table}

\subsection{More Visualizations and Application}
\begin{figure*}[ht]
    \centering
    \includegraphics[width=0.77\linewidth]{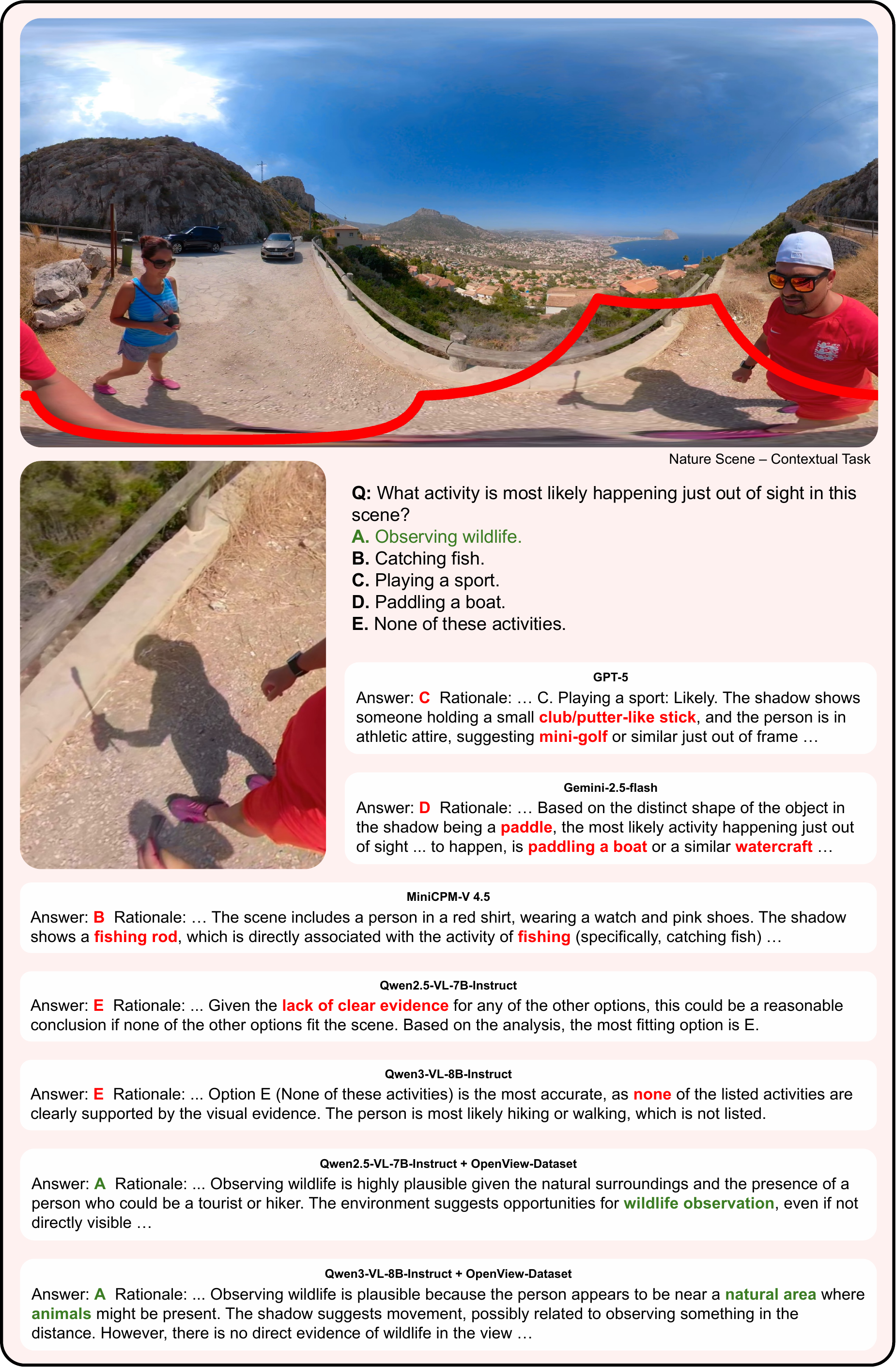}
    \caption{Outputs of top-performing MLLMs and two fine-tuned models on an example contextual OOV question in a nature scene. 
    While base models tend to rely heavily on the shadow of the person holding the 360-degree camera to infer possible activities, they fail to recognize the camera rig and show limited understanding of the surrounding environment.
    In contrast, our fine-tuned models capture a broader scene context and the natural surroundings, enabling more accurate and grounded reasoning about the correct answer.}
    \label{fig:test_case_1}
\end{figure*}

\begin{figure*}[ht]
    \centering
    \includegraphics[width=0.78\linewidth]{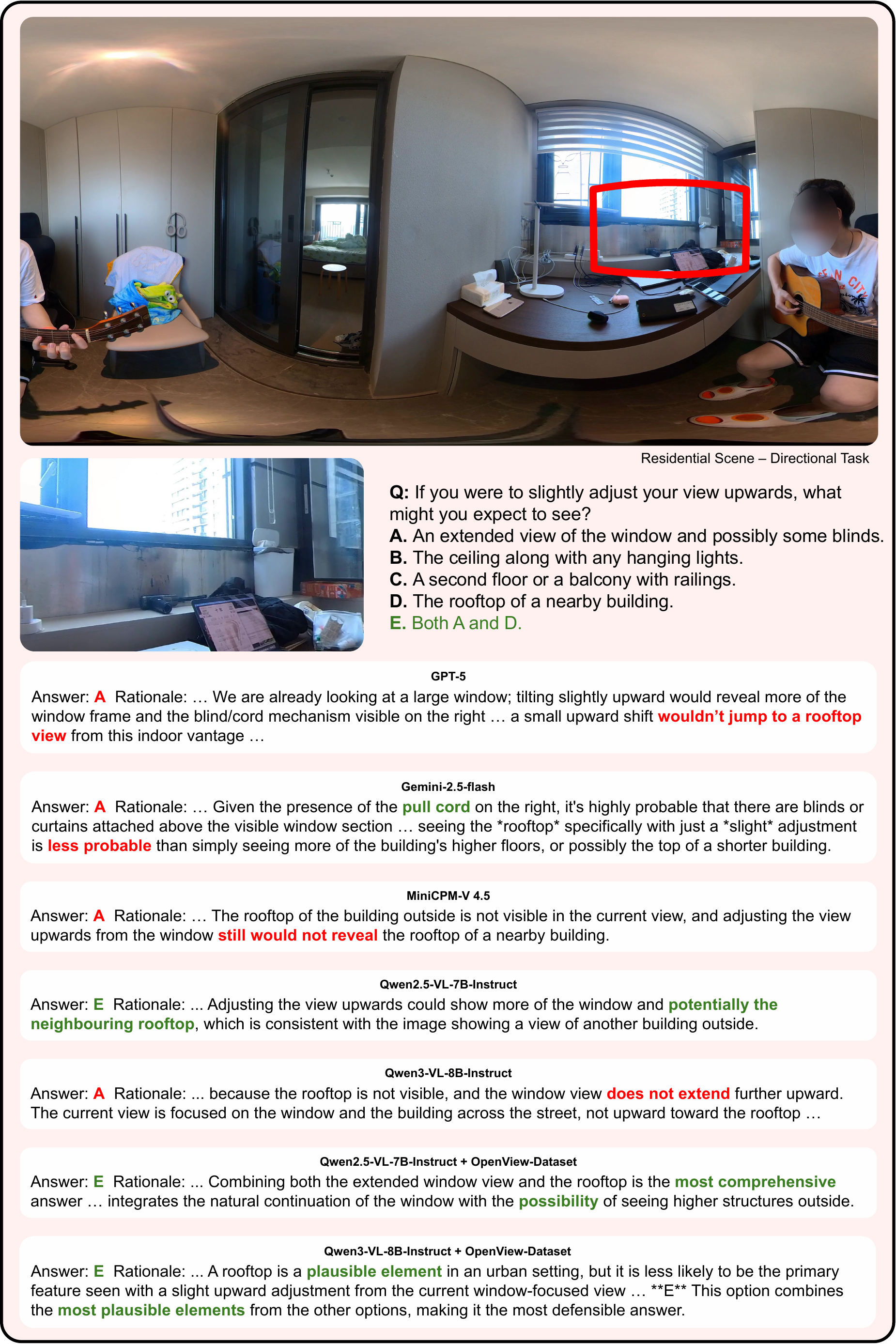}
    \caption{Outputs of top-performing MLLMs and two fine-tuned models on an example directional OOV question in a residential scene. 
    Most base models predict with high confidence that slightly tilts the camera upward would not reveal any rooftop of a nearby building.
    In contrast, our fine-tuned models consider both options as plausible, demonstrating a more nuanced and fine-grained visual understanding and better uncertainty awareness in OOV understanding.}
    \label{fig:test_case_2}
\end{figure*}

\begin{figure*}[ht]
    \centering
    \includegraphics[width=0.77\linewidth]{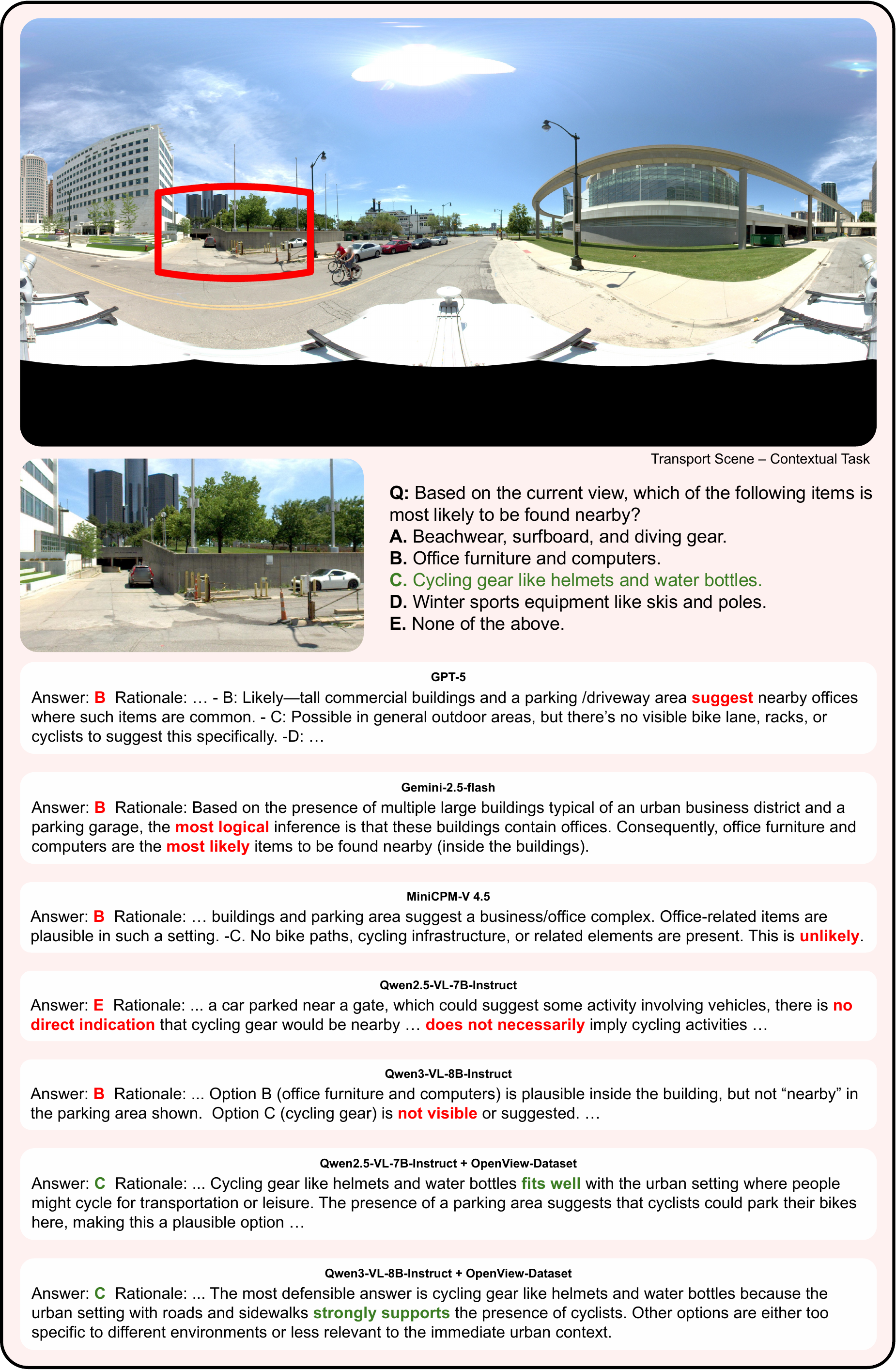}
    \caption{Outputs of top-performing MLLMs and two fine-tuned models on an example contextual OOV question in a transport scene.
    Most of the base models identify the building but incorrectly infer office furniture, revealing limited understanding of the broader street scene contents.
    In contrast, our fine-tuned models are able to correctly interpret the scene as an urban transport environment and infer plausible items such as cycling gear, demonstrating stronger context reasoning in OOV settings.
    }
    \label{fig:test_case_3}
\end{figure*}

\textbf{Visualizations of model performance.} The visualizations on the results of the top-performing MLLMs and two fine-tuned variants are provided in Figures~\ref{fig:test_case_1},~\ref{fig:test_case_2} and~\ref{fig:test_case_3}.
After fine-tuning on \ourdatasetnospace, all models produce more coherent, scene-aware rationales that are better aligned with the visual evidence.
We also observe improved performance on both contextual and directional OOV questions, suggesting that the synthetic OOV supervision effectively strengthens their ability to reason unseen regions.

\section{Application to Image Outpainting}
\label{supp:application}

\noindent\textbf{Fine-tuned MLLM Supports Outpainting with Informative Descriptions.}
\label{sec:outpainting_exp}
We demonstrate the practical benefit of the fine-tuned model by improving text guidance for image\&text-conditioned panorama generation.
Instead of relying on manually written multi-view descriptions as in MVDiffusion~\cite{Tang2023mvdiffusion}, we prompt Qwen3-VL-8B-Instruct and its fine-tuned variant to (i) describe the conditioned image, and (ii) iteratively infer adjacent views by rotating 45$^\circ$ to the right, generating descriptions within a 90$^\circ$ FoV and continuing this process for seven successive steps to form a closed viewing loop. 
This setup enables automatic multi-view captioning that reflects the OOV capability of the two models. 

\begin{figure*}[ht]
    \centering
    \includegraphics[width=\linewidth]{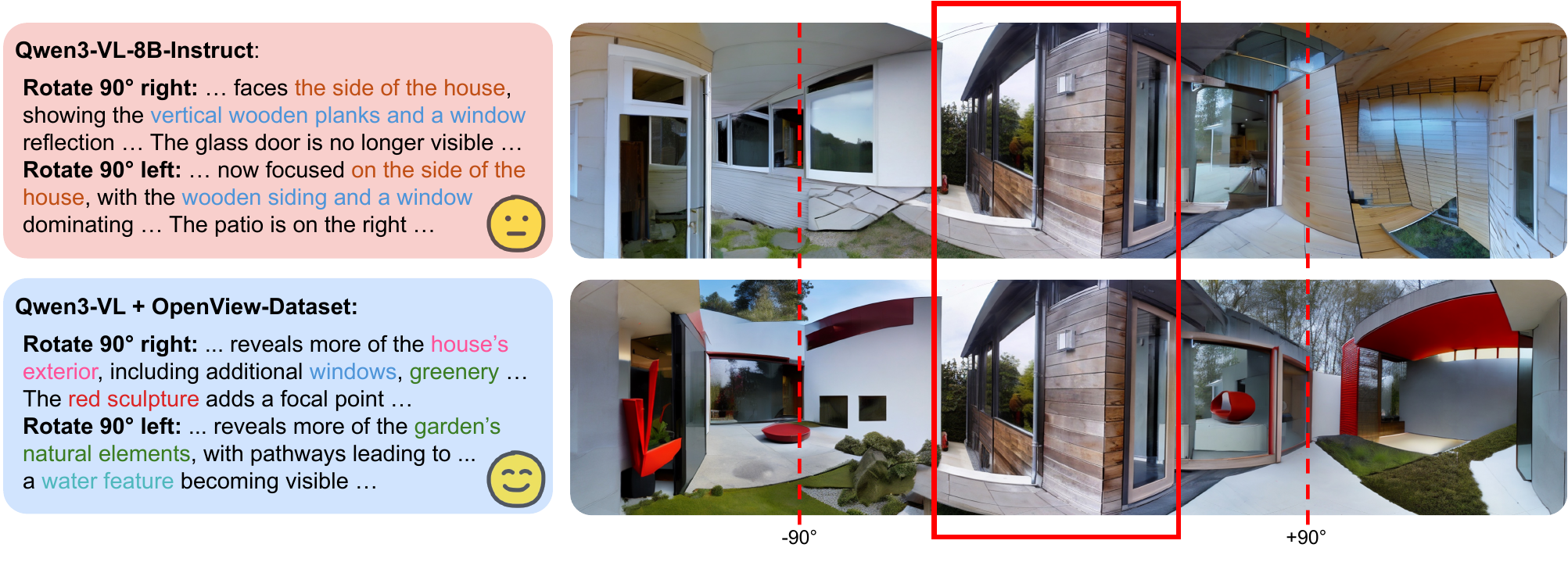}
    \caption{\textbf{Example of outdoor Image\&text-conditioned panorama generation results.} 
    We employ Qwen3-VL-8B-Instuct and its fine-tuned variant as assistants for adjacent-view description generation. 
    Conditioned views are bounded in red. Objects are colorized to highlight repetitive or salient elements.
    For fair OOV comparison, we show generated descriptions for the left and right 90$^\circ$ views, which are non-overlapping yet closest to the conditioned image.
    The fine-tuned model provides more informative captions than base model, it also offers improved guidance for rear-facing views.
    }
    \label{fig:outpainting_1}
\end{figure*}

\begin{figure*}[ht]
    \centering
    \includegraphics[width=\linewidth]{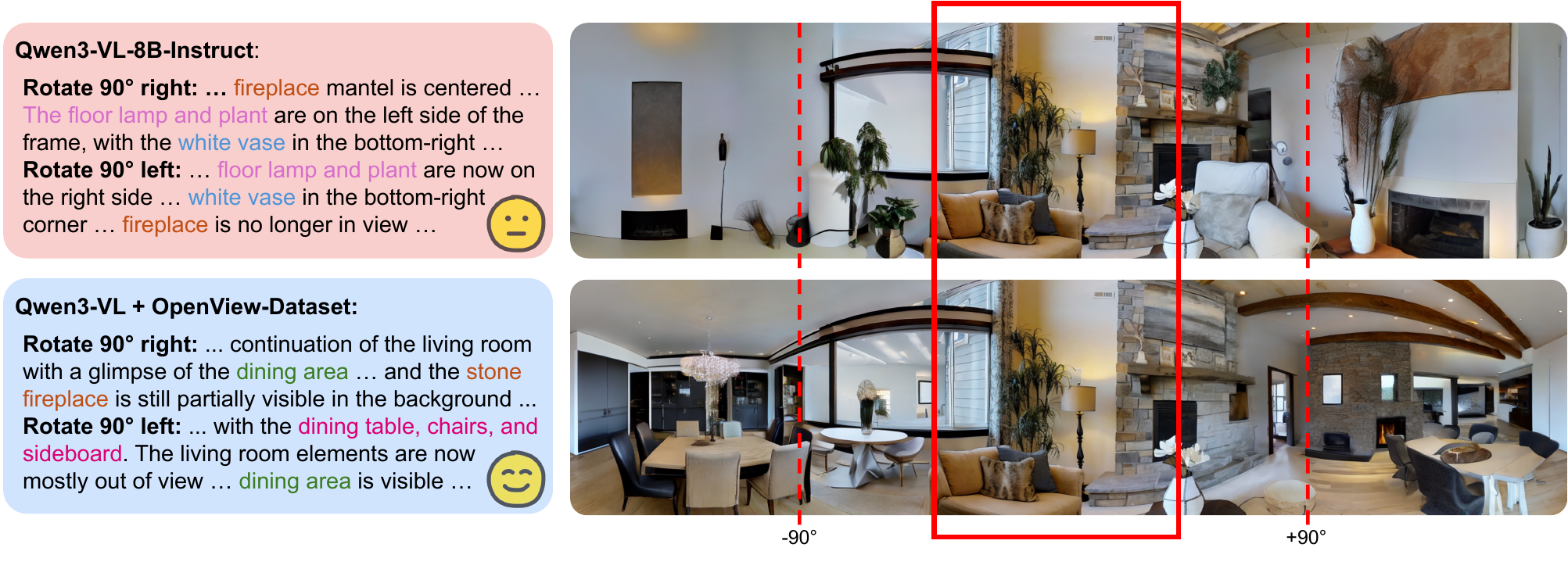}
    \caption{\textbf{Example of indoor Image\&text-conditioned panorama generation results.}
    Conditioned views are bounded in red. Objects are colorized to highlight repetitive or salient elements.
    }
    \label{fig:outpainting_2}
\end{figure*}

\begin{figure*}[ht]
    \centering
    \includegraphics[width=\linewidth]{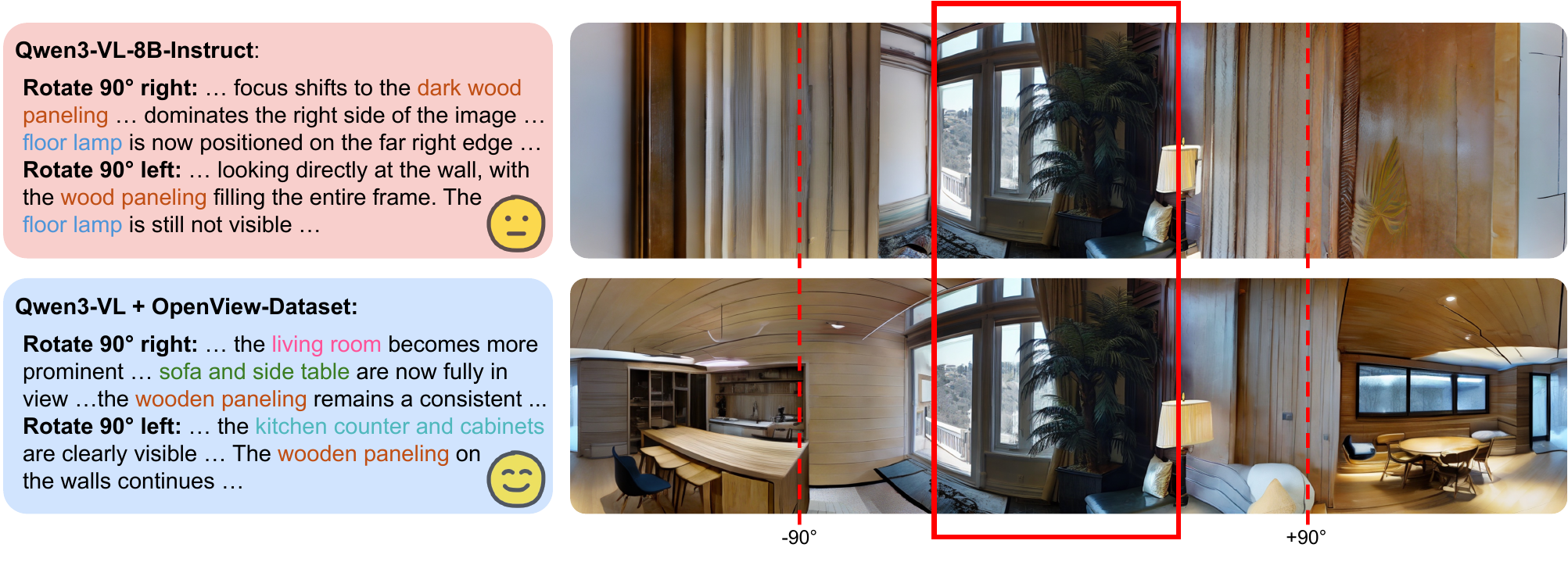}
    \caption{\textbf{Example of indoor Image\&text-conditioned panorama generation results.}
    Conditioned views are bounded in red. Objects are colorized to highlight repetitive or salient elements.
    }
    \label{fig:outpainting_3}
\end{figure*}

\begin{figure*}[ht]
    \centering
    \includegraphics[width=\linewidth]{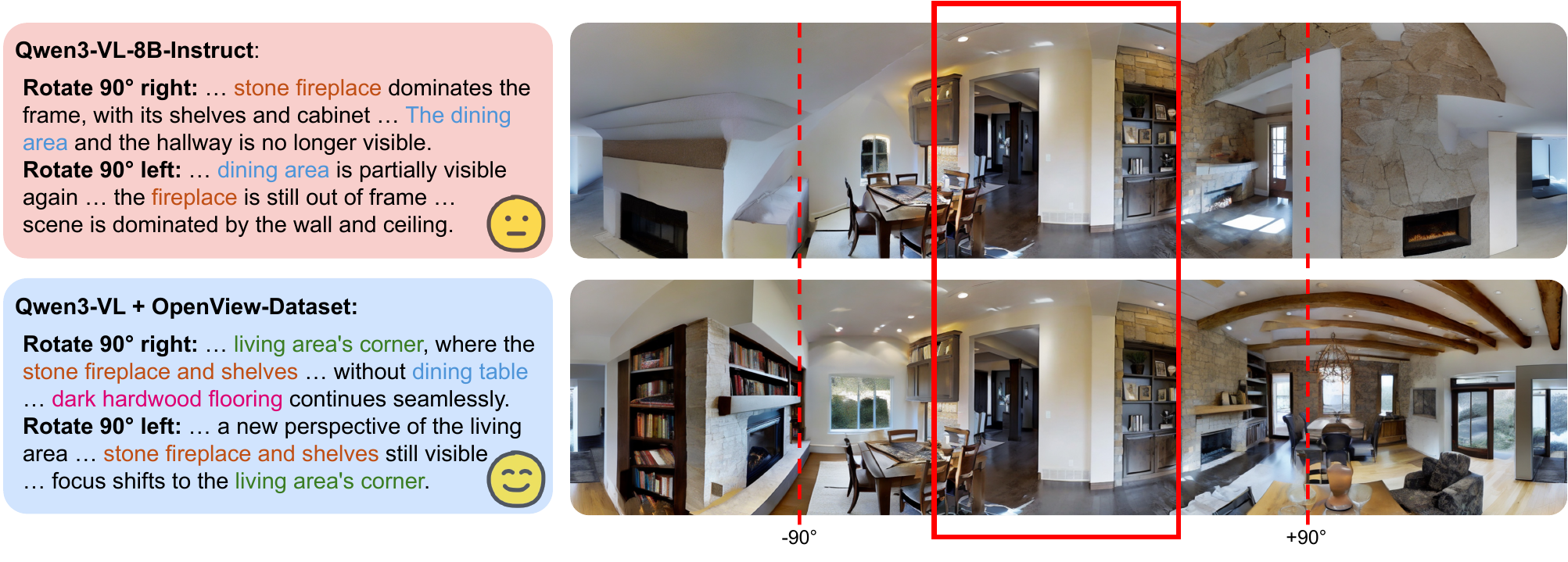}
    \caption{\textbf{Example of indoor Image\&text-conditioned panorama generation results.} 
    Conditioned views are bounded in red. Objects are colorized to highlight repetitive or salient elements.
    }
    \label{fig:outpainting_4}
\end{figure*}

\begin{figure*}[ht]
    \centering
    \includegraphics[width=\linewidth]{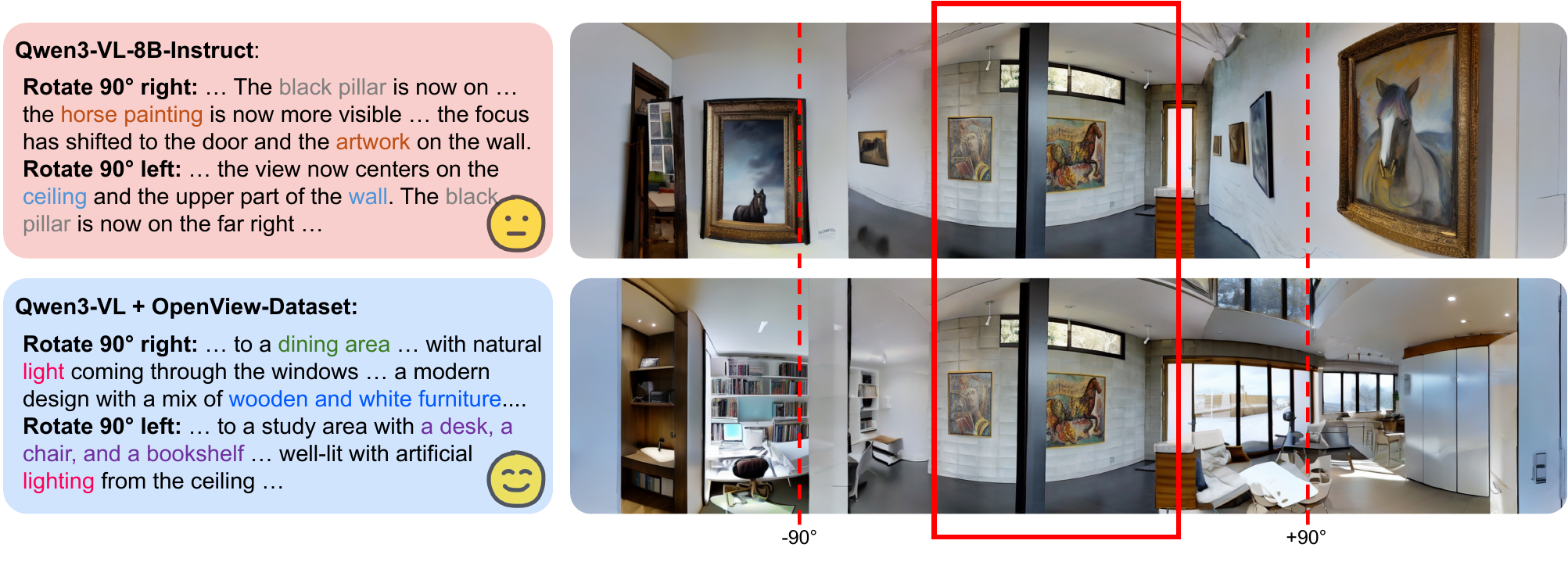}
    \caption{\textbf{Example of indoor Image\&text-conditioned panorama generation results.}
    Conditioned views are bounded in red. Objects are colorized to highlight repetitive or salient elements.
    }
    \label{fig:outpainting_5}
\end{figure*}

\begin{figure*}[ht]
    \centering
    \includegraphics[width=\linewidth]{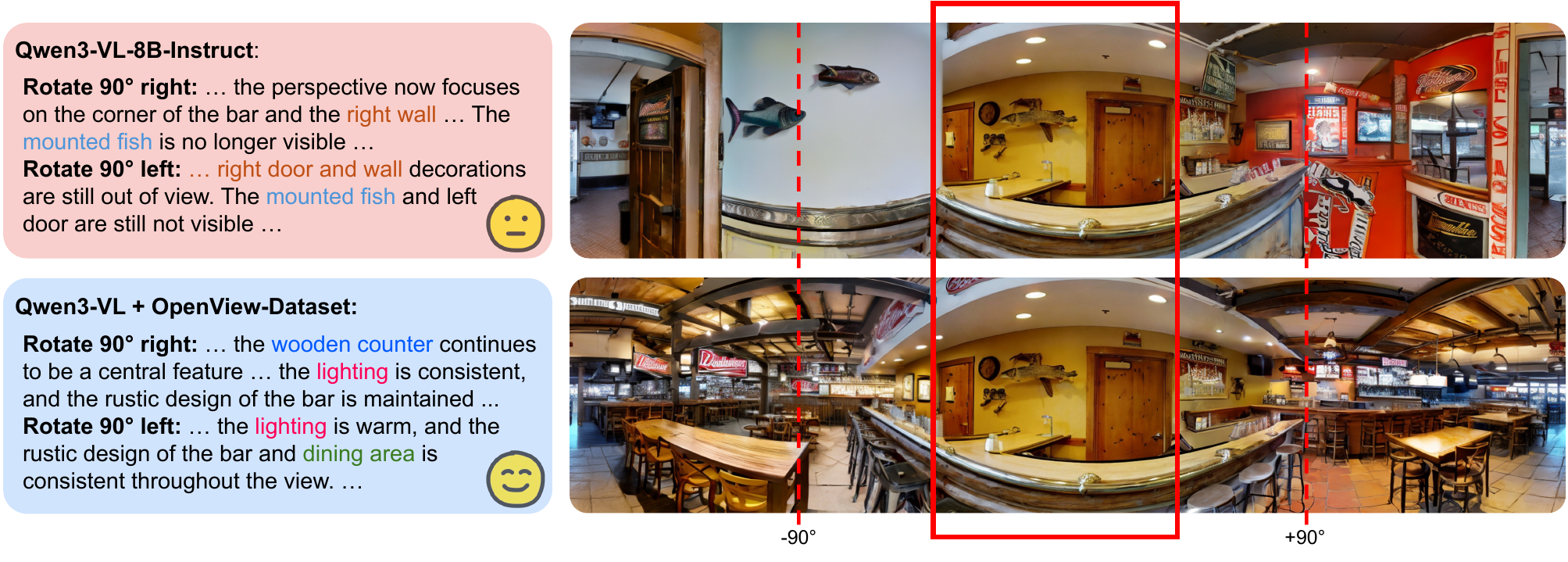}
    \caption{\textbf{Example of indoor Image\&text-conditioned panorama generation results.}
    Conditioned views are bounded in red. Objects are colorized to highlight repetitive or salient elements.
    }
    \label{fig:outpainting_6}
\end{figure*}

Qualitative results in Figure~\ref{fig:outpainting_1} to~\ref{fig:outpainting_6} reveal clear visual differences between the two models.
Although the base model understands the general notion of camera rotation, it often repeats references to objects from the conditioned view and lacks new contextual or spatial details.
In contrast, the fine-tuned model produces more informative and spatial-coherent descriptions by introducing plausible unseen objects and correctly identifies structural transitions in the environment, resulting in more reasonable and complete scene interpretations.
These observations indicate that \ourdataset effectively empowers MLLMs, thereby supporting informative textual conditioning for outpainting and potentially other view-extrapolation tasks. 
Table~\ref{tab:prompt_outpainting} shows the prompt template for producing the textual descriptions conditioned on the given projected view.

\clearpage

\begin{table*}[ht]
\centering
\caption{The prompt template for prompt-based filtering in the Data Preprocessing stage.}
\label{tab:prompt_stage1_filter}
\begin{tcolorbox}[width=\linewidth, colframe=black, colback=cyan!5, boxsep=0mm, arc=2mm, left=4mm, right=4mm, top=2mm, bottom=2mm]
\hspace{-2mm}\textbf{SYSTEM PROMPT}\\[2pt]
You are a helpful panorama checking assistant.\\

\hspace{-2mm}\textbf{USER PROMPT}\\[2pt]
Examine the given image and judge if it is a suitable panorama source.\\

\textbf{Validation criteria:}\\[-4.2mm]
\begin{itemize}[leftmargin=3mm]
    \item[-] format: The image must be a 360$^\circ$ panorama with Equirectangular Projection (ERP). Mark \textbf{invalid} if:
    \begin{itemize}[leftmargin=3mm]
        \item[-] The image looks like non-ERP panoramic format. (e.g., flat perspective, dual fisheye, cube-map, cylindrical, little planet, or any projection other than ERP)
    \end{itemize}
    \item[-] informative: The image must contain clear, meaningful scene content without obstructions. Mark \textbf{invalid} if:
    \begin{itemize}[leftmargin=3mm]
        \item[-] The image contains any watermark, logo, text, or overlay, regardless of size or placement.
        \item[-] The image shows compression artifacts, stitching errors, severe motion blur, or pixelation.
        \item[-] The scene is too dark, obscured, or lacks visible detail (e.g., low-light/night scenes with little visibility).
        \item[-] The content is almost empty or uniform (e.g., mostly blank sky, solid color areas).
        \item[-] The image is rendered from a virtual environment.
    \end{itemize}
\end{itemize}
\vspace{1.5mm}

\textbf{Task constraints:}\\
- Give concise reasons for both judgments.\\
- A distortion is acceptable only if it follows ERP format.\\
- Be strict: if there is any doubt about ERP format or informativeness, mark \textbf{invalid}.\\
- Your response \textbf{MUST} follow the output schema in English.\\

\textbf{Output schema}\\[-4.2mm]
\begin{verbatim}
{
    "format_reason": "<short reason within 20 words>",
    "format": "valid" | "invalid",
    "informative_reason": "<short reason within 20 words>",
    "informative": "valid" | "invalid"
}
\end{verbatim}
\end{tcolorbox}
\end{table*}

\begin{table*}[ht]
\centering
\caption{The prompt template for perspective-projected view analysis in the Visual Analyzer stage.}
\label{tab:prompt_stage2_caption}
\begin{tcolorbox}[width=\linewidth, colframe=black, colback=cyan!5, boxsep=0mm, arc=2mm, left=4mm, right=4mm, top=2mm, bottom=2mm, fontupper=\small]

\hspace{-2mm}\textbf{SYSTEM PROMPT}\\[2pt]
You are a precise visual analyzer. Examine the image and produce a structured, factual description.\\

\textbf{Task constraints:}\\ 
- Describe ONLY what is visible; no external knowledge or speculation.\\ 
- Be objective and accurate; nouns for objects, short phrases for relations.\\ 
- Your response \textbf{MUST} follow the output schema in English.\\

\textbf{Object locations (must use one of these 9 tokens):}\\
top-left, top, top-right, left, center, right, bottom-left, bottom, bottom-right.\\ 
- If an object spans areas, record all the areas it covers.\\ 
- If multiple same instances exist, group them together with a plural noun (e.g., ``benches'') and do not repeat the same object name in the list.\\

\textbf{Output schema}\\[-4.2mm]
\begin{verbatim}
{
    "caption": "<scene description>",
    "objects": ["<object_noun> in/at/on <location_token>", ...],
    "spatial_facts": ["<relations>", ...]
}
\end{verbatim}

\textbf{Notes}\\
- Each item in ``objects'' is a string: ``\textless object\textgreater ~in/at/on \textless location\textgreater''.\\
- If no objects or relations are visible, return ``objects'': [], ``spatial\_facts'': [].\\
- Keep spatial facts concrete (e.g., ``bench (left) faces kiosk (center)'', ``sign (top) above kiosk (center)'').\\

\textbf{Example output}\\[-4.2mm]
\begin{verbatim}
{
    "caption": "A small plaza with benches and a kiosk by ...",
    "objects": [
        "red bench on the left",
        "green bench at the bottom-left",
        "kiosk in the center",
        "sign at the top"
    ],
    "spatial_facts": [
        "red bench (left) faces kiosk (center)",
        "green bench (bottom-left) faces kiosk (center)",
        "sign (top) above kiosk (center)"
    ]
}
\end{verbatim}

\hspace{-2mm}\textbf{USER PROMPT}\\[2pt]
Analyze this image and provide a detailed visual analysis.
\end{tcolorbox}
\end{table*}

\begin{table*}[ht]
\centering
\caption{The prompt template for panorama analysis in the Visual Analyzer stage.}
\label{tab:prompt_stage2_summary}
\begin{tcolorbox}[width=\linewidth, colframe=black, colback=cyan!5, boxsep=0mm, arc=2mm, left=4mm, right=4mm, top=2mm, bottom=2mm]

\hspace{-2mm}\textbf{SYSTEM PROMPT}\\[2pt]
You are a professional panorama summarizer. Your task is to provide a concise, comprehensive overview and high-level understanding of the entire panorama, assign a scene label and check whether the scene is an outdoor scene.\\

You will be given:
\begin{itemize}[leftmargin=3mm]
    \item[-] A list of 0-roll perspective-projected views of the panorama, each with:
    \begin{itemize}[leftmargin=3mm]
        \item[-] uv\_norm: normalized (u,v) coordinates on the panorama, indicating the center of the view.
        \item[-] diag\_FoV: diagonal field of view in degrees.
        \item[-] aspect\_ratio: the view’s width-to-height ratio.
        \item[-] neighbor views: the neighbor views of the view.
        \item[-] visual analysis: visible objects with location in the view and spatial facts.
    \end{itemize}
\end{itemize}
\vspace{1.5mm}
\textbf{Task constraints:}
- Provide a concise and high-level summary that captures the essence of the panorama.\\
- Emphasize the main environment, setting, and key spatial relationships.\\
- Avoid excessive detail that obscures the main scene understanding.\\
- Assign exactly one label from: \texttt{\{scene\_labels\_str\}}\\
- Check the label with the definition above.\\
- Check whether the scene is an outdoor scene.\\
- Your response *MUST* follow the output schema in English.\\

\textbf{Output schema}\\[-4.2mm]
\begin{verbatim}
{
    "summary": "A concise, comprehensive overview and high-level 
    understanding of the entire panorama scene",
    "label": "The label of the scene, *MUST* be one of the label 
    above."
    "outdoor": "Whether the panorama is an outdoor scene, *MUST* 
    be True or False."
}
\end{verbatim}

\hspace{-2mm}\textbf{USER PROMPT}\\[2pt]
Check the list of perspective-projected views of a panorama and the visual analysis of each view as detailed references.\\

Visual analysis: \texttt{\{list\_of\_analyses\}}
\end{tcolorbox}
\end{table*}

\begin{table*}[ht]
\centering
\caption{The base system prompt template (Part 1) in the Proposal Generator stage.}
\label{tab:prompt_stage3_part1}
\begin{tcolorbox}[width=\linewidth, colframe=black, colback=cyan!5, boxsep=0mm, arc=2mm, left=4mm, right=4mm, top=2mm, bottom=2mm]

\hspace{-2mm}\textbf{BASE SYSTEM PROMPT}\\[2pt]
You are a professional Multi-Choice VQA Designer. You will be given:
\begin{itemize}[leftmargin=3mm]
    \item[-] A \textbf{panorama image} in 2:1 aspect ratio.
    \item[-] A \textbf{summary} and the \textbf{category} of the scene.
    \item[-] A list of 0-roll perspective-projected views of the panorama, each with:
    \begin{itemize}[leftmargin=3mm]
        \item[-] uv\_norm: normalized (u,v) coordinates on the panorama, indicating the center of the view.
        \item[-] diag\_FoV: diagonal field of view in degrees.
        \item[-] aspect\_ratio: the view’s width-to-height ratio.
        \item[-] neighbor views: the neighbor views of the view.
        \item[-] visual analysis: visible objects with location in the view and spatial facts.
    \end{itemize}
\end{itemize}
\textbf{Overall objectives:}\\
- The core goal is to test the user’s diverse knowledge and reasoning ability about what lies beyond the directly visible content of the chosen view, and what is possible to observe from out of view.\\
- Questions must require inference about out-of-view functions, context, spatial relations, temporal cues, causal dependencies or commonsense implications.\\
- The full panorama is provided only as design context: it allows you, the question designer, to understand the overall scene in order to craft out-of-view reasoning questions.\\
- You will design several multi-choice QA pairs with different perspective-projected views from the panorama. The user will see only this selected view when answering the questions, not the panorama.\\
- Each question should encourage reasoning that connects visible evidence in the view with what is likely outside of it, avoiding trivial tasks such as object detection, counting, or describing what is directly seen.\\
- The QA must be challenging, informative, and non-trivial: answering correctly should require bridging in-view cues with out-of-view reasoning rather than relying on surface-level observation.\\
- Your response \textbf{MUST} follow the output schema in English.\\

\textbf{Your design workflow must follow three explicit reasoning steps:}

\#\#\# \textbf{Step 1 - View Reasoning}
\begin{itemize}[leftmargin=3mm]
    \item[-] First, review the full panorama image along with its summary and scene category to understand the global context. Then, frame an appropriate perspective-projected view location from anywhere on the panorama for question design.
    \item[-] Choose u\_norm and v\_norm $\in [0,1]$ to specify the center of the selected view based on your reasoning.
    \item[-] Choose a diag\_FoV $\in [40,100]$ and an aspect\_ratio $\in $ \{4:3, 3:4, 3:2, 2:3, 16:9, 9:16, 1:1\} to best frame the reasoning target.
    \item[-] Selection should maximize the potential to design a challenging, reasoning-based VQA.
\end{itemize}
\end{tcolorbox}
\end{table*}

\begin{table*}[ht]
\centering
\caption{The base system prompt template (Part 2) in the Proposal Generator stage.}
\label{tab:prompt_stage3_part2}
\begin{tcolorbox}[width=\linewidth, colframe=black, colback=cyan!5, boxsep=0mm, arc=2mm, left=4mm, right=4mm, top=2mm, bottom=2mm]
\begin{itemize}[leftmargin=3mm]
    \item[-] Consider:
        \begin{itemize}[leftmargin=3mm]
            \item[-] Question potential: select a region whose visible cues best support reasoning about out-of-view context, relations, or commonsense implications rather than simple recognition or localization.
            \item[-] View diversity: avoid redundant or trivial viewpoints; sample positions as diverse as possible and avoid regions dominated by artifacts (e.g., bottom camera rig or tripod).
            \item[-] Evidence sufficiency: ensure the chosen view provides enough contextual cues to justify the reasoning required in the planned question.
            \item[-] FoV use: Adjust the field of view according to the size and distance of the main object/subject. Use narrow FoV $\in [40^\circ-70^\circ]$ to constrain information and highlight decisive details for reasoning, especially for distant or small objects. Use wide FoV $\in [70^\circ-100^\circ]$ when nearby or large elements require more surrounding context to remain interpretable. Always aim to provide limited but sufficient visual evidence that forces deeper reasoning, rather than making the answer obvious.
            \item[-] Aspect ratio: landscape (16:9, 4:3, 3:2) for wide settings; portrait (3:4, 2:3, 9:16) for tall/narrow contexts; 1:1 for balanced or centered views. Be cautious with tall ratios that might expose upper/lower adjacent view content; maintain perspective consistency.
            \item[-] Output the justification before the parameters as ``view\_reasoning", then provide ``u\_norm'', ``v\_norm'', ``diag\_FoV'', ``aspect\_ratio''.
        \end{itemize}
\end{itemize}
\#\#\# \textbf{Step 2 - Question Reasoning}
\begin{itemize}[leftmargin=3mm]
    \item[-] Design an out-of-view multi-choice question based on the chosen view, taking into account the adjusted FoV and aspect ratio from Step 1.
    \item[-] Question design: Inspect the selected view region carefully, and use the corresponding analysis from the view list as reference when constructing the question. Design a question asking the model to choose the most or least plausible option.
    \item[-] Option design: Provide exactly five options (option\_a to option\_e).
    \begin{itemize}[leftmargin=3mm]
        \item[-] option\_a to option\_d: must be plausible, mutually exclusive, and non-trivial distractors or candidates.
        \item[-] option\_e: a fixed interference option with a logical relation to the others (e.g., ``None of the above'', ``All of the above'', ``Both A and C''). This must sometimes serve as the correct answer.
        \item[-] All options should be concise and avoid absurdity, correctness must depend on reasoning with the chosen view + commonsense.
    \end{itemize}
    \item[-] Reason explicitly about:
    \begin{itemize}[leftmargin=3mm]
        \item[-] View-Question fit: why this view enables the question; what cues support the intended inference.
        \item[-] Option design: how the one correct option is truly defensible among the four distractors.
        \item[-] Reasoning demand: how the question forces integration of visible cues with commonsense/contextual knowledge for out-of-view reasoning, rather than simple recognition or counting.
    \end{itemize}
\end{itemize}
\end{tcolorbox}
\end{table*}

\begin{table*}[ht]
\centering
\caption{The base system prompt template (Part 3) in the Proposal Generator stage.}
\label{tab:prompt_stage3_part3}
\begin{tcolorbox}[width=\linewidth, colframe=black, colback=cyan!5, boxsep=0mm, arc=2mm, left=4mm, right=4mm, top=2mm, bottom=2mm]
\begin{itemize}[leftmargin=3mm]
    \item[-] No leakage: do not reference knowledge that is not in the chosen view but from the panorama; all reasoning must be legitimate from the chosen view’s context.
    \item[-] Make challenge: frame the stem as brief and general (e.g., ``In this scenario...'', ``Given the view...''), without naming specific visible objects or narrow categories. Avoid giving hints that reveal the answer.
    \item[-] Output: Provide ``question\_reasoning'' first, then ``question'', ``option\_a" to ``option\_e'', and the selected ``answer''.
\end{itemize}

\#\#\# \textbf{Step 3 - Answer Reasoning}
\begin{itemize}[leftmargin=3mm]
    \item[-] Keep the selected correct answer from Step 2 unchanged.
    \item[-] Do not reference or rely on panorama-wide or unseen information when reasoning.
    \item[-] Provide concise and individual reasoning for each option (``option\_a''-``option\_e''), based solely on visible cues from the chosen view and relevant knowledge.
    \begin{itemize}[leftmargin=3mm]
        \item[-] Explain why the option could be plausible or correct given the view.
        \item[-] Explain why it is ultimately less plausible or incorrect, if it’s not the correct answer.
    \end{itemize}
    \item[-] After describing all options, provide a short contrastive conclusion that:
    \begin{itemize}[leftmargin=3mm]
        \item[-] Summarizes why the chosen answer is the most defensible based on the view.
        \item[-] Briefly contrasts it with why each distractor fails or is less consistent with the visible evidence.
        \item[-] Do not include the option name (A, B, C, D, E) in the reasoning, just describe the option in general terms.
    \end{itemize}
    \item[-] After reasoning, give a confidence score for the design of the proposal in the range of 1(low) to 3(high).
    \item[-] No need to explain the confidence score in any reasoning.
    \item[-] Output this reasoning as ``option\_a\_reasoning'' to ``option\_e\_reasoning'', ``conclusion\_reasoning'' and ``confidence\_score'' as the last seven fields.
\end{itemize}

\textbf{Output schema}\\[-4.2mm]
\begin{verbatim}
[{
    "view_reasoning": "<why this uv, diag_fov, aspect_ratio was chosen>",
    "u_norm": "<float in [0,1]>", "v_norm": "<float in [0,1]>",
    "diag_fov": "<float in [40, 100]>", 
    "aspect_ratio": "<string in [4:3, 3:4, 3:2, 2:3, 16:9, 9:16, 1:1]>", 
    "question_reasoning": "<why this question, options and answer 
    are suitable designed for the view>", "question": "<string>", 
    "option_a": "<string>", ..., "option_e": "<string>", "answer": 
    "<string>", "option_a_reasoning", ..., "option_e_reasoning", 
    "conclusion_reasoning", "confidence_score": "<int in [1, 3]>",
 }, {...}, ...]
\end{verbatim}
\end{tcolorbox}
\end{table*}

\begin{table*}[ht]
\centering
\caption{The system prompt template for contextual task design in the Proposal Generator stage.}
\label{tab:prompt_stage3_contextual}
\begin{tcolorbox}[width=\linewidth, colframe=black, colback=cyan!5, boxsep=0mm, arc=2mm, left=4mm, right=4mm, top=2mm, bottom=2mm]
\hspace{-2mm}\textbf{SYSTEM PROMPT FOR CONTEXTUAL QUESTIONS}\\[2pt]
\textbf{Question type:} Contextual question

\textbf{Design Objectives:}\\
- Create multi-choice questions that test whether the user can judge which objects, actions, conditions, or scenarios are plausible or implausible outside of the given view.\\
- First, framing a base view. Then, finding its neighbor views, and use it only during VQA generation as the ground truth for the correct option.\\
\textbf{Task constraints:}\\[-4.2mm]
\begin{itemize}[leftmargin=3mm]
    \item[-] Base view coordinate (u,v) framed from the panorama. You may adjust diag\_fov and aspect\_ratio to provide limited but sufficient evidence.
    \item[-] The chosen view should provide contextual cues (environment type, spatial layout, activities) that support reasoning about out-of-view plausibility.
    \item[-] The user can not see the neighbor views, they should rely only on cues visible in the chosen view plus commonsense/contextual reasoning.
    \item[-] No panorama leakage: In question stem, options and reasoning must never reference the panorama or the neighbor view directly.
    \item[-] \textbf{Question stem:}
    \begin{itemize}[leftmargin=3mm]
        \item[-] Keep the question stem brief and neutral, without describing any visible contents. Frame it in general terms (e.g., ``Which object/event/condition would most likely appear outside of the view?'' or ``Which option would be least plausible to see outside of the view?'').
    \end{itemize}
    \item[-] \textbf{Options:}
    \begin{itemize}[leftmargin=3mm]
        \item[-] Options may be single items (e.g., ``umbrella'') or small sets/lists of items (e.g., ``apple, banana, and orange'').
        \item[-] Options must be mutually exclusive, with distractors that are contextually reasonable but incorrect. Avoid absurd or random sets. 
        \item[-] Set of items should form a correlative group (``ski poles, snow boots, sled'', not ``lamp, fish, shoe''). 
        \item[-] Avoid speculative or overly detailed predictions that cannot be inferred with confidence from the base view.
    \end{itemize}
    \item[-] \textbf{Answer reasoning:}
    \begin{itemize}[leftmargin=3mm]
        \item[-] Justify the correct answer based on cues in the base view (e.g., visible building edges, road alignment, horizon, continuation of features).
        \item[-] Do not justify correctness by referencing what is seen in the neighbor view. Neighbor view is only a hidden reference to ensure one option is correct.
    \end{itemize}
    \item[-] \textbf{Confidence score:} Provide a confidence score for the proposal.
\end{itemize}

\textbf{Examples of valid questions:}\\[-4.2mm]
\begin{itemize}[leftmargin=3mm]
    \item[-] Which of the following sets of objects would most likely appear outside?
    \begin{itemize}[leftmargin=3mm]
        \item[-] option\_a: Beach ball, umbrella, and towel 
        \item[-] option\_b: Ski poles, snow boots, and sled
        \item[-] option\_c: Laptop, projector, and whiteboard
        \item[-] option\_d: Pots, pans, and oven mitts
        \item[-] option\_e: Both C and D
    \end{itemize}
    \item[-] Which of the following activities would be least likely to occur nearby?
    \item[-] Which type of seating would be most plausible in this area?
\end{itemize}
\end{tcolorbox}
\end{table*}

\begin{table*}[ht]
\centering
\caption{The system prompt template for directional task design in the Proposal Generator stage.}
\label{tab:prompt_stage3_directional}
\begin{tcolorbox}[width=\linewidth, colframe=black, colback=cyan!5, boxsep=0mm, arc=2mm, left=4mm, right=4mm, top=2mm, bottom=2mm]
\hspace{-2mm}\textbf{SYSTEM PROMPT FOR DIRECTIONAL QUESTIONS}\\[2pt]
\textbf{Question type:} Directional question

\textbf{Design objectives:}\\
- Test whether the user can infer the out-of-view content conditioned on a specified camera rotation (left, right, up, down or set of rotations), using only the current base view’s cues.\\
- First, framing a base view. Then, defining rotation instruction(s) (e.g., ``turn left 60$^\circ$'', ``tilt up slightly'', ``rotate right 30$^\circ$'', ``turn left 45$^\circ$ and tilt up slightly'') for the prediction task. Finally, find the neighbor view and use it only during VQA generation as the ground truth for the correct option. The user can not see this neighbor view.

\textbf{Task constraints:}\\[-4.2mm]
\begin{itemize}[leftmargin=3mm]
    \item[-] Base view coordinate (u,v) framed from the panorama. You may adjust diag\_fov and aspect\_ratio to provide limited but sufficient evidence.
    \item[-] No panorama leakage: In question stem, options and reasoning must never reference the panorama or the neighbor view directly.
    \item[-] Handle uncertainty by refining the stem (e.g., ``immediately visible,'' ``dominant feature,'' ``center of frame'') so the prediction is focused and non-ambiguous, and make the question challenging.
    \item[-] \textbf{Question stem:}
    \begin{itemize}[leftmargin=3mm]
        \item[-] Keep the question stem brief and neutral, without describing any visible contents. Frame it in general terms. The rotation instruction may be single or two camera rotations.
    \end{itemize}
    \item[-] \textbf{Options:}
    \begin{itemize}[leftmargin=3mm]
        \item[-] Exactly one option must align with the ground truth neighbor view, but phrased only in terms of what is reasonably implied from the base view + rotation instructions.
        \item[-] Distractors must be plausible in the broader scene type but logically contradict the rotated direction or visible cues from the base view.
        \item[-] Avoid speculative or overly detailed predictions that cannot be inferred with confidence from the base view.
    \end{itemize}
    \item[-] \textbf{Answer reasoning:}
    \begin{itemize}[leftmargin=3mm]
        \item[-] Justify the correct answer based on cues in the base view (e.g., visible building edges, road alignment, horizon, continuation of features).
        \item[-] Do not justify correctness by referencing what is seen in the neighbor view. Neighbor view is only a hidden reference to ensure one option is correct.
    \end{itemize}
    \item[-] \textbf{Confidence score:} Provide a confidence score for the proposal.
\end{itemize}

\textbf{Examples of valid questions:}\\[-4.2mm]
\begin{itemize}[leftmargin=3mm]
    \item[-] If you turn left about 40$^\circ$ and tilt up slightly, what feature would most likely come into view first?
    \item[-] After tilting upward slightly, which element would you most expect to see appear?
    \item[-] By rotating right about 60$^\circ$, what would you most likely see prominently?
\end{itemize}
\end{tcolorbox}
\end{table*}

\begin{table*}[ht]
\centering
\caption{The user prompt template in the Proposal Generator stage.}
\label{tab:prompt_stage3_user}
\begin{tcolorbox}[width=\linewidth, colframe=black, colback=cyan!5, boxsep=0mm, arc=2mm, left=4mm, right=4mm, top=2mm, bottom=2mm]
\hspace{-2mm}\textbf{USER PROMPT}\\[2pt]
Scene Category: \texttt{\{category\}}

Summary of the panorama:
\texttt{\{summary\}}

Perspective-projected views and visual analysis of each view:
\texttt{\{list\_of\_analyses\}}

Task:
Generate \texttt{\{k\}} VQAs.  
Each question must follow the JSON schema, return a \textbf{JSON list}.
\end{tcolorbox}
\end{table*}

\begin{table*}[ht]
\centering
\caption{The prompt template for quality control in the Proposal Refiner stage.}
\label{tab:prompt_stage4}
\begin{tcolorbox}[width=\linewidth, colframe=black, colback=cyan!5, boxsep=0mm, arc=2mm, left=4mm, right=4mm, top=2mm, bottom=2mm]

\hspace{-2mm}\textbf{SYSTEM PROMPT}\\[2pt]
You are a strict JSON format corrector. \\

You will be given a string that should represent a JSON list, but it may contain multiple formatting errors. Check the error message carefully and fix the issues. After fixing the current error, re-check for further errors until the output is fully valid JSON.

\textbf{Task constraints:}
\begin{itemize}
    \item[-] Fix only formatting errors (commas, quotes, colons, brackets, extra/trailing chars, markdown fences, etc.).  
    \item[-] Do not modify keys or values, only repair structure.
    \item[-] Clean redundant list items if applicable.  
    \item[-] Remove any URLs in the string.  
    \item[-] For any list mixing strings and dicts, output as a JSON array of strings:  
    \begin{itemize}
        \item[-] Example: input `[``Person'', ``Car'':``moving'']' → output `[``Person'', ``Car: moving'']'.  
        \item[-] Ensure no colon appears outside of quotes.  
    \end{itemize}
    \item[-] Replace all '' with ' inside string values. 
    \item[-] Final output must be valid JSON and nothing else (no explanations).
\end{itemize}
\vspace{1.5mm}

\hspace{-2mm}\textbf{USER PROMPT}\\[2pt]
Raw message: \texttt{\{raw\_content\}}\\
Error message: \texttt{\{error\_msg\}}
\end{tcolorbox}
\end{table*}

\begin{table*}[ht]
\centering
\caption{The user prompt template for inference in \ourbenchnospace.}
\label{tab:prompt_inference}
\begin{tcolorbox}[width=\linewidth, colframe=black, colback=cyan!5, boxsep=0mm, arc=2mm, left=4mm, right=4mm, top=2mm, bottom=2mm]
\hspace{-2mm}\textbf{USER PROMPT}\\[2pt]
Question: \texttt{\{question\}}\\
Options: \texttt{\{options\}}\\
Instructions:\\
Analyze each option's correctness with reasoning or justification based on the image and the question.
Then, conclude with the single correct option in the format: \texttt{<answer>...</answer>}. 
\end{tcolorbox}
\end{table*}

\begin{table*}[ht]
\centering
\caption{The user prompt template for rationale correctness evaluation in \ourbenchnospace.}
\label{tab:prompt_eval}
\begin{tcolorbox}[width=\linewidth, colframe=black, colback=cyan!5, boxsep=0mm, arc=2mm, left=4mm, right=4mm, top=2mm, bottom=2mm]
\hspace{-2mm}\textbf{USER PROMPT}\\[2pt]
You are a rigorous evaluator for an out-of-view (OOV) multi-choice visual question answering task.
Your job is to determine whether the model's response is logically consistent with the ground truth and factual correct. Notice that the image is not provided for this evaluation. Evaluate both the option-level rationale and the final answer justification holistically.\\

Case of INCORRECT:\\
1. Any answers based primarily on guessing, imagination or scene stereotypes without referring to image evidence.\\
2. Any rationale that only say ``not visible'', ``unlikely'', ``probably'', or other vague statements without supporting evidence.\\
3. Missing the rationale for any five options.\\
4. Responses that omit essential evidence, or only provide the final choice should be considered incorrect. \\

Important reminders:\\
- Long explanations do NOT necessarily mean correct reasoning.\\
- Be strict with each criterion.\\

Question: \texttt{\{question\}}\\
Options: \texttt{\{options\}}\\
Ground Truth Answer: \texttt{\{answer\_rationale\}}\\
Response (Rationale): \texttt{\{response\}}\\

Respond only with one of the following:
\texttt{<answer>Yes</answer>} or \texttt{<answer>No</answer>}.
\end{tcolorbox}
\end{table*}

\begin{table*}[ht]
\centering
\caption{The prompt template for generating textual descriptions from a single view, serving as a semantic support for outpainting.}
\label{tab:prompt_outpainting}
\begin{tcolorbox}[width=\linewidth, colframe=black, colback=cyan!5, boxsep=0mm, arc=2mm, left=4mm, right=4mm, top=2mm, bottom=2mm]

\hspace{-2mm}\textbf{USER PROMPT}\\[2pt]
\textbf{Instructions:}\\
Examine the provided image and generate detailed descriptions for eight sequential perspective views. 
Each view should have a horizontal field of view of 90°, with each subsequent view rotated 45° to the right from the previous one (resulting in a 50\% overlap between adjacent views).
Start from the current viewpoint (the original image) and continue rotating right until all seven rotated views are covered.\\

Here is the output schema:\\[-4.2mm]
\begin{verbatim}
<current_view> description of the current view here </current_view>
<view_1> description of the first rotated view here </view_1>
<view_2> description of the second rotated view here </view_2>
<view_3> description of the third rotated view here </view_3>
<view_4> description of the fourth rotated view here </view_4>
<view_5> description of the fifth rotated view here </view_5>
<view_6> description of the sixth rotated view here </view_6>
<view_7> description of the seventh rotated view here </view_7>
\end{verbatim}
\end{tcolorbox}
\end{table*}

\clearpage


\newpage
\section*{NeurIPS Paper Checklist}

\begin{enumerate}

\item {\bf Claims}
    \item[] Question: Do the main claims made in the abstract and introduction accurately reflect the paper's contributions and scope?
    \item[] Answer: \answerYes{} 
    \item[] Justification: The main claims in the abstract and introduction largely align with the presented contributions, including the proposed benchmark and analysis of out-of-view reasoning.

\item {\bf Limitations}
    \item[] Question: Does the paper discuss the limitations of the work performed by the authors?
    \item[] Answer: \answerYes{} 
    \item[] Justification: We discuss the limitations related to the use of synthetic data at the end of the main paper.

\item {\bf Theory assumptions and proofs}
    \item[] Question: For each theoretical result, does the paper provide the full set of assumptions and a complete (and correct) proof?
    \item[] Answer: \answerNA{} 
    \item[] Justification: The paper does not present formal theoretical proof or results.

    \item {\bf Experimental result reproducibility}
    \item[] Question: Does the paper fully disclose all the information needed to reproduce the main experimental results of the paper to the extent that it affects the main claims and/or conclusions of the paper (regardless of whether the code and data are provided or not)?
    \item[] Answer: \answerYes{} 
    \item[] Justification: The paper provides sufficient details to support reproducibility of the main experimental results in Section~\ref{sec:methods}, Section~\ref{sec:bench} and Section~\ref{sec:experiments}. It clearly describes the dataset construction pipeline, benchmark settings, evaluation protocols, and model configurations.

\item {\bf Open access to data and code}
    \item[] Question: Does the paper provide open access to the data and code, with sufficient instructions to faithfully reproduce the main experimental results, as described in supplemental material?
    \item[] Answer: \answerYes{} 
    \item[] Justification: The code and links to the dataset and benchmark annotations are publicly available at \url{https://github.com/q1xiangchen/OpenView}. The repository provides instructions for data preparation, dataset generation, inference, and evaluation.

\item {\bf Experimental setting/details}
    \item[] Question: Does the paper specify all the training and test details (e.g., data splits, hyperparameters, how they were chosen, type of optimizer) necessary to understand the results?
    \item[] Answer: \answerYes{} 
    \item[] Justification: The paper provides sufficient experimental details, including training setup, evaluation protocols, and inference settings, in both main paper and appendix.

\item {\bf Experiment statistical significance}
    \item[] Question: Does the paper report error bars suitably and correctly defined or other appropriate information about the statistical significance of the experiments?
    \item[] Answer: \answerNo{} 
    \item[] Justification: The paper does not report error bars, which is due to the computational cost of multi-runs for more than ten MLLMs.

\item {\bf Experiments compute resources}
    \item[] Question: For each experiment, does the paper provide sufficient information on the computer resources (type of compute workers, memory, time of execution) needed to reproduce the experiments?
    \item[] Answer: \answerYes{} 
    \item[] Justification: The paper provides sufficient information on computational resources, including reported costs of the OpenView pipeline in Section~\ref{sec:inference_cost} and details of the GPUs used for training in Section~\ref{sec:experiments}.
    
\item {\bf Code of ethics}
    \item[] Question: Does the research conducted in the paper conform, in every respect, with the NeurIPS Code of Ethics \url{https://neurips.cc/public/EthicsGuidelines}?
    \item[] Answer: \answerYes{} 
    \item[] Justification: The research adheres to the NeurIPS Code of Ethics, with no apparent ethical concerns. It focuses on dataset construction and model evaluation without involving sensitive data or harmful applications.

\item {\bf Broader impacts}
    \item[] Question: Does the paper discuss both potential positive societal impacts and negative societal impacts of the work performed?
    \item[] Answer: \answerYes{} 
    \item[] Justification: The paper discusses both positive and negative societal impacts in the beginning of the appendix.
    
\item {\bf Safeguards}
    \item[] Question: Does the paper describe safeguards that have been put in place for responsible release of data or models that have a high risk for misuse (e.g., pre-trained language models, image generators, or scraped datasets)?
    \item[] Answer: \answerNA{} 
    \item[] Justification: The paper does not introduce specific safeguards, as the released data and models are based on publicly available sources, we follow and release them ``for research purposes only'' and ``no redistribution''.

\item {\bf Licenses for existing assets}
    \item[] Question: Are the creators or original owners of assets (e.g., code, data, models), used in the paper, properly credited and are the license and terms of use explicitly mentioned and properly respected?
    \item[] Answer: \answerYes{} 
    \item[] Justification: The paper properly credits the original sources of datasets and models used, with appropriate citations. The licenses and terms of use for these assets are respected.

\item {\bf New assets}
    \item[] Question: Are new assets introduced in the paper well documented and is the documentation provided alongside the assets?
    \item[] Answer: \answerYes{} 
    \item[] Justification: We introduce a syntheic dataset and human-verifed benchmark based on the five existing datasets with new annotations. Details are mentioned in Section~\ref{sec:dataset_analysis}, Section~\ref{supp:more_dataset} and Section~\ref{supp:more_benchmark}.

\item {\bf Crowdsourcing and research with human subjects}
    \item[] Question: For crowdsourcing experiments and research with human subjects, does the paper include the full text of instructions given to participants and screenshots, if applicable, as well as details about compensation (if any)? 
    \item[] Answer: \answerNo{} 
    \item[] Justification: Benchmark verification and the user study involved participants known to the authors who took part voluntarily without compensation. No identifying or sensitive participant information was collected. Section~\ref{supp:more_benchmark} describes the procedures and provides a screenshot of the verification interface, but does not reproduce the full text of participant instructions.

\item {\bf Institutional review board (IRB) approvals or equivalent for research with human subjects}
    \item[] Question: Does the paper describe potential risks incurred by study participants, whether such risks were disclosed to the subjects, and whether Institutional Review Board (IRB) approvals (or an equivalent approval/review based on the requirements of your country or institution) were obtained?
    \item[] Answer: \answerNo{} 
    \item[] Justification: Participants voluntarily verified VQA items and answered benchmark questions. No identifying or sensitive participant information was collected for these activities.

\item {\bf Declaration of LLM usage}
    \item[] Question: Does the paper describe the usage of LLMs if it is an important, original, or non-standard component of the core methods in this research? Note that if the LLM is used only for writing, editing, or formatting purposes and does \emph{not} impact the core methodology, scientific rigor, or originality of the research, declaration is not required.
    \item[] Answer: \answerYes{} 
    \item[] Justification: The paper clearly describes the use of LLMs as part of the core methodology, including their role in dataset construction and evaluation, making their usage explicit and reproducible.

\end{enumerate}

\end{document}